\documentclass{article}

\usepackage{PRIMEarxiv}

\usepackage{amssymb}
\usepackage{comment}
\usepackage{mathtools}
\usepackage{amsmath}
\usepackage{lscape}
\usepackage{subfig}
\usepackage{graphicx}
\usepackage{etoolbox}
\usepackage{multirow}
\usepackage{multicol}
\usepackage{lscape}
\usepackage{makecell}
\usepackage[dvipsnames]{xcolor}
\usepackage{longtable,geometry}
\usepackage{rotating}
\usepackage[square,numbers]{natbib}

\graphicspath{{media/}}     % organize your images and other figures under media/ folder

\title{Semantic relatedness in DBpedia: \\ A comparative and experimental assessment
\thanks{\textit{\underline{Accepted manuscript}}:\\ 
\textbf{Cite as:} Anna Formica, Francesco Taglino.
Semantic relatedness in DBpedia: A comparative and experimental assessment, Information Sciences, Volume 621, April 2023, 474-505, ISSN 0020-0255, https://doi.org/10.1016/j.ins.2022.11.025.} 
}

\author{
  Anna Formica, Francesco Taglino \\
  Istituto di Analisi dei Sistemi ed Informatica ``Antonio Ruberti''\\
  Consiglio Nazionale delle Ricerche (IASI-CNR)\\
  Via dei Taurini 19\\
  I-00185, Rome\\
  Italy\\
\texttt{\{anna.formica, francesco.taglino\}@iasi.cnr.it} \\
}

\begin{document}
\maketitle

\begin{abstract}

Evaluating semantic relatedness of Web resources is still an open challenge. This paper focuses on knowledge-based methods,  which represent an alternative to corpus-based approaches, and rely in general on the availability of knowledge graphs. In particular, we have selected 10 methods from the existing literature, that have been organized according to  {\it adjacent resources}, {\it triple patterns}, and {\it triple weights} based methods. They have been implemented and evaluated by using DBpedia as reference RDF knowledge graph. Since DBpedia is continuously evolving, the experimental results provided by these methods in the literature are not comparable. For this reason, in this work such methods have been experimented by running them all at once on the same DBpedia release and against 14 well-known {\it golden} datasets. On the basis of the correlation values with human judgment obtained according to the experimental results,  weighting the RDF triples in combination with evaluating {\it all} the directed paths linking the compared resources is  the best strategy in order to compute semantic relatedness in DBpedia.

\end{abstract}


%\end{frontmatter}

%\linenumbers 

\section{Introduction}
\label{sec:Introduction}

{\it How much are two given words related?} In general, the way of automatically computing a degree of relatedness between words falls into one of the following categories of methods \cite{Taieb2019ASO}: \emph{corpus-based methods}, which use large corpora of natural language texts, and exploit co-occurrences of words, as for instance \cite{Ponza2020}; \emph{knowledge-based methods}, which rely on structured resources, as for instance \cite{ElVaigh2020}; and \emph{hybrid methods}, which are a mix of the two, as for instance \cite{Mohamed2O20}. 
Corpus-based methods benefit from the huge availability of textual documents and the advancements in the field of natural language processing and, for this reason, they have been widely investigated  in the literature for a long time.
Knowledge-based methods mainly depend on the availability and the quality of a proper knowledge base, such as a knowledge graph or an ontology. These methods require words to be associated with resources in the knowledge base in order to shift from a pure linguistic dimension to a knowledge-based one.
This paper focuses on knowledge-based methods.

Since the advent of the Semantic Web, ontologies have become significant knowledge representation tools, especially when advanced reasoning is required.
However, ontologies suffer from some drawbacks: (i) they are usually manually or semi-manually created and maintained, and this can be very costly; (ii) general purpose ontologies, such as {\it WordNet}\footnote{http://wordnet.princeton.edu}, contain a limited number of relations between concepts, mainly hierarchical relations (\textit{is\_a} and \textit{part\_of}) and a few non-hierarchical or thematic ones \cite{Kacmajor20};
(iii) domain specific ontologies are available only in a few cases.
Furthermore, when knowledge-based techniques are applied, they often exploit taxonomies and, therefore, the focus is limited to the notion of semantic {\it similarity} \cite{Chandra2021},  which is a particular case of semantic relatedness.

With the advent of  \emph{Linked Data}\footnote{http://www.w3.org/TR/2015/REC-ldp-20150226/}, a new frontier appeared, enabling the generation of large knowledge graphs (or semantic networks), such as \emph{DBpedia}\footnote{https://www.dbpedia.org/}, which is the result of an ongoing project aiming at producing structured content from \emph{Wikipedia}\footnote{https://www.wikipedia.org/}.
Since the number of published Linked Data datasets is growing, the interest in exploiting knowledge graphs for 
knowledge-based applications is increasing as well \cite{HARARI2022}.

Knowledge graphs are fundamental in several research areas, such as for instance pattern mining \cite{Fournier2020}, social network analysis \cite{Dougnon2016}, etc.. In this work the focus is on semantic relatedness, which is a key feature 
in Word Disambiguation \cite{Wang2020}, Entity Linking \cite{Oliveira2021}, Recommendation Systems \cite{NikzadK-Khasmakhi2019}, Data Mining \cite{Ristoski2019}, Information Retrieval \cite{Yang2020}, Question Answering \cite{Zhang2021},  etc. 
Semantic relatedness 
captures two main key dimensions: taxonomic and non-taxonomic relations \cite{Kacmajor20}. In general, taxonomic relatedness concerns semantic similarity, that has been extensively analyzed in the literature \cite{Chandra2021}, whereas non-taxonomic relations
are fundamental in the evaluation of the more general notion of semantic relatedness.
With this regard, to our knowledge, one of the most recent and relevant surveys on semantic relatedness is \cite{Taieb2019ASO}, which defines the guidelines to select, develop, and evaluate semantic relatedness measures, although  a benchmarking of the existing methods is not provided. To date, computing semantic relatedness is both conceptually and practically an open challenge \cite{Taieb2019ASO}.

Among the existing approaches, this paper focuses on the methods for evaluating semantic relatedness of Web resources in DBpedia. As known, DBpedia is a continuously evolving knowledge graph, and the methods defined in the literature provide their own experimental results, whose correlations with human judgment are often non-comparable because they have been evaluated in different time periods. Therefore, an experiment on the same DBpedia release and against the same datasets is missing.  For this reason, in this paper we selected and compared 10 representative proposals, by benchmarking them all at once  against 14 {\it golden}  datasets addressed in the literature, by using the same  DBpedia  release. These methods have  been compared by providing first an informal description and some intuitive examples about them. Successively, they have been formally recalled and a  technical  running example has been given  in order to highlight the key aspects characterizing the different approaches. To the best of our knowledge this work provides the first comparative experiment in this direction.

The paper is organized as follows.  In Section \ref{relatedwork} the related work is given, where semantic relatedness has been analyzed by focusing first on semantic similarity and, then, on methods relying on  WordNet, Wikipedia, and Machine Learning techniques.  Section \ref{sec:semantic_relatedness} introduces the notion of semantic relatedness, and  a classification of  semantic relations  in line with \cite{Kacmajor20}.  Section \ref{sec:rdf_linked_data} provides an introduction about  RDF\footnote{http://www.w3.org/TR/2014/REC-rdf11-concepts-20140225/}, the W3C\footnote{The World Wide Web Consortium. https://www.w3.org/} specifications for conceptual description and modeling of information, and DBpedia.
In Section \ref{sec:methods_for_semantic_relatedness}  the 10 methods  are informally presented, by providing simple  examples in order to highlight their differences and commonalities. In particular, they have been organized according to 
 three groups, namely {\it adjacent resources}, {\it triple patterns}, and {\it triple weights} based methods.
In Section \ref{sec:experiment} the experimentation is presented, with the evaluation of the results and a discussion about them. Section \ref{sec:conclusion} concludes. Finally, in the Appendix, the 10 methods are formally recalled, and a technical running example is provided in order to better illustrate the different approaches.

\section{Related Work} \label{relatedwork}
Semantic relatedness is a fundamental research topic not only in computer science \cite{Taieb2019ASO}, but also in other disciplines, such as  economic and social sciences \cite{Tacchella2021}, however it is still a challenge.
In the literature there is a significant amount of works addressing semantic similarity which, as mentioned in the Introduction, is a particular case of semantic relatedness \cite{Chandra2021, Kacmajor20}, and has been investigated also by the authors within Formal Concept Analysis \cite{FORMICA2006, F19}, and  Semantic Web search  \cite{FMPT2008, FMPT13}. With the advent of Wikipedia, i.e., the {\it Web of Documents} and, successively, Linked Data (and, in particular, DBpedia), i.e, the {\it Web of Data},  further approaches for evaluating semantic similarity have been proposed, such as for instance \cite{HUSSAIN2020}. 
In particular, this work relies on the semi-structured taxonomy called Wikipedia
Category Graph (WCG), and proposes a method to measure the semantic similarity between Wikipedia concepts. 
In order to improve the   efficiency of  semantic similarity methods, other approaches exploit the advantages of combining Wikipedia with WordNet, as for instance \cite{Li2020}.
However, the focus of all the aforementioned papers is on semantic similarity rather than relatedness. 
It is worth mentioning  that, among the various approaches,  in \cite{Piao2015}    the authors propose the {\it Resim} (Resource Similarity) measure for evaluating  semantic similarity of DBpedia resources and, successively,  in \cite{Piao2016}   they address the more general problem of relatedness, and propose an approach that is one of the  10 methods selected for the experimentation of this work (see Section \ref{Methods_patterns} and also Section \ref{PiaoBreslin2016_2}).
Note that similarity is fundamental  also in clustering \cite{Carpineto2009}, aimed at  partitioning 
 data into similar groups, which has been extensively investigated in the literature. For example in ontology matching, 
in order to deal with large scale ontologies,  it is necessary to decompose the
huge number of instances into a small number of clusters.
Clustering is addressed for ontology matching for instance in  \cite{Djenouri2021}. In particular, the proposed approach aims at extracting  sets
of instances from a given ontology and grouping them into  subsets in order to evaluate the
 common instances between different ontologies. 
Clustering on semantic spaces is also used for the summarization of image
collections and self-supervision, as for instance in \cite{Sharma2022}.

In the following, we restrict our attention to the literature addressing semantic relatedness, that is the focus of this paper, rather than the more specific notion of semantic similarity. 
%In particular, the approaches recalled below have been organized according to the exploited knowledge graph. 
Note that  semantic relatedness measures defined for specific domains and experimented on specific datasets (as for instance in biomedicine \cite{Kirkpatrick2022}) have not been addressed in this paper
because experiments show that some of them, that are effective for a specific task or an application area, do not perform well in general \cite{Taieb2019ASO}.

Below, the approaches from the literature have been organized according to three main groups, relying on WordNet, Wikipedia, and Machine Learning techniques, respectively. Before introducing them, it is worth mentioning two recent methods presented in \cite{Natarajan2022} and \cite{Ahmed2021}, respectively. The former proposes a new measure within recommender systems
which evaluates the closeness of items across domains
in order to generate relevant recommendations for new
users in the target domain. Essentially, such a measure is  based on the total number of web pages where the  words describing the compared items occur together. According to the latter, semantic relatedness is evaluated for unstructured data by relying on fuzzy vectors and by using different semantic relatedness techniques. However, both these approaches are not knowledge-based and for this reason they have not been considered in our experiment.

%\begin{itemize}

\vspace{0.15 cm}

%\item{ \textit{WordNet} -
\noindent \textbf{WordNet}. WordNet  can be considered as a relatively simple knowledge graph designed to semantically model  the English  lexicon. It  contains mainly  taxonomic relations ({\it is\_a}), and  part-whole (\textit{part\_of}) relations, whereas a few thematic relations are present  (see the next section where  semantic relations have been recalled).
In the literature, several approaches for computing semantic relatedness have been proposed by leveraging  WordNet knowledge graph, as for instance   \cite{TAGARELLI2013, AlMousa2021, Yang2020}.
In particular, in \cite{TAGARELLI2013}, the problem of measuring semantic relatedness in labeled tree data is addressed by leveraging the {\it is\_a} and  \textit{part\_of} hierarchies of WordNet.
In \cite{AlMousa2021}, the authors state  that  the majority of the proposed methods rely on the {\it is\_a} relation, and 
%emphasize the importance of non-taxonomic information. For this reason, they  
introduce a new
approach to measure semantic relatedness between concepts based on weighted paths defined by non-taxonomic relations in  WordNet.
In \cite{Yang2020}, semantic relatedness is evaluated by following different strategies in order to improve computation performances, by combining WordNet with word embedding methods.
Furthermore, it is worth recalling that in \cite{Torres2018} the authors define an algorithm for semantic relatedness relying on random  walks, i.e.,  generalizations of paths where cycles are allowed, that has been evaluated on WordNet. However, as mentioned by the same authors, WordNet is relatively small, and   an evaluation of the performances of their proposal on larger knowledge graphs, such as DBpedia, is missing.
In this paper the  approaches designed, and somehow limited,  to evaluate semantic relatedness in  WordNet have not been addressed since here the focus is on the methods that have been experimented on  larger knowledge graphs, i.e., that  contain a more heterogeneous set of relations.

\vspace{0.15 cm}
%\item{\textit{Wikipedia} -
\noindent \textbf{Wikipedia}. Wikipedia is a free, multilingual, online encyclopedia and, to date, the English version edition is composed of more than 6 million articles written and maintained by a community of volunteers. It can be seen as a large
corpus where entities are described by natural language, and therefore it contains a huge amount of unstructured information.
For this reason, 
 methods for evaluating relatedness between Wikipedia entities require a significant pre-processing effort in order to extract structured information  from the natural language descriptions. 
With this regard, the WCG mentioned above is a hybrid structure, i.e, it is not a rigorous \textit{is\_a} taxonomy  that has been conceived in order to facilitate the management of Wikipedia articles. Therefore, an interesting research direction concerns the analysis of  the trade-off between the expressivity of natural language queries and their ``usability'' over Linked Data.
For instance, in \cite{FREITAS2013} the TREO system  has been presented where Linked Data are queried by combining entity search, the TF-IDF method \cite{aizawa2003} for link weighting, spreading activation models, and 
\textit{WLM}  (one of the methods selected for comparison in this work, see  Subsection \ref{MethodsAdjacent} and \ref{sec:WLM_2}).
In the same direction,  in \cite{Vakulenko2019},
knowledge graphs are used in combination with text similarity techniques for improving the efficiency of complex question answering.

In this paper  the proposals focusing on the relatedness of entities in Wikipedia have not been addressed  because, as mentioned in the Introduction, in order to analyze and extract keywords from Wikipedia documents, they rely on corpus-based approaches and, therefore, on natural language processing techniques that   go beyond the scope of this work. However, this does not hold for \textit{WLM} that is a pure knowledge base approach and, as shown in the next sections, it has been included in the paper by using its RDF graph formulation.
%}

\vspace{0.15 cm}
\noindent \textbf{Machine Learning}. Recently, some works have proposed to apply Machine Learning techniques to compute semantic relatedness, by encoding the available knowledge as numerical vectors.
When the available knowledge is in the form of textual documents, this step is referred to as \textit{word embedding}, whereas, when dealing with graph-shaped knowledge, as \textit{graph embedding}.
Examples about word embedding for semantic relatedness are proposed in \cite{li2021}, \cite{sarwar2022}, and \cite{Zhou2022}. In particular, \cite{li2021} aims at
achieving a better accuracy on the semantic relatedness of both isolated words and words in contexts. 
In  \cite{sarwar2022},  word embedding is applied to represent keyphrases in a corpus of textual documents in order to find similar news articles. 
In \cite{Zhou2022}, a semantic relatedness graph is constructed in order to detect sentiment polarities in a long sentence towards multiple aspect categories.
However, the first two proposals are corpus-based, whereas the third one is an hybrid method combining semantic similarity on a taxonomy and a distributional approach over a corpus of documents. Therefore, these three methods have not been addressed in our experiment.
Concerning graph embedding, in \cite{Ristoski2019}
 the RDF2Vec approach for evaluating semantic relatedness of Linked Data has been proposed by relying on Neural Network models. The mentioned paper is, to the best of our knowledge, the first proposal that leverages the graph structure using neural language modeling for the purpose of entity relatedness and similarity.  However, the computation of embedding is time-consuming \cite{Chatzakis2021}, and the experiments, even on small RDF datasets, do not terminate in a reasonable number of days or run out of memory. Along this research direction computational efficiency is still an open problem \cite{Chandra2021} that goes beyond the scope of this paper. 
 Finally, it is worth mentioning \cite{Huang2021}, which applies Machine Learning techniques to images representing  words in order to investigate the cognitive mechanism underlying semantic relatedness by using  deep convolutional neural networks. However, also this approach does not involve graph-based knowledge that is the focus of our work.

\section{Semantic Relatedness}
\label{sec:semantic_relatedness}
\begin{figure}[ht]
    \centering
    \includegraphics[width=0.90\textwidth]{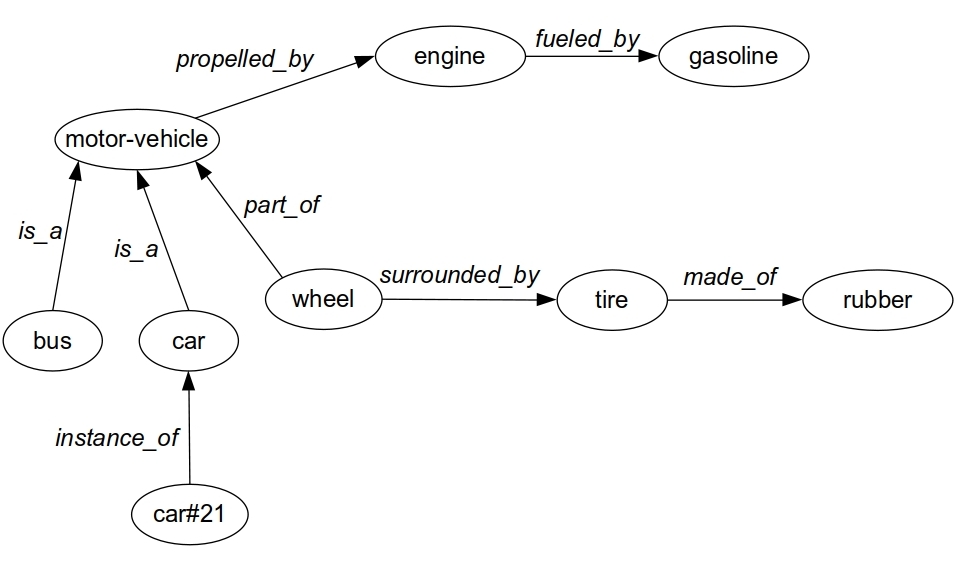}
    \caption{A simple knowledge graph}
    \label{fig:simple_sem_net}
\end{figure}

The Merriam-Webster dictionary defines the term \textit{related} as: ``\textit{connected by reason of an established or discoverable relation}''. According to this definition,  \textit{established} can be intrerpreted as \textit{explicit}, and  \textit{discoverable} as \textit{implicit}. Let us consider a knowledge graph where nodes represent entities (concepts or real world objects) and arcs stand for  relations between them. An explicit relation between  entities can be seen as an existing edge between the corresponding nodes, whereas an implicit relation can be identified by a chain of edges connecting the related nodes.
For instance, Figure \ref{fig:simple_sem_net} shows a simple semantic network where the node \textit{car} is related to the node \textit{motor-vehicle} by means of the explicit (or established)
\textit{is\_a} relation, i.e., \textit{car} $\rightarrow$  \textit{is\_a} $\rightarrow$ \textit{motor-vehicle}, whereas \textit{motor-vehicle} is related to \textit{gasoline} by means of an implicit (or discoverable) relation corresponding to the path \textit{motor-vehicle} $\rightarrow$  \textit{propelled\_by} $\rightarrow$  \textit{engine} $\rightarrow$  \textit{fueled\_by} $\rightarrow$  \textit{gasoline}.

In general, a  relation is semantic when it is based on the meaning of the involved words. For example, \textit{tire} $\rightarrow$ \textit{made\_of} $\rightarrow$ \textit{rubber} represents a semantic relation because rubber is the material a tire is made of. On the contrary, for instance, \textit{car} $\rightarrow$ \textit{rhymes\_with} $\rightarrow$ \textit{star} is not a semantic relation because it holds due to the assonance between the words. According to \cite{Kacmajor20}, semantic relations can be organized according to the following classification:

\begin{itemize}
\item Taxonomic relation
\begin{itemize}
\item Specialization relation (\textit{is\_a})
\end{itemize}
\item Non-taxonomic relation
\begin{itemize}
\item Part-whole relation (\textit{part\_of})
\item Idiosyncratic relation
\item Thematic relation
\item Instance relation
\item ...
\end{itemize}
\end{itemize}

\noindent where, with respect to the classification presented in \cite{Kacmajor20}, the part-whole relations have been highlighted among the non-taxonomic ones according to \cite{AlMousa2021}. 
In general, in the literature, a taxonomic relation refers to the notion of specialization, i.e. the well-known {\it is\_a} relation that involves concepts with common features and functions. In particular, this relation  allows the identification of concepts that are semantically similar, as for instance {\it knife} and {\it fork} that are both \textit{cutlery} \cite{Lin1998}.

Within the non-taxonomic ones, that concern concepts that co-occur in any sort of context, an important role is represented by the part-whole, or meronymic, relations \cite{AlMousa2021},  i.e., semantic relations between a meronym  denoting a part and a holonym denoting a whole. These can be further distinguished according to different types of meronymy, such as (i) component-integral object, as for instance {\it pedal} and {\it bike}, (ii) member-collection, as for instance  {\it ship} and {\it fleet}, (iii) portion-mass, as for instance {\it slice} and {\it pie}, etc. \cite{WINSTON87}.
Non-taxonomic relatedness is often characterised in terms of free associations relying on the probability for one concept to evoke another concept \cite{Nelson2004}. With this regard, the idiosyncratic relations originate from subjective  perceptions associated with  autobiographic memories, as for instance {\it coffee} and {\it beard}, that can be related for someone because they are  often associated with morning activities, but this of course may not be true for someone else. 
Thematic relations involve concepts performing complementary roles in a given context, as for instance {\it river} and {\it bridge}. 
Note that pairs of concepts taxonomically related can also be thematically related, as for instance {\it doctor} and {\it nurse} that are similar because they are both health
professionals, but they are also thematically related, because they perform
complementary roles, for example during surgery \cite{Kacmajor20}.

In a knowledge graph different kinds of semantic relations coexist, as shown for instance by the graph of Figure \ref{fig:simple_sem_net}.
In particular, the nodes \textit{car} and \textit{bus} are both related by \textit{is\_a} arcs to the more general concept \textit{motor-vehicle} (\textit{car} $ \rightarrow $ \textit{is\_a} $\rightarrow$  \textit{motor-vehicle}, and \textit{bus} $\rightarrow$  \textit{is\_a} $\rightarrow$  \textit{motor-vehicle}). For this reason, \textit{car} and \textit{bus} are sibling concepts  sharing the meaning  of their parent \textit{motor-vehicle} and, therefore, are similar \cite{FT2021, Lin1998}.
Furthermore, \textit{wheel} $\rightarrow$ \textit{part\_of} $\rightarrow$  \textit{motor-vehicle} represents an example of meronymy, where \textit{wheel} is the part and  \textit{motor-vehicle} is the whole. 
\textit{Motor-vehicle} $\rightarrow$ \textit{propelled\_by} $\rightarrow$ \textit{engine}, and \textit{engine} $\rightarrow$ \textit{fueled\_by} $\rightarrow$ \textit{gasoline} represent examples of thematic relatedness, since these relations pertain to a certain theme (i.e., the automotive).
Finally, \textit{car\#21 $\rightarrow$ instance\_of $\rightarrow$ car} is an example of an instance relation, since it involves a real world entity and its type. 

%\begin{figure}
%\centering
%\subfloat[][\label{fig:simple_sem_net}] 
%{\includegraphics[width=.48\textwidth]{simple_sem_net.jpg}} \quad
%\subfloat[][\label{fig:Rome}]
%{\includegraphics[width=.48\textwidth]{Rome.jpg}}\hfill \\
%\caption{(a) A simple knowledge graph and (b) an example of RDF graph}
%\end{figure}

According to \cite{Taieb2019ASO}, in the following ``we use the term semantic relatedness in a general sense, i.e. how much connection humans perceive between two concepts''. Hence, in this paper all kinds of semantic relations have been addressed, without making any assumption about the causes of a given perception.

\section{Resource Description Framework (RDF) and DBpedia}
\label{sec:rdf_linked_data}
The Resource Description Framework (RDF) is a family of specifications designed as a standard model for data interchange on the Web. In particular, RDF is used for the conceptual description or modeling of information of Web  resources, each identified by a Uniform Resource Identifier (URI). RDF is based upon the idea of making statements about resources by means of expressions in the form of triples following  a \textit{subject}$-$\textit{predicate}$-$\textit{object} pattern. The subject denotes the resource that is being described, and the predicate expresses a relation between the subject and the object, which can be a resource or a literal (e.g., a string, a number).

Let $R=\{r_1, r_2, ..., r_n\}$ be a finite set of URIs each representing a resource, and $L=\{l_1, l_2, ..., l_m\}$ a finite set of literals, an RDF triple (or statement) has the form:

\begin{center}
%$<$\textit{s}, \textit{p}, \textit{o}$>$
$\langle s, p, o\rangle$,
\end{center}
where
\textit{s} $\in R$ is the subject, \textit{p} $\in R$ is the predicate, and \textit{o} $\in R \cup L$ is the object.

\begin{figure}
    \centering
    \includegraphics[width=0.90\textwidth]{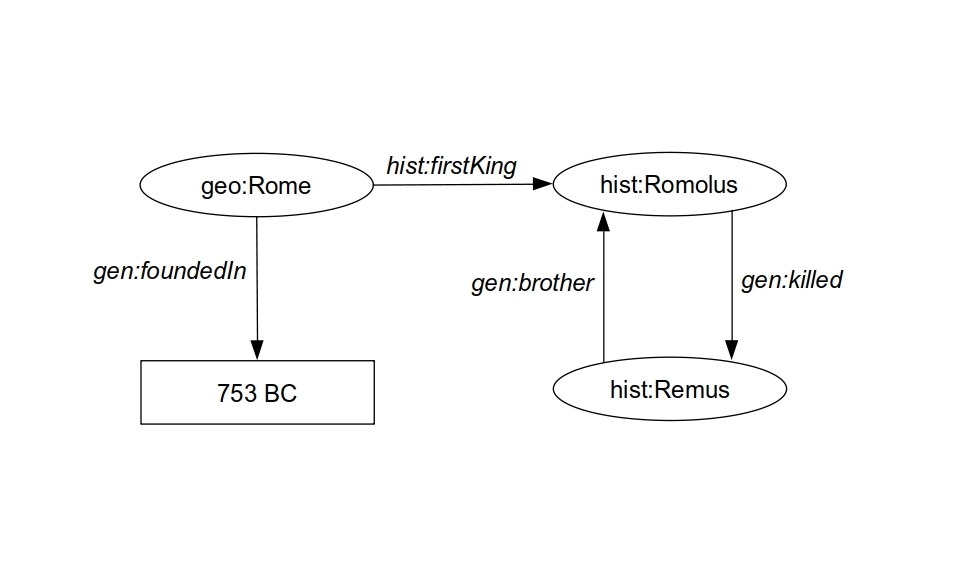}
    \caption{An example of RDF graph}
    \label{fig:simple_rdf_graph}
\end{figure}

An RDF graph $\mathcal{G}$ is a set of RDF triples,  where subjects and objects are nodes, and predicates are directed arcs (also called links, edges or arrows).   
For instance, in Figure \ref{fig:simple_rdf_graph}  an RDF graph is shown. 
The triples $\langle\textit{geo:Rome}, \textit{hist:firstKing}, $ $ \textit{hist:Romolus}\rangle$,  and $\langle \textit{geo:Rome}, \textit{gen:foundedIn}, \textit{753 BC}\rangle$ express  that the first king of Rome was Romolus, and the city of Rome was founded in 753 before Christ, respectively. In the proposed examples, \textit{geo:}, \textit{hist:}, and \textit{gen:} are prefixes for namespaces\footnote{A namespace is a collection of terms  that allows them to be  uniquely identified.} that are assumed to contain geographical, historical and generic terms, respectively.

In the following, a triple \textit{t} $=$ $\langle s, p, o\rangle$ represents a directed link, labelled as \textit{p}, from the resource \textit{s} to the resource \textit{o}. The predicate \textit{p} is said to be \textit{outgoing} from \textit{s} and \textit{incoming} to  \textit{o}.
%A \textit{walk W} of length \textit{n} in a RDF graph is an ordered list of $n$ triples [$t_1$, $t_2$, ... $t_n$], where $o_i$, the object of the triple $t_i$, coincides with $s_{i+1}$, the subject of the triple $t_{i+1}$, for $1\leq i \leq n-1$. 

Given two resources $r_a$ and $r_b$, a \textit{directed path} $P$ of length \textit{n} from $r_a$ to $r_b$ is a list of $n$ triples [$t_1$, $t_2$, ... $t_n$] where $r_a$ coincides with $s_1$, the subject of the triple $t_1$,  $r_b$ coincides with $o_n$, the object of the triple $t_n$, and $o_i$, the object of the triple $t_i$, coincides with $s_{i+1}$, the subject of the triple $t_{i+1}$, for $1\leq i \leq n-1$.
%In the following, we assume that paths are acyclic, **** that is, $o_i \neq s_1$, and $o_i \neq o_j$ for any $i$ and $j$ such that  $1\leq i, j \leq n$, and  $i \neq j$.
For instance, the sequence of triples [$\langle$\textit{geo:Rome}, \textit{hist:firstKing}, \textit{hist:Romulus}$\rangle$, $\langle$\textit{hist:Romulus}, \textit{gen:killed}, \textit{hist:Remus}$\rangle$] represents a directed path of length $2$, from the resource \textit{geo:Rome} to the resource \textit{hist:Remus}.  

An \textit{undirected path} connecting two resources is a path in which the predicates can be traversed in both directions, i.e., they represent undirected links. For instance,
the list of triples: [$\langle$\textit{geo:Rome}, \textit{hist:firstKing}, \textit{hist:Romulus}$\rangle$, $\langle$\textit{hist:Remus}, \textit{gen:brother}, \textit{hist:Romulus}$\rangle$] represents an undirected path  connecting the resources \textit{geo:Rome} and \textit{hist:Remus}, where the predicate \textit{gen:brother} is traversed from the object \textit{hist:Romulus} to the subject \textit{hist:Remus}.

Note that, although an RDF graph can be cyclic, we consider only acyclic paths, i.e., paths where there are no repetitions of nodes, therefore {\it walks} \cite{Torres2018} are not allowed.

%The \textit{distance} between two resources $r_a$ and $r_b$ is defined as the length of the shortest path connecting $r_a$ and $r_b$, which can change whether directed or undirected paths are considered.

RDF identifies a vocabulary for making assertions on resources. For instance, the predicate \textit{rdf:type}\footnote{\textit{rdf:} is the prefix for the RDF namespace.} is used to state that a resource is an instance of another resource.  
Furthermore, RDF is used for defining further vocabularies. For instance, the RDF Schema (RDFS)\footnote{https://www.w3.org/TR/rdf-schema/}, and the Web Ontology Language (OWL)\footnote{https://www.w3.org/OWL/} are two RDF vocabularies. The former  introduces, among the others, the resource \textit{rdfs:Class} and the predicate \textit{rdfs:subClassOf}, which can be used for defining taxonomies, whereas the latter, which is built on top of RDFS, defines more sophisticated constructs for representing computational ontologies. 
Finally, SPARQL\footnote{http://www.w3.org/TR/2013/REC-sparql11-query-20130321/}, which is a recursive acronym for SPARQL Protocol and RDF Query Language, is an RDF query language based on a SELECT-FROM-WHERE syntax, where variables begin with a question mark (?). For instance, in SPARQL,  $\langle r, ?x, ?y \rangle$ represents any triple having the resource $r$ as subject.

%The definitions of walk and path given above, assume that arcs representing predicates must be traversed according to their directions. However, when dealing with knowledge graphs, it is feasible to consider that predicates, and consequently their directions, are the results of implementation choices. In fact, it can be assumed that for each predicate, e.g., \textit{gen:killed}, an inverse predicate, e.g., \textit{gen:killedBy}, could be defined. When this assumption is assumed, predicates are considered bidirectional (or undirected), and can be traversed in both directions. 
%For instance, p = [$<$\textit{geo:Rome}, \textit{hist:firstKing}, \textit{hist:Romulus}$>$, $<$\textit{hist:Remus}, \textit{gen:brother}, \textit{hist:Remulus$>$}] is an example of path where predicates are assumed to be bidirectional. 

DBpedia is a very huge RDF knowledge graph, and is the result of an ongoing process aimed at semi-automatically extracting 
information from Wikipedia,  in order to represent it in RDF. 
 Therefore, for each Wikipedia article a corresponding RDF resource exists.  Note that, a significant part of DBpedia comes from the information in the \textit{infoboxes} of the Wikipedia articles\footnote{An infobox is a panel that summarizes the key features of the Wikipedia article.}.  The infobox contains data  represented as property-value pairs provided by articles' editors that are, in general,  an excerpt of the relevant information of  a DBpedia resource.

\section{Methods for Computing Semantic Relatedness}
\label{sec:methods_for_semantic_relatedness}
In this section, we recall 10 methods for computing semantic relatedness in RDF graphs that have been selected from the literature. These methods have been chosen on the basis of the following criteria: 
\begin{itemize}
    \item They are pure knowledge-based approaches. For this reason, we did not consider any corpus-based and hybrid method and, in general, any method addressing the use of natural language processing techniques.
    \item They can be applied to any RDF graph, without making any assumption about  the   types of nodes and predicates. Therefore, we did not take into consideration the methods  defined for specific knowledge graphs, as for instance  WordNet. 
\end{itemize}
\noindent Furthermore, these methods have been identified by:
\begin{itemize}
    \item Searching on Google Scholar, in September 2021,  by using the following keywords: [``semantic relatedness'' ``rdf''], and by considering only the articles published    since 2015.
    \item Analyzing the first 40 pages of results (400 in total), and by selecting the articles about semantic relatedness methods, according to the above criteria. This search allowed us to identify 4 methods, namely, \textit{Linked Data Semantic Distance with Global Normalization} (here referred to as \textit{LDSDGN}) \cite{Piao2016}, \textit{Propagated Linked Data Semantic Distance} (\textit{PLDSD}) \cite{Alfarhood2017},  \textit{Exclusivity-based measure} (here referred to as \textit{ExclM}) \cite{Hulpus2015}, and \textit{ASRMP$_m$} \cite{ElVaigh2020}.
    \item Selecting 6 further methods on the basis of the bibliographic references of the papers about the 4 methods above. These methods are: \textit{Wikipedia Link-based Measure} (\textit{WLM}) \cite{Witten2008},  \textit{Linked Open Data Description Overlap} (\textit{LODDO}) \cite{Zhou2012}, \textit{Linked Data Semantic Distance} (\textit{LDSD}) \cite{Passant2010}, \textit{IC-based measure} (here referred to as \textit{ICM}) \cite{Schuhmacher2014}, \textit{REWOrD} \cite{Pirro2012}, and \textit{Proximity-based Method} (here referred to as \textit{ProxM}) \cite{Leal2013}. 
\end{itemize}

In this paper the selected methods have been classified according to the following three groups:

\begin{enumerate}
\item Methods based on adjacent resources.
\item Methods based on triple patterns.
\item Methods based on  triple weights.
\end{enumerate}

\begin{figure}[ht]
    \centering
    \includegraphics[width=0.65\textwidth]{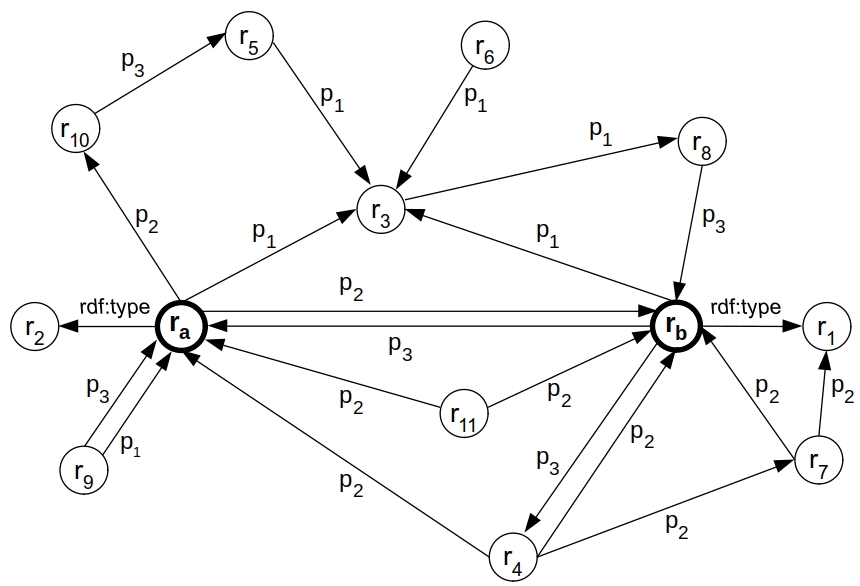}
    \caption{The running example graph $\mathcal{G}$}
    \label{fig:master}
\end{figure}

Some of these  methods have been originally conceived for computing a  distance. Hence, in these cases we adopted   the corresponding relatedness formulation, based on the assumption that the shorter the distance the greater the relatedness. 
In the Appendix the 10 methods are formally recalled and, in order to achieve a more effective comparison among them, a running example is used, based on the graph $\mathcal{G}$ shown in Figure \ref{fig:master}. Such a graph contains 13 nodes (resources), linked with directed edges labeled with 4 possible predicates, namely, $p_1$, $p_2$, $p_3$, and \textit{rdf:type}. Among the 13 resources,  $r_a$ and $r_b$ are the ones whose relatedness will be  addressed when describing each method.

 Below the 10 methods are informally summarized, and their main characteristics are recalled, but  
 readers interested in the formal aspects can refer to the Appendix.

\subsection{Methods based on adjacent resources}
\label{MethodsAdjacent}
In this subsection the methods belonging to the first group are described, that are based on resources' adjacent nodes, i.e., nodes that are linked to the compared resources via paths of length 1 in the knowledge graph.
They are \textit{Wikipedia Link-based Measure} (\textit{WLM}), and  \textit{Linked Open Data Description Overlap} (\textit{LODDO}).\\
%The focus on local knowledge,  with respect to the compared resources, makes these methods very efficient from a computational complexity point of view.   

\noindent \textbf{Wikipedia Link-based Measure (\textit{WLM}).} \label{sec:WLM}
In \cite{Witten2008}, the \textit{Wikipedia Link-based Measure} (\textit{WLM}) is presented.
This  measure,  which originally exploits the hyperlinks within Wikipedia articles, derives from the well-known {\it Normalized Google Distance} ($NGD$) \cite{Cilibrasi2007}, which is based on the assumption that, given two terms, the more pages contain them the more related  they are.
In this paper, the  \textit{WLM} approach  is recalled by applying it to an RDF graph. In particular, rather than considering the Wikipedia articles or Google pages containing a given term, the set of RDF  triples of the graph whose objects correspond to such a term are addressed.
Then, the set of the resources that are the subjects of such triples are considered.

For instance consider the graph of Figure \ref{fig:master} and the resources $r_a$ and $r_b$. In order to evaluate their relatedness, two sets of RDF triples have to be addressed, one for each of the compared resources. For example in the case of $r_a$, this set is given by the triples of the graph whose objects correspond to such a resource, i.e., \{$\langle r_4, p_2, r_a\rangle$,  $\langle r_9, p_3, r_a\rangle$, $\langle r_9, p_1, r_a\rangle$, $\langle r_{11}, p_2, r_a\rangle$, $\langle r_b, p_3, r_a\rangle$\}. Therefore, regarding $r_a$,  the set of resources  \{$r_4$, $r_9$, $r_{11}$, $r_b$\} will be considered, i.e., all the resources with incoming predicates to $r_a$.\\

\noindent \textbf{Linked Open Data Description Overlap (\textit{LODDO}).}
The \textit{Linked Open Data Description Overlap} (\textit{LODDO}) method \cite{Zhou2012} is based on the notion of 
\textit{description} of a resource, which is the set of the resources linked to it, either via an incoming, or an outgoing predicate, excluding \textit{rdf:type}, and including the resource itself. In other words, a resource \textit{r$_i$}, different from $r$, belongs to the description of  \textit{r} if it participates in a triple with \textit{r}, either as  subject or object. For instance, in the graph of Figure \ref{fig:master}, the description of $r_a$, say $D(r_a)$, is given by the following set \{$r_a$, $r_b$, $r_3$, $r_4$, $r_9$, $r_{10}$, $r_{11}$\}.
 The approach proposes two strategies, namely \textit{LODOverlap} and \textit{LODJaccard}, sharing the rationale that the more the descriptions of two resources have in common, the greater their relatedness.
According to \cite{Zhou2012}, the \textit{LODOverlap} strategy performs better than the \textit{LODJaccard} one, and this is the strategy that has been considered in our experimentation.

\subsection{Methods based on triple patterns}
\label{Methods_patterns}
These methods are based on the identification of \textit{path patterns} in the knowledge graph, i.e., paths satisfying specific conditions with respect to the compared resources.
Note that these methods  represent distances and, as mentioned above, the shorter the distance the greater the relatedness.
They are \textit{Linked Data Semantic Distance} (\textit{LDSD}), \textit{LDSD with Global Normalization} (\textit{LDSDGN}), and \textit{Propagated Linked Data Semantic Distance} (\textit{PLDSD}). 
\\

\noindent \textbf{Linked Data Semantic Distance (\textit{LDSD}).\label{sec:Passant}}
In \cite{Passant2010}, Passant proposes a theoretical definition of Linked Data and shows how relatedness between resources can be evaluated by using the semantic distance measure introduced by Rada \cite{Rada1989}. With respect to the traditional approach of Rada which focuses  on hierarchical relations,  the proposed distance takes into account  any
kind of links. In particular, a family of measures for semantic distance has been defined, named \textit{Linked Data Semantic Distance} (\textit{LDSD}). 
In the Appendix, the three measures belonging to this family are recalled.
The first one focuses on direct links ($LDSD_{dw}$), the second one on indirect links ($LDSD_{iw}$), and the third one on a combination of both direct and indirect links between the compared resources ($LDSD_{cw}$).
As mentioned by the author in \cite{Passant2010}, among these three measures, the best one is $LDSD_{cw}$, which has been considered in our experiment. In essence, it addresses direct paths, and indirect paths  adhering to  a specific pattern, that is, two links labeled with the same predicate both outgoing from (incoming to) a third resource and incoming to (outgoing from) the compared resources.

For instance, consider the graph in Figure \ref{fig:master}, in comparing the resources $r_a$ and $r_b$, besides the  direct paths, which are [$\langle r_a, p_2, r_b\rangle$] and [$\langle r_b, p_3, r_a\rangle$], the following indirect paths contribute: (i) [$\langle r_4, p_2, r_a\rangle$, $\langle r_4, p_2, r_b\rangle$], whose  links are labeled with $p_2$  and are outgoing from the third resource $r_4$,
(ii) [$\langle r_{11}, p_2, r_a\rangle$, $\langle r_{11}, p_2, r_b\rangle$],  where  links are labeled with $p_2$ and are outgoing from the third resource $r_{11}$, and (iii) [$\langle r_a, p_1, r_3\rangle$, $\langle r_b, p_1, r_3\rangle$], whose  links are labeled with $p_1$ and are incoming to the third resource $r_3$. 
Note that, as clarified by the formalization of the measures provided in the Appendix, both the $LDSD_{iw}$ and $LDSD_{cw}$ are not symmetric, i.e., they are independent of the order of the compared resources.\\

\noindent \textbf{LDSD with Global Normalization (\textit{LDSDGN}).} \label{PiaoBreslin2016}
The measures  presented in \cite{Piao2016}, in this paper referred to as \textit{LDSD with Global Normalization} (\textit{LDSDGN}), represent an evolution of the approach proposed by Passant \cite{Passant2010}. 
In \cite{Piao2016} the authors present three strategies, namely $LDSD_\alpha$,  $LDSD_\beta$,  and $LDSD_\gamma$. In the first case, they assume that resources are more related if there is a great number of them linked to the compared resources via a given predicate. In the second strategy further assumptions are considered in order to achieve symmetry. In the third case, the contribution of the indirect paths is normalized with respect to the global number of occurrences of the corresponding patterns in the whole graph. According to the authors, $LDSD_\gamma$ is the best strategy, and it has been selected for the experiment of this paper. 

In the case of the graph of Figure \ref{fig:master}, on the basis of  the third strategy, the contribution of the path [$\langle r_a, p_1, r_3\rangle$, $\langle r_b, p_1, r_3\rangle$], linking $r_a$ and $r_b$ by means of the predicate $p_1$, is normalized by taking into account the cardinality of the set of paths having a similar pattern. In this case, they are two links labeled with $p_1$ that are incoming to the same resource, which by chance is always $r_3$. In particular, this set is the following: 

\noindent \{[$\langle r_a, p_1, r_3\rangle$, $\langle r_5, p_1, r_3\rangle$], [$\langle r_a, p_1, r_3\rangle$, $\langle r_6, p_1, r_3\rangle$], [$\langle r_a, p_1, r_3\rangle$, $\langle r_b, p_1, r_3\rangle$], 

[$\langle r_b, p_1, r_3\rangle$, $\langle r_5, p_1, r_3\rangle$], [$\langle r_b, p_1, r_3\rangle$, $\langle r_6, p_1, r_3\rangle$], [$\langle r_5, p_1, r_3\rangle$, $\langle r_6, p_1, r_3\rangle$]\}. 
\\

\noindent \textbf{Propagated Linked Data Semantic Distance (\textit{PLDSD}).}
The measure proposed in \cite{Alfarhood2017} originates from the need to overcome some drawbacks of the families of methods
 illustrated above. Indeed, according to them, semantic distance is evaluated by focusing on the resources that are either directly or indirectly linked by means of a single intermediate resource.
Therefore, all the resources belonging to longer paths are not involved in the relatedness evaluation. 
For this reason, in the aforementioned paper Alfarhood et al.  present a measure, named \textit{Propagated Linked Data Semantic Distance} (\textit{PLDSD}), that extends the previous approaches in this direction.
%In \cite{Alfarhood2017}, the authors consider the approach proposed by Passant limited by the fact that it considers only indirect paths between two compared resources with at most one intermediate node. 
In particular, in the proposed method, all the paths between the compared resources, up to a given length $h$, are taken into account, and for each pair of adjacent resources in these paths the original measure of Passant is computed. Therefore, for each triple of a path, the $PLDSD$ method applies the $LDSD_{cw}$ to the pair of resources formed by  the subject and the object of the triple. For instance, consider $r_a$ and $r_b$ in the graph of Figure \ref{fig:master}, if we assume $h$ equal to $2$, all the paths linking $r_a$ and $r_b$ with length not greater than $2$ have to be taken into account. For example, if we focus on the path [$\langle r_b, p_3, r_4\rangle$, $\langle r_4, p_2, r_a\rangle$],  $LDSD_{cw}$ is applied to the   pairs of resources ($r_b$, $r_4$) and ($r_4$, $r_a$).

\subsection{Methods based on triple weights}
\label{Methods_weights}
In this subsection the third group of methods is described. It  concerns five different approaches that, in order to compute  semantic relatedness between resources, require the association of weights with triples that allow to evaluate the overall paths. 
They are \textit{Information Content-based Measure} (\textit{ICM}), \textit{REWOrD}, \textit{Exclusivity-based Measure} (\textit{ExclM}), \textit{ASRMP$_m$}, and \textit{Proximity-based Method} (\textit{ProxM}).\\

\noindent \textbf{Information Content-based Measure (\textit{ICM}).}
The method presented in \cite{Schuhmacher2014}, here referred to as \textit{Information Content-based Measure} (\textit{ICM}), relies on the computation of the weights of the triples occurring in the undirected paths connecting the compared resources, up to a given length. 
The weight is evaluated on the basis of the \textit{information content} (\textit{IC})  notion, which needs a probability distribution $P(X)$ over a random variable $X$ to be given, and is defined as $IC(X)$ = $-log P(X)$. The method proposes three strategies, that differ for the adopted probability distribution. The \textit{Joint Information Content} (\textit{jointIC}) strategy considers the joint probability of the predicate and the object of a triple by assuming they are not independent, the \textit{Combined Information Content} (\textit{combIC})  addresses again the joint probability but the predicates and the objects are supposed to be   mutually independent, and the \textit{Information Content and Pointwise Mutual Information} (\textit{IC+PMI}) considers the deviation from independence between the predicate and the object.
According to the evaluation presented in \cite{Schuhmacher2014}, \textit{combIC} outperforms the others and, for this reason, it has been considered in the experimentation of the present work. 

As an example, in order to weigh the triple $\langle r_a, p_2, r_b\rangle$ in the graph of Figure \ref{fig:master} by using the \textit{combIC} strategy, the joint probability of $p_2$ and $r_b$ needs to be computed. Then, since the strategy assumes that predicates and objects are independent, the required probability is given by the sum of the probabilities of $p_2$ and $r_b$. In particular, these two  probabilities are equal to $\frac{9}{22}$ and $\frac{5}{22}$, where $9$ and  $5$  are  the numbers of occurrences of the triples with predicate $p_2$ and object $r_b$, respectively, and $22$ is the total number  triples in the graph. \\

\noindent \textbf{REWOrD.}
The {\it REWOrD} method \cite{Pirro2012} is based on the notion of \textit{informativeness} of predicates, which is inspired by the \textit{Term Frequency-Inverse Document Frequency} (\textit{TF-IDF}).
\textit{TF-IDF} is commonly used in information retrieval to estimate how important a term \textit{w} is in a document \textit{d} belonging to a collection \textit{D} of documents.
When applied to an RDF graph, \textit{TF-IDF} deals with predicates instead of terms, and resources and triples instead of documents, therefore becomes \textit{Predicate Frequency-Inverse Triple Frequency} (\textit{PF-ITF}). 

%In particular, \textit{PF} is ... and \textit{ITF} is ... Then \textit{PF-ITF} is the product of the two.\\

 According to this approach, we need to distinguish between   {\it outgoing} and {\it incoming} \textit{Predicate Frequency} (\textit{PF}).
 In particular, the outgoing \textit{PF} of a predicate $p$ with respect to the resource $r$, say $PF_o^{r}(p)$,  is the ratio between  the number of triples with subject $r$ and predicate $p$, and the number of triples in which $r$ appears either as  subject or  object. 
 Furthermore, the \textit{Inverse Triple Frequency} of the predicate $p$, say $ITF(p)$, is equal to the logarithm of the ratio between the total number of triples in the graph and the number of triples with predicate $p$.
  Then $\textit{PF-}ITF_{o}^{r}(p)$ is defined as the product of $PF_o^{r}(p)$ and $ITF(p)$.
Analogously, the incoming $\textit{PF-}ITF_{i}^{r}(p)$ of the predicate $p$ with respect to the resource $r$ can be defined.

As a result, the weight of a triple $t = \langle r_{k}, p, r_{j}\rangle$, also referred to as the \textit{informativeness} of $t$, takes into account both the $\textit{PF-}ITF_{o}^{r_{k}}(p)$ and the $\textit{PF-}ITF_{i}^{r_{j}}(p)$.

For instance, consider the triple $\langle r_{a}, p_2, r_{b}\rangle$ of the graph in Figure \ref{fig:master}. In order to compute $ITF(p_2)$,  we need: (i) the total number of triples in the graph, that is $22$, and (ii) the number of triples with predicate $p_2$, that is $9$. In addition, in order to compute \textit{PF}$_{o}^{r_{a}}(p_2)$, we have to consider: (i) the  number of outgoing links from $r_a$ with predicate $p_2$, that is $2$, and (ii) the number of triples with $r_a$ either as subject or object, that is 9. Analogously, in order to compute \textit{PF}$_{i}^{r_{b}}(p_2)$ we have to address: (i) the  number of incoming links to $r_b$ with predicate $p_2$, that is $4$, and (ii) the number of triples with $r_b$ either as subject or object, that is $9$.

 According to this method, given an undirected path, its informativeness is the sum of the informativeness of the triples of the path divided by the length of the path. In particular, the \textit{most informative path} (\textit{mip}) is the path with  the greatest informativeness among those connecting the  resources, up to a given length.  \\
In order to evaluate the overall relatedness between resources, we need to build their \textit{relatedness spaces}, i.e.,  vectors of weighted predicates computed according to five alternative strategies. The first strategy focuses on the incoming predicates, the second one on the outgoing predicates, the third one on both the incoming and the outgoing predicates, the fourth one on the \textit{mip}, and the fifth one, which has been addressed in the experimentation of this paper (and  here referred to  as \textit{reword}), on both the incoming predicates and the \textit{mip}.\\ 

\noindent \textbf{Exclusivity-based Measure (\textit{ExclM}).}
The approach proposed in \cite{Hulpus2015}, here referred to as \textit{Exclusivity-based Measure} (\textit{ExclM}), relies on the notion of \textit{exclusivity} of triples. The assumption is that, given two resources connected through a predicate, the less the number of resources linked to them through that predicate, the stronger the relation between them.
In particular, given a triple $t = \langle r_i, p, r_j \rangle$ in an RDF graph, the exclusivity of $t$, which represents the weight of the triple $t$, is defined as the probability to randomly select the triple $t$ out of the set of all the triples with predicate $p$ and subject $r_i$, and  all the triples with predicate $p$ and object $r_j$. 

As an example, in order to associate a weight with the triple $\langle r_a, p_2, r_b \rangle$ in the graph of Figure \ref{fig:master}, two sets of triples have to be considered. In particular, according to the SPARQL notation introduced in Section \ref{sec:rdf_linked_data}, they are the set of  triples of the form $\langle r_a, p_2, ?x \rangle$, i.e., the ones with the outgoing  predicates $p_2$ from $r_a$, and the set  of triples of the form $\langle ?x, p_2, r_b \rangle$, i.e., those with the incoming predicate $p_2$ to $r_b$. 
Then, on the basis of the triple weights, the set of $k$ undirected paths with the greatest weights between the compared resources are considered. Furthermore,
as experimented in the mentioned paper, longer paths contribute less to the relatedness of the compared resources according to a given parameter $\alpha$.
In our experiment, $k$ and $\alpha$ are set to $5$ and $0.25$, respectively, since these are the values suggested by the authors.
\\

\noindent \textbf{ASRMP$_m$.} 
In \cite{ElVaigh2020},  El Vaigh et al. propose the $ASRMP_m$ family of relatedness measures, originating from a previous proposal of the authors, referred to as {\it Weighted Semantic Relatedness Measure} ($WSRM$) \cite{ElVaigh2019}. They state that a well-founded relatedness measure should meet the following three requirements: (i) to have a formal semantics in order to be defined on a knowledge graph such as RDF or OWL (as opposed to Wikipedia), (ii) to have a reasonable computational cost, (iii) to be transitive, in order to capture directly or indirectly related resources, and symmetric.
This family 
is based on the assumption that the more predicates between resources, the stronger their relatedness.  It relies on  the $WSRM$ measure of an ordered pair of resources standing for the subject and the object of a given triple. In particular, such a measure is given by the number of outgoing links from the first resource, i.e., the triple's subject, to the second resource, i.e.,  the triple's object,  normalized to the total number of outgoing links from the triple's subject.

For instance, consider the pair of resources $r_a$, $r_b$ of Figure \ref{fig:master}. Then $WSRM(r_a,r_b)$ = $\frac{1}{4}$ because there is one direct link from $r_a$ to $r_b$, and the total number of outgoing links from $r_a$ is 4.
This family of measures consists of three strategies, namely {\textit {ASRMP}$_m^a$}, {\textit {ASRMP}$_m^b$}, {\textit {ASRMP}$_m^c$}, that  consider all the directed paths between the compared resources, where paths and triples are aggregated by using fuzzy logic operators. In particular, the first strategy addresses paths of a given length, say $m$, the second one of length \textit{less than or equal} to $m$, and the third one also provides a criterion  for which  paths are weighted depending on their lengths. Since paths are directed, the  relatedness of $r_a$ to $r_b$ is first evaluated and then, in order to achieve symmetry, also  the  relatedness of $r_b$ to $r_a$  is computed and their average is considered. In our experiment the  {\textit {ASRMP}$_m^a$} strategy, which is the best measure according the authors, has been addressed.\\

\noindent \textbf{Proximity-based Method (\textit{ProxM}).}
\label{sec:proximity}
The \textit{Proximity-based Method} (\textit{ProxM}) \cite{Leal2013} focuses on the notion of \textit{proximity}, which  has been conceived in order to measure how related two resources are in terms of number of paths between them, rather than addressing the shortest path (distance) between them. A resource may be at the same distance from other resources but it may have more connections (in this proposal undirected paths are considered) with one of them with respect to the others.
Therefore, according to the author, the more paths between resources, the higher their proximity. 
 In order to compute it, in this proposal all the paths connecting the resources, up to a maximum length $h$, are considered. However, in general shorter paths contribute more than longer paths. With regard to triple weights, in the experiment given in \cite{Leal2013} they are manually assigned, therefore the method does not provide any built-in function for weighing triples.\\

In Table \ref{tab:relatednessMethods}, for each of the 10 methods addressed in the paper, some key aspects are summarized that are: the main features of the method; the contributing links of the compared resources or the contributing paths between them; the maximum distance (\textit{Max dist.}) between the compared resources in order to have a non-null semantic relatedness degree; whether the method is symmetric (\textit{Symm.}), i.e., if the order of the resources impacts on the results. Note that, in the case of  \textit{WLM}, \textit{LODDO}, \textit{LDSD}, and \textit{LDSDGN},  if the length of the shortest path between the resources exceeds $2$ their relatedness degree is null, whereas the remaining proposals do not have any constraints about this. Furthermore, all the methods except for \textit{LDSD} are symmetric.

\begin{landscape}
\begin{table*}\caption{Semantic relatedness methods used in the experiment}
\scriptsize
    \begin{tabular}{p{1.0cm} p{2.0cm} p{7.9cm} p{4.9cm} p{0.6cm} p{0.6cm}}
    \hline
    & Method & Main features & Contributing  & Max  & Symm. \\
    & (strategy) &  $ t = \langle s, p, o\rangle$ is a triple  & - links of the compared resources &  dist. &\\
    & & $w(t)$ is the weight of $t$ & - paths b/w the compared resources &  &\\

    \hline
    \multirow{4}{*}{\parbox{1.2cm}{Adjacent resources}}
    &\textit{WLM} \cite{Witten2008} & Inspired by the \textit{Normalized Google Distance} measure. & All the \textit{incoming} links. & 2 & YES \\ \\
    
    &\textit{LODDO}\textcolor{white}{a}\cite{Zhou2012} \hspace{0.3 cm} (\textit{LODOverlap}) & Overlapping of resources descriptions. & All the \textit{incoming} and the \textit{outgoing} links. & 2 & YES\\
    
    \hline
    
    \multirow{8}{*}{\parbox{1.3cm}{Triple patterns}}
    &\textit{LDSD}\textcolor{white}{a}\cite{Passant2010} \hspace{0.3 cm} (\textit{LDSD$_{cw}$}) & Semantic distance  between  resources. & All the \textit{undirected} paths with incoming (outgoing) links labeled with the same predicate. & 2 & NO  \\ \\
    
    &\textit{LDSDGN}\textcolor{white}{a}\cite{Piao2016} \hspace{0.3 cm} (\textit{LDSD$_{\gamma}$}) & Evolution of \textit{LDSD} with global normalization. & All the \textit{undirected} paths as in \textit{LDSD} and extended to the whole graph. & 2 & YES  \\ \\
    
    &\textit{PLDSD} \cite{Alfarhood2017} & \textit{LDSD$_{cw}$} propagated to paths between the compared resources. & The \textit{undirected} path leading to the shortest distance. & ANY & YES \\
    \hline
    \multirow{16}{*}{\parbox{1.2cm}{Triple weights}}
    
    &\textit{ICM}\textcolor{white}{a}\cite{Schuhmacher2014} \hspace{2.0 cm} (\textit{combIC}) & Information Contents (\textit{IC}) of predicates and resources. \hspace{0.5 cm}The weight $w(t)$ is the sum of $IC(p)$ and $IC(o)$. & The \textit{undirected} path with the greatest weight. &  ANY & YES \\ \\
    
    & \textit{REWOrD}\textcolor{white}{a}\cite{Pirro2012} \hspace{0.3 cm} (\textit{reword}) & Predicate Frequency and Inverse Triple Frequency (\textit{PF-ITF}). The weight $w(t)$ is the \textit{Informativeness} of $p$ based on \textit{PF-ITF}. & The \textit{undirected} path with the greatest informativeness (\textit{mip}). &  ANY & YES \\ \\
    
    &\textit{ExclM} \cite{Hulpus2015} &  Longer paths contribute less to the relatedness. 
    
    The weight $w(t)$ is the probability of selecting a triple out of all those with predicate $p$ and either subject $s$ or object $o$. & The k \textit{undirected} paths with the greatest weights. &  ANY & YES \\ \\
    
    &\textit{$ASRMP_m$}\textcolor{white}{a}\cite{ElVaigh2020} \hspace{0.3 cm} ({\textit {ASRMP}$_m^a$}) & The more paths the stronger the relatedness. \hspace{1.4 cm} Fuzzy Logic operators for triple and path aggregations. & All the \textit{directed} paths. &  ANY & YES \\ 
    & & The weight $w(t)$ is the number of links from $s$ to $o$,  normalized to the number of outgoing links from $s$.
    \\ \\
    &\textit{ProxM} \cite{Leal2013} & The more paths the higher the proximity. & All the \textit{undirected} paths. & ANY & YES \\ 
    & & No built-in function for weighting a triple. \\
    
    \hline
    \end{tabular}
\label{tab:relatednessMethods}
\end{table*}
\end{landscape}

\section {Experimentation and Evaluation}
\label{sec:experiment}
In order to evaluate the 10 methods recalled in the previous section, we performed an experimentation by applying them to 14 benchmark  golden datasets, and considering a subgraph of the whole DBpedia knowledge graph, as described below. For each dataset,  we compared the  semantic relatedness values obtained for each method against the  human judgment values provided in  the dataset.
In the next subsections, the portion of the DBpedia knowledge graph and the selected benchmark datasets are outlined. Furthermore, additional details about the set up of the experiment are given and, finally, the evaluation of the methods is illustrated.

\subsection{DBpedia data collections used in the experiment}
\label{sec:dbpediadatasets}
The knowledge graph  addressed in the experimentation is a subgraph of the whole DBpedia obtained by considering  a subset of  its data collections according to the following critera.
Firstly, we referred to the most recent version of the essential DBpedia data focused on English\footnote{https://databus.dbpedia.org/dbpedia/collections/latest-core}.
Secondly, we selected all the data collections containing triples having a resource as object rather than a literal. This choice is in line with the one made by all the methods considered in this paper. In the case of data collections containing triples involving literals as objects, such triples have been removed.
Thirdly, we selected the data collections containing the triples representing the  hyperlinks that appear in the texts of Wikipedia articles. 
It is important to note that all these triples have the same predicate name, i.e., \textit{dbo:wikiPageWikiLink}, and correspond to a very huge number  in the DBpedia graph. Such triples, although with the same predicate name, gather a relevant piece of information for each resource.
Hence, including them in the knowledge graph means to significantly  enrich the  information provided  by the resources' infoboxes that, often, contain just a summary of the most representative information of a given resource. 
For instance, the infobox associated with the resource Michael Jackson contains the information related to the dates of his birth and death, the names of his spouses, children, awards, etc. However, it does not specify anything about, for example, the names of his most popular songs, such as {\it Beat It}, {\it Billie Jean}, or {\it Thriller} that, instead, are described in the corresponding Wikipedia article. 

\begin{figure}[ht]
%\centering
	\includegraphics[width=1.0 \textwidth]{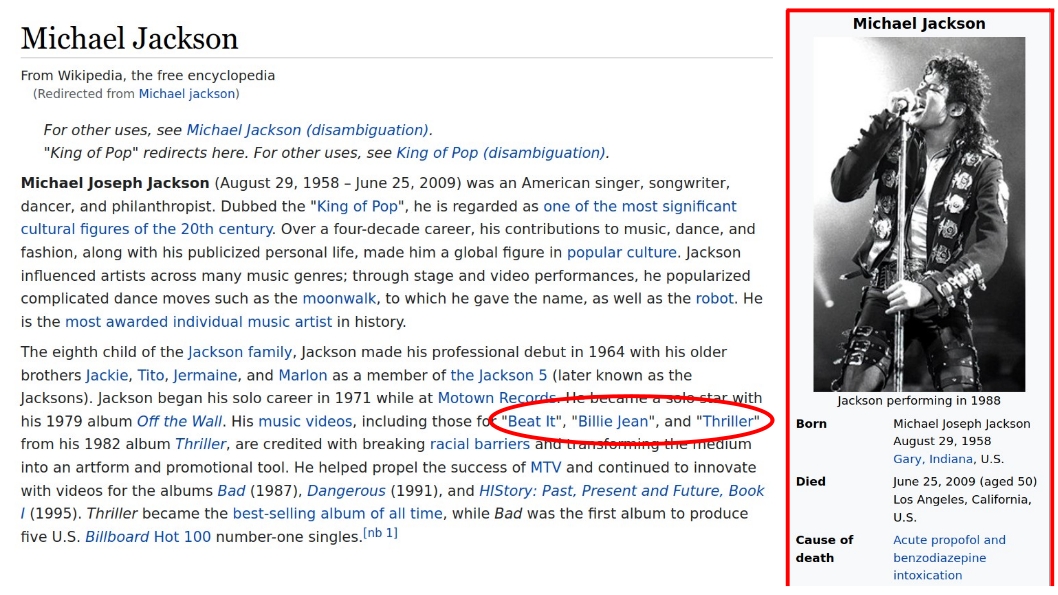}
	\caption{A fragment of the Wikipedia article about Michael Jackson and, on the right side, part of the related infobox}
	\label{fig:MJackson}
\end{figure}

Therefore, excluding the triples with \textit{dbo:wikiPageWikiLink} as predicate in the experimentation means to have, for each resource, a significantly less number of triples to be evaluated, and hence semantic relatedness is computed by relying on the information provided by the resources's infoboxes, mainly. 
For this reason, in order to analyse the relevance of the information contained in the infoboxes in the evaluation of  semantic relatedness, we ran two experiments, the  first by excluding such triples, and the second by including them. 
The total dimension of the DBpedia data collections used in the experimentation is around 61 GB, corresponding to 380,891,403 triples\footnote{All the data collections were downloaded on the 3rd September 2021 from the https://databus.dbpedia.org/dbpedia/collections/latest-core web page, except for the page$\_$links$\_$en.ttl dataset, which was downloaded from the https://wiki.dbpedia.org/downloads-2016-10$\#$h26493-2 web page.}. By removing the triples with the \textit{dbo:wikiPageWikiLink} links, the data decrease to 33 GB, and 207,266,671 triples.

\subsection{Benchmark datasets used in the experimentation}
\label{sec:golden_datasets}
Traditionally, computer-aided tasks are evaluated by comparing the behaviour of the computer against the one of human beings. In the case of methods for automatically evaluating semantic relatedness, they are assessed by setting up experiments where people are asked to express numerical values representing how much, according to their opinion, pre-defined pairs of terms are related. These human judgment values are then compared against the  automatically computed ones. 
Such collections of pairs of terms, where each pair is associated with a human judgment value, represent benchmark datasets. 
In the literature, several benchmark datasets have been defined, often referred to as golden datasets.
In this paper, we considered 14 benchmark datasets from the most representative ones presented in \cite{Taieb2019ASO}. In particular, we selected the datasets in the English language that have been conceived for evaluating semantic relatedness.
%In particular, we selected the  following benchmark datasets
They are\footnote{The number appearing in the dataset name  stands for the number of pairs contained in the dataset.}: Atlasify240 (here, Atlasify for short) \cite{Brent2012}, B\textsubscript{0} (25 pairs) and  B\textsubscript{1} (30 pairs) \cite{Ziegler2006}, GM30 \cite{Gracia2008}, MTurk287 (here, MTurk for short) \cite{Radinsky2011}, Rel122 \cite{Szumlanski2013}, WRG (252 pairs) \cite{Agirre2009}, and KORE (420 pairs) \cite{Hoffart2012} organized into five datasets, namely, KORE-IT, KORE-HW, KORE-VG, KORE-TV, and KORE-CN. 
In addition, we included two datasets, namely, RG65 \cite{Rubenstein1965} and MC30 \cite{Miller1991}, which are traditionally considered milestones in order to assess semantic similarity.

It is important to observe that, among the above datasets, all the  terms in the KORE collections correspond to  DBpedia URIs.
All the other datasets contain words that, in some cases, either do not have a  straightforward correspondence with a DBpedia resource, or correspond to  different DBpedia resources depending on the possible different meanings they have. 
For this reason, in line with \cite{FT2021}, a {\it disambiguation} step has been introduced, as described below.

For each word occurrence in a given dataset, the corresponding resource in DBpedia has been manually selected in accordance with the following disambiguation criteria:

\begin{itemize}
\item If a word is present in the dataset in a plural form, we transformed it into its singular form.
\item If a word in a pair has more than one meaning, and hence can be mapped to more than one DBpedia resource,  we selected the resource whose acceptation is more semantically related to the other word of the pair.
For instance, the word \textit{crain}, in the MC30 dataset, leads to two   DBpedia resources,
namely, \textit{http://dbpedia.org/resource/Crane\_(bird)}, and\\ \textit{http://dbpedia.org/resource/Crane\_(machine)}.\\ Hence, in the case of the pair (\textit{crain}, \textit{bird}), we selected the first resource, which refers to \textit{crane as a bird}, whereas in the case of the pair (\textit{crain}, \textit{implementation}), we selected the second resource, which refers to \textit{crane as a machine} \cite{FT2021}.
\item If a word is a terminological variant, e.g., a synonym, or an acronym, of the name of a  given DBpedia resource, for such a word we selected  that resource.  For instance, in  the case of the acronym \textit{FBI}, we considered the \textit{http://dbpedia.org/resource/Federal\_Bureau\_of\_Investigation} resource.
\end{itemize}

\noindent According to the above criteria, for each dataset except for the ones in KORE that do not need the disambiguation step, we built another dataset, and in this paper we refer to the former as the \textit{original}, and to the latter as the \textit{disambiguated} dataset. Hence, we experimented the methods illustrated in Section \ref{sec:methods_for_semantic_relatedness}  on the selected datasets, according to  both their original and  disambiguated forms.

\subsection{Further experimentation details}
In the experimentation, for each of the 10 methods we considered the strategy, or variant, that according to the authors provides the best performances.
Hence, in the case of \textit{LODDO}, \textit{LDSD}, \textit{LDSDGN}, \textit{ICM}, \textit{REWOrD}, \textit{ExclM}, and \textit{$ASRMP_m$}, we have selected the corresponding variants \textit{$LODOverlap$}, \textit{$LDSD_{cw}$}, \textit{$LDSD_{\gamma}$},  
 \textit{ICM} with \textit{combIC} as weighting function,
%\textit{combIC-R}\footnote{Note that, \textit{combIC-R} stands for the \textit{IC-based$_h$} measure of Eq. \ref{eq:ic_relatedness}
\textit{reword},  \textit{ExclM} with \textit{$k = 5$}  and \textit{$\alpha = 0.25$}, and \textit{$ASRMP_m^a$} with $m = 2$.  
Furthermore, as recalled in Section \ref{Methods_weights} (see also \ref{sec:proximity_2}), \textit{ProxM} does not have a built-in function $w(p_i)$ for weighting a predicate $p_i$. 
In particular, in \cite{Leal2013}, weights  are assigned to predicates by domain experts manually because the graph addressed in the experiment contains a limited number of predicates.
However, assigning weights manually is not a scalable approach with respect to the number of predicates in the graph.
For this reason,  due to the dimension of the DBpedia knowledge graph, in the case of  \textit{ProxM}, in this experiment the weight of a predicate has been defined as its information content, which is a notion that has been attracting a lot of attention in the literature for years \cite{Lin1998}. 
Therefore,  we implemented $w(p_i) = -log(Pr(p_i))$, in accordance with the information content definition provided in Eq. \ref{eq:IC}.

As mentioned, for those methods that natively compute a semantic distance, i.e., \textit{WLM}, \textit{LDSD}, \textit{LDSDGN}, and \textit{PLDSD}, in the experimentation we consider the corresponding relatedness formulation. 
In particular, this formulation depends on whether the method returns a value $v$ in the range $[0, \dots, 1]$, as for \textit{LDSD}, \textit{LDSDGN}, and \textit{PLDSD}, or in the range $[0, \dots, +\infty)$, as for \textit{WLM}. 
In the former case, the corresponding relatedness formulation is defined as $1-v$, whereas in the latter case $\frac{1}{1+v}$.

It is important to recall that, as experimented in \cite{Hulpus2015}, in general the longer the paths the weaker the semantic relation, in the sense that the smaller the inﬂuence of  longer paths, the better the correlation with human judgment. Besides \textit{ExclM}, this  is also the underlying assumption of most of the methods based on triple weights, as for instance \textit{ASRMP$_m$}, and it is in line with  the implicit assumptions made by   \textit{WLM} and \textit{LODDO}, which are  based on adjacent nodes,
and also in line with \textit{LDSD} and \textit{LDSDGN}, which rely on patterns represented by paths of length 2. Therefore,  in order to compare the 10 methods under the same hypotheses,  in our experimentation  the length of the contributing paths is not greater than 2.

In the first experiment, the one without  the \textit{dbo:wikiPageWikiLink} links  in the knowledge graph, we evaluated the 10 methods also on  \textit{clean} datasets, i.e., the disambiguated datasets where the pairs of terms that are not connected by any path of length less than or equal to 2 have been removed. Whereas, in the experiment including the \textit{dbo:wikiPageWikiLink} links, clean datasets have not been addressed since there is a limited number of such pairs that can be neglected.   

In order to compare the methods against human judgment, we considered both the Spearman's and Pearson's correlations.
However, for the five KORE datasets we computed only the Spearman's correlation because for these datasets only  pairwise rankings  are provided without relatedness values.

In the case of the experiment with the \textit{dbo:wikiPageWikiLink} links, we also analyzed the performances of the 10 methods when dealing with pairs of disambiguated terms representing common nouns and proper nouns separately. For this purpose, from each dataset we extracted two additional smaller datasets,  one containing only pairs of common nouns and the other including only pairs of proper nouns. Then, for each method and each of these additional datasets, we computed the Spearman's and the Pearson's correlations against human judgement.

The experimental results are presented and discussed in the next subsection, and all the data are  available at \cite{Taglino2022}. %``https://data.mendeley.com/datasets/78gxwmc6zr/2".

\subsection {Evaluation}
\label{sec:evaluation}

In this section the results of the two experiments are presented and are shown in  Tables \ref{tab:spearman} and  \ref{tab:pearson}, where the Spearman's and Pearson's correlations are given, respectively. As mentioned above, the first experiment concerns DBpedia without including the \textit{dbo:wikiPageWikiLink} links (see columns \textit{w/o} in the tables, where \textit{w/o} stands for \textit{without dbo:wikiPageWikiLink}), whereas in the second experiment these links have been considered (see columns \textit{w} in the tables, where \textit{w} stands for \textit{with dbo:wikiPageWikiLink}). In each table, for each dataset, the results corresponding to the original (o)  and disambiguated (d) datasets are shown in the first and the second rows, respectively.
In addition, in the first experiment,  i.e., the one without \textit{dbo:wikiPageWikiLink} links, the values obtained by considering the clean datasets (c) are given whereas, as mentioned above, in the second experiment these values have not been considered (see the symbol  ``$-$'' in rows c in the tables).
Furthermore, in the tables, the best values are highlighted in bold, and the average correlations (Avg.) for each method are also shown.

\vspace{0.15 cm}
\noindent \textbf{Experiment 1:} DBpedia without \textit{dbo:wikiPageWikiLink} links.

\noindent In the case the triples with the \textit{dbo:wikiPageWikiLink} predicate are not considered in the knowledge graph, both the Spearman's and Pearson's correlations do not provide satisfactory values (see columns \textit{w/o} in Tables  \ref{tab:spearman} and \ref{tab:pearson}, respectively). 
Indeed, for some methods and some datasets, it is not even possible to compute such correlations. For instance, if we consider  the \textit{$ASRMP_m$} method, and the original  golden dataset B$_0$,  for any pair of the dataset there are no directed paths of length 2 connecting the related resources, and then the relatedness values returned by the method are null  for all the pairs of the dataset (see the symbol ``$-$'' in the tables).
Note that in the case of the Spearman's correlation (see Table \ref{tab:spearman}),  \textit{LODDO}  outperforms the other methods in all the three cases, i.e., with original ($0.34$), disambiguated ($0.49$), and clean datasets ($0.62$). According to  Pearson (see Table \ref{tab:pearson}),  when the original datasets are considered,
 \textit{LODDO} and   \textit{ExclM}  provide the highest, although low, results ($0.25$) whereas, in the cases of the disambiguated and clean datasets, \textit{LDSD}  shows the best results by improving its performances from $0.21$ to $0.39$ and $0.50$, respectively.

Overall, the correlation values with human judgment obtained without the  \textit{dbo: wikiPageWikiLink} triples, i.e., by relying mainly on the information of the Wikipedia's infoboxes, are  low. Indeed, the removal of almost half of the triples from the knowledge graph has a great impact on the computation of  semantic relatedness because,  as shown in the second experiment, although  all these triples have the same predicate name, they convey a significant amount of information  for each resource that cannot be ignored.

\vspace{0.15 cm}
\noindent \textbf{Experiment 2:} DBpedia with \textit{dbo:wikiPageWikiLink} links.

\noindent In the presence of  the \textit{dbo:wikiPageWikiLink} predicate, for the majority of the methods
 both the Spearman's and Pearson's correlations   significantly increase  with respect to the results obtained in the Experiment 1 (see columns \textit{w} in  Tables \ref{tab:spearman} and \ref{tab:pearson}, respectively).
Note that, analogously to the previous  experiment, in general the performances of the 10 methods improve by considering the disambiguated datasets. 
 In particular, with regard to the Spearman's correlation, \textit{LODDO} outperforms the other methods when the original datasets are addressed ($0.59$) and, overall, the methods based on adjacent resources show good performances.  This result is interesting, especially if we consider that 
 the methods based on adjacent resources rely on  information that are local to the compared resources,  and therefore they require a smaller number of queries and, of course,  lower computational complexity costs with respect to the other methods. 

%AGGIUNGERE QUI
It is worth noting that, in the case of the disambiguated datasets, on average, both the methods based on adjacent resources and triples patterns give good results. The role of disambiguation is more evident if we observe the results obtained in Table  \ref{tab:spearman}, columns \textit{w}, for  \textit{ExclM}, with $k=5$ and $\alpha = 0.25$, that outperforms the other methods ($0.70$). 

In the case of the Pearson's correlation, for instance, \textit{LDSDGN} increases of $0.21$, and both \textit{WLM} and \textit{LDSD}  of $0.18$.
Furthermore, it is interesting to observe that
 \textit{ASRMP$_m$} outperforms on average the other methods against both the original ($0.48$) and the disambiguated versions of the datasets ($0.63$). Note that, if we consider the datasets individually, \textit{ASRMP$_m$} performs better in half of the cases. 
More specifically, in the case of the Atlasify, MTurk, and WRG datasets, \textit{ASRMP$_m$} provides better results than the other methods, with respect to both the original and the disambiguated versions. 

Overall, if compared to the corresponding correlation values obtained without the \textit{dbo:wikiPageWikiLink} triples, the \textit{ASRMP$_m$} method  significantly   improves its performances. In particular,  in the case of disambiguated datasets, it increases on average not only according to  Spearman ($0.63$ with respect to $0.21$) but also according to Pearson ($0.63$ with respect to $0.14$).
This occurs because  this method relies on directed paths, and the absence of such  triples implies that several pairs of the compared resources are not connected in the graph, leading  therefore   to  null  relatedness degrees.
Indeed,  \textit{ASRMP$_m$} shows the best performance according to the means of the averages of the Spearman's and Pearson's correlations  with \textit{dbo:wikiPageWikiLink} triples ($0.63$), as shown in Table \ref{tab:overall}.

\begin{figure}[ht]
    \centering
    \includegraphics[width=1.0\textwidth]{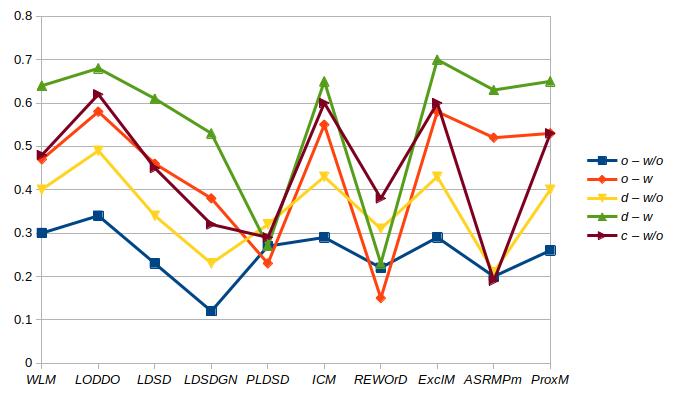}
    \caption{Average Spearman's  correlations line plot}
    \label{fig:Spearman_chart}
\end{figure}

\begin{figure}[ht]
    \centering
    \includegraphics[width=1.0\textwidth]{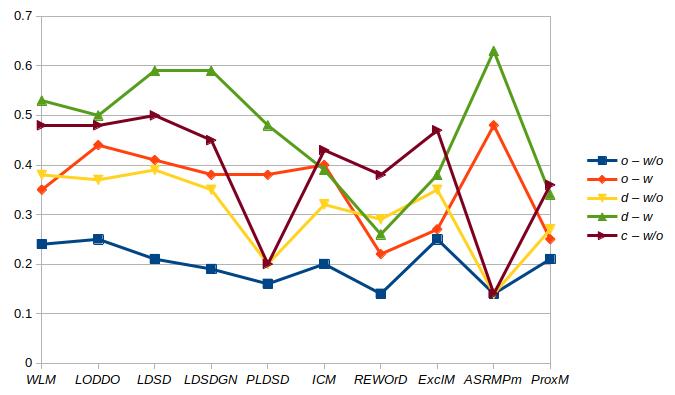}
    \caption{Average Pearson's  correlations line plot}
    \label{fig:Pearson_chart}
\end{figure}

The line plots of the average Spearman's and Pearson's correlation values obtained according to the experimental results are  shown in Figures  \ref{fig:Spearman_chart} and \ref{fig:Pearson_chart}, respectively.

As already mentioned, in the case of the Experiment 2, we also studied the correlations of the 10 methods in the presence  of disambiguated datasets when only pairs of common nouns or pairs of proper nouns are addressed.
Tables   \ref{tab:spearman_common_proper} and \ref{tab:pearson_common_proper} show the experimental results about this further analysis for Spearman and Pearson, respectively. Note that the symbol ``$-$'' in the tables means that either the corresponding dataset does not contain pairs of a given type (e.g., GM30 does not include any pair of proper nouns) or it is not possible to compute the Spearman's correlation (as in the case of the KORE datasets for which only  pairwise rankings  are provided without relatedness values).
The experimental results show that, when considering only pairs of common nouns, according to Spearman \textit{LODDO} outperforms all the other methods ($0.73$), whereas the best Pearson's correlation is achieved by \textit{ASRMP$_m$} ($0.68$). However, if we compute the means of the average Spearman's and Pearson's correlations, both \textit{ASRMP$_m$} and \textit{LODDO} show the best performances ($0.65$). 
In the case of pairs of proper nouns,  \textit{ExclM} provides the best Spearman's correlation ($0.73$), whereas \textit{LDSD} shows the best performance according to Pearson ($0.72$).
Furthermore, \textit{ExclM} outperforms the other methods if we consider the means of the average Spearman's and Pearson's correlations ($0.69$).

%%%%%%%%%%% TABLE 2 %%%%%%%%%%%%%
\begin{landscape}
%\begin{center}
\begin{table*}[ht]
\tiny
\caption{Spearman's correlation for original (o), disambiguated (d), and clean (c) datasets, and DBpedia without \textit{dbo:wikiPageWikiLink} links (\textit{w/o}), and with \textit{dbo:wikiPageWikiLink} links (\textit{w})}
\centering
    \label{tab:spearman}
    \begin{tabular} 
{p{1.7cm} 
p{0.3cm} p{0.2cm} 
p{0.3cm}  p{0.2cm} 
p{0.5cm} p{0.3cm} 
p{0.5cm} p{0.3cm}
p{0.3cm} p{0.5cm} 
p{0.5cm} p{0.3cm} 
p{0.5cm} p{0.5cm} 
p{0.5cm} p{0.3cm} 
p{0.2cm} p{0.2cm} 
p{0.5cm} p{0.2cm}}
    \hline
     Dataset & \multicolumn{2}{c}{\textit{WLM} \cite{Witten2008}} & 
     \multicolumn{2}{c}{\textit{LODDO} \cite{Zhou2012}} & 
     \multicolumn{2}{c}{\textit{LDSD} \cite{Passant2010}}  & 
     \multicolumn{2}{c}{\textit{LDSDGN} \cite{Piao2016}} & 
     \multicolumn{2}{c}{\textit{PLDSD} \cite{Alfarhood2017}} & 
     \multicolumn{2}{c}{\textit{ICM} \cite{Schuhmacher2014}} & 
     \multicolumn{2}{c}{\textit{REWOrD} \cite{Pirro2012}} & 
     \multicolumn{2}{c}{\textit{ExclM} \cite{Hulpus2015}}  & 
     \multicolumn{2}{c}{\textit{ASRMP$_m$} \cite{ElVaigh2020}} & 
     \multicolumn{2}{c}{\textit{ProxM} \cite{Leal2013}}\\ 

\hline
& 
\textit{w/o} & \textit{w} &
\textit{w/o} & \textit{w} &
\textit{w/o} & \textit{w} &
\textit{w/o} & \textit{w} &
\textit{w/o} & \textit{w} &
\textit{w/o} & \textit{w} &
\textit{w/o} & \textit{w} &
\textit{w/o} & \textit{w} &
\textit{w/o} & \textit{w} &
\textit{w/o} & \textit{w} \\

%% Atlasify %%
      \hline
\hfill o & 
.43 & .67 & 
.46 & .69 & 
.26  & .62 & 
.24  & .61 & 
.23  & .37 & 
.39  & .69 & 
\textbf{.47} & .33 & 
.39 & \textbf{.72} & 
.12  & .64 & 
.38  & .66 \\
\textbf{Atlasify}\hfill d & 
.43 & .62 & 
\textbf{.46} & .66 &  
.23 & .58 & 
.20 & .57 & 
.23 & .33 & 
.37 & .66 & 
.44 & .29 & 
.37 & \textbf{.68} & 
.12 & .62 &
.35 & .64 \\
\cite{Brent2012} \hfill c & 
.62 & - & 
\textbf{.69} & - & 
.35  & - & 
.23  & - & 
.24  & - & 
.68  & - & 
.55 & - & 
.67 & - & 
.08  & - & 
.61  & - \\ 
\hline

%% B0 %%
\hfill o & 
\textbf{.32} & .36 & 
.29 & \textbf{.68} & 
.14  & .40 & 
.16  & .35 & 
-  & .46 & 
.16  & .63 & 
.18 & .30 & 
.15 & .60 & 
- & .47 & 
.15  & .43 \\
\textbf{B$_0$}\hfill d & 
.53 & .70 & 
\textbf{.60} & .76 &  
.42 & .67 & 
.44 & .67 & 
.25 & .48 & 
.44 & .76 & 
.36 & .62 & 
.43 & \textbf{.79} & 
.25 & .75 &
.40 & .59\\
\cite{Ziegler2006} \hfill c & 
.77 & - & 
.83 & - & 
.84  & - & 
.87  & - & 
.02  & - & 
\textbf{.90}  & - & 
.75 & - & 
.87 & - & 
.20  & - & 
.68  & - \\ 
\hline

%% B1 %%
\hfill o & 
.16 & \textbf{.30} & 
.13 & .23 & 
.00  & .17 & 
.00  & .19 & 
\textbf{.31}  & .11 & 
.01  & .28 & 
.21 & -.19 & 
.00 & .28 & 
\textbf{.31}  & .25 & 
.01  & .19 \\
\textbf{B$_1$}\hfill d & 
.33 & .69 & 
\textbf{.48} & .66 &  
.26 & .48 & 
.18 & .55 & 
.31 & .34 & 
\textbf{.48} & .71 & 
.38 & .16 & 
.38 & \textbf{.72} & 
.31 & .56 & 
.42 & .54 \\
\cite{Ziegler2006} \hfill c & 
.54 & - & 
.73 & - & 
.53  & - & 
.29  & - & 
.37  & - & 
\textbf{.79}  & - & 
.50 & - & 
.64 & - & 
.37  & - & 
.67  & - \\ 
\hline

%% GM30 %%
\hfill o & 
.22 & .44 & 
.23 & \textbf{.69} & 
\textbf{.25}  & .46 & 
\textbf{.25}  & .47 & 
-  & .27 & 
.20  & .67 & 
.15 & .40 & 
.21 & .63 & 
-  & .58 & 
.21  & .58 \\
\textbf{GM30}\hfill d & 
.22 & .54 & 
\textbf{.50} & .79 &  
.48 & .72 & 
.46 & .59 & 
.16 & .40 & 
.45 & .78 & 
.42 & .28 & 
.46 & \textbf{.82} & 
.00 & .65 &
.43 & .69 \\
\cite{Gracia2008} \hfill c & 
.19 & - & 
\textbf{.65} & - & 
.61  & - & 
.59  & - & 
.07  & - & 
.61  & - & 
.31 & - & 
\textbf{.65} & - & 
.00  & - & 
.48  & - \\ 
\hline

%% MTURK %%
\hfill o & 
.15 & .41 & 
.21 & \textbf{.49} & 
\textbf{.26}  & .43 & 
\textbf{.26}  & .41 & 
.14  & .37 & 
.22  & .46 & 
.20 & .29 & 
.23 & .48 & 
.14  & .45 & 
.21  & .45 \\
\textbf{MTurk}\hfill d & 
.24 & .51 & 
\textbf{.32} & .49 &  
.28 & .49 & 
.28 & .47 & 
.22 & .36 & 
.29 & .52 & 
.23 & .23 & 
.29 & \textbf{.53} & 
.19 & .52 &
.26 & .51 \\
\cite{Radinsky2011} \hfill c & 
.22 & - & 
.37 & - & 
.36  & - & 
.30  & - & 
.26  & - & 
\textbf{.39}  & - & 
.33 & - & 
.37 & - & 
.23  & - & 
.18  & - \\ 
\hline

%% Rel122 %%
\hfill o & 
.15 & .39 & 
.17 & \textbf{.46} & 
.08  & .33 & 
.07  & .36 & 
.02  & .35 & 
.10  & .41 & 
\textbf{.19} & .31 & 
.11 & .44 & 
.02  & .39 & 
.01  & .39 \\
\textbf{Rel122}\hfill d & 
.25 & \textbf{.66} & 
\textbf{.36} & .65 &  
.17 & .58 & 
.15 & .57 & 
.09 & .48 & 
.26 & .61 & 
.12 & .30 & 
.27 & \textbf{.66} & 
.09 & .58 &
.25 & .57 \\
\cite{Szumlanski2013} \hfill c & 
.32 & - & 
\textbf{.55} & - & 
.20  & - & 
.06  & - & 
.09  & - & 
.41  & - & 
.19 & - & 
.53 & - & 
.09  & - & 
.36  & - \\ 
\hline

%% WRG %%
\hfill o & 
.25 & .33 & 
\textbf{.26} & .46 & 
.14  & .33 & 
.14  & .37 & 
.18  & .42 & 
.24  & .46 & 
-.03 & .00 & 
.24 & \textbf{.47} & 
.16  & \textbf{.47} & 
.24  & .41 \\
\textbf{WRG}\hfill d & 
.32 & .51 & 
\textbf{.33} & \textbf{.58} &  
.18 & .44 & 
.19 & .49 & 
.20 & .39 & 
.29 & .55 & 
.14 & .09 & 
.29 & .57 & 
.15 & .54 &
.29 & .56 \\
\cite{Agirre2009} \hfill c & 
.40 & - & 
.45 & - & 
.18  & - & 
.26  & - & 
.29  & - & 
.50  & - & 
.06 & - & 
.52 & - & 
.20  & - & 
\textbf{.56}  & - \\ 
\hline

%% RG65 %%
\hfill o & 
.19 & .36 & 
.23 & \textbf{.68} & 
.18  & .40 & 
.18  & .35 & 
\textbf{.32}  & .46 & 
.21  & .63 & 
.21 & .30 & 
.20 & .60 & 
.20  & .50 & 
.21  & .54 \\
\textbf{RG65}\hfill d & 
.55 & .70 & 
\textbf{.62} & .76 &  
.54 & .67 & 
.55 & .67 & 
.24 & .48 & 
.58 & .76 & 
.52 & .62 & 
.58 & \textbf{.79} & 
.00 & .75 &
.58 & .75 \\
\cite{Rubenstein1965} \hfill c & 
.70 & - & 
\textbf{.78} & - & 
.66  & - & 
.68  & - & 
.20  & - & 
.77  & - & 
.66 & - & 
.77 & - & 
.00  & - & 
\textbf{.78}  & - \\ 
\hline

%% MC30 %%
\hfill o & 
.08 & .16 & 
.01 & \textbf{.54} & 
-.18  & .26 & 
-.19  & .15 & 
\textbf{.33}  & .26 & 
-.09  & .45 & 
.09 & .23 & 
-.11 & .42 & 
\textbf{.33}  & .42 & 
-.09  & .30 \\
\textbf{MC30}\hfill d & 
.35 & .81 & 
\textbf{.36} & \textbf{.86} &  
.16 & .76 & 
.17 & .76 & 
.20 & .30 & 
.25 & .81 & 
.28 & .54 & 
.25 & .79 & 
.23 & .79 &
.25 & .80 \\
\cite{Miller1991} \hfill c & 
.64 & - & 
.75 & - & 
.52  & - & 
.58  & - & 
.00  & - & 
.73  & - & 
.54 & - & 
\textbf{.76} & - & 
.00  & - & 
.73  & - \\ 
\hline

%% KORE-IT %%
\hfill o & 
.49 & .63 & 
\textbf{.69} & \textbf{.76} & 
.51 & .68 & 
.06 & .32 & 
.65 & -.01 & 
.64 & .62 & 
.40 & -.06 & 
.62 & .75 & 
.57 & .65 & 
.58 & .68  \\
\textbf{KORE-IT}\hfill d & 
.49 & .63 & 
\textbf{.69} & \textbf{.76} & 
.51 & .68 & 
.06 & .32 & 
.65 & -.01 & 
.64 & .62 & 
.40 & -.06 & 
.62 & .75 & 
.57 & .65 & 
.58 & .68  \\
\cite{Hoffart2012} \hfill c & 
.49 & - & 
\textbf{.69} & - & 
.51 & - & 
.06 & - & 
.65 & - & 
.64 & - & 
.40 & - & 
.62 & - & 
.57 & - & 
.58 & -  \\ 
\hline

%% KORE-HW %%
\hfill o & 
.33 & .53 & 
.61 & .52 & 
.62 & \textbf{.73} & 
.32 & .44 & 
\textbf{.63} & -.08 & 
.62 & .66 & 
.46 & .37 & 
.59 & .71 & 
.31 & .54 & 
.62 & .72  \\
\textbf{KORE-HW}\hfill d & 
.33 & .53 & 
.61 & .52 & 
.62 & \textbf{.73} & 
.32 & .44 & 
\textbf{.63} & -.08 & 
.62 & .66 & 
.46 & .37 & 
.59 & .71 & 
.31 & .54 & 
.62 & .72  \\
\cite{Hoffart2012} \hfill c & 
.33 & - & 
.61 & - & 
.62 & - & 
.32 & - & 
\textbf{.63} & - & 
.62 & - & 
.46 & - & 
.59 & - & 
.31 & - & 
.62 & -  \\ 
\hline

%% KORE-VG %%
\hfill o & 
.42 & \textbf{.70} & 
\textbf{.51} & .60 & 
.19 & .48 & 
.35 & .51 & 
.33 & -.11 & 
.44 & .54 & 
.27 & .10 & 
.49 & .67 & 
.36 & .62 & 
.43 & .56  \\
\textbf{KORE-VG}\hfill d & 
.42 & \textbf{.70} & 
\textbf{.51} & .60 & 
.19 & .48 & 
.35 & .51 & 
.33 & -.11 & 
.44 & .54 & 
.27 & .10 & 
.49 & .67 & 
.36 & .62 & 
.43 & .56  \\
\cite{Hoffart2012} \hfill c & 
.42 & - & 
\textbf{.51} & - & 
.19 & - & 
.35 & - & 
.33 & - & 
.44 & - & 
.27 & - & 
.49 & - & 
.36 & - & 
.43 & -  \\ 
\hline

%% KORE-TV %%
\hfill o & 
\textbf{.54} & .64 & 
.45 & \textbf{.71} & 
.39 & .61 & 
-.01 & .43 & 
.53 & .01 & 
.48 & .68 & 
.19 & -.41 & 
.46 & .62 & 
.18 & .59 & 
.46 & .70  \\
\textbf{KORE-TV}\hfill d & 
\textbf{.54} & .64 & 
.45 & \textbf{.71} & 
.39 & .61 & 
-.01 & .43 & 
.53 & .01 & 
.48 & .68 & 
.19 & -.41 & 
.46 & .62 & 
.18 & .59 & 
.46 & .70  \\
\cite{Hoffart2012} \hfill c & 
\textbf{.54} & - & 
.45 & - & 
.39 & - & 
-.01 & - & 
.53 & - & 
.48 & - & 
.19 & - & 
.46 & - & 
.18 & - & 
.46 & -  \\ 
\hline

%% KORE-CN %%
\hfill o & 
.53 & .69 & 
\textbf{.55} & .73 & 
.34 & .59 & 
-.13 & .38 & 
.43 & .37 & 
.42 & .49 & 
.13 & .09 & 
.47 & \textbf{.74} & 
.13 & .66 & 
.28 & .65 \\
\textbf{KORE-CN}\hfill d & 
.53 & .69 & 
\textbf{.55} & .73 & 
.34 & .59 & 
-.13 & .38 & 
.43 & .37 & 
.42 & .49 & 
.13 & .09 & 
.47 & \textbf{.74} & 
.13 & .66 & 
.28 & .65 \\
\cite{Hoffart2012} \hfill c & 
.53 & - & 
\textbf{.55} & - & 
.34 & - & 
-.13 & - & 
.43 & - & 
.42 & - & 
.13 & - & 
.47 & - & 
.13 & - & 
.28 & -  \\ 
\hline

%% Avg. %%
\hline
\hline
\hfill o & 
.30 & .47 & 
\textbf{.34} & \textbf{.59} & 
.23  & .46 & 
.12  & .38 & 
.27  & .23 & 
.29  & .55 & 
.22 & .15 & 
.29 & .58 & 
.20  & .52 & 
.26  & .53 \\
\textbf{Avg.}\hfill d & 
.40 & .64 & 
\textbf{.49} & .68 &  
.34 & .61 & 
.23 & .53 & 
.32 & .27 & 
.43 & .65 & 
.31 & .23 & 
.43 & \textbf{.70} & 
.21 & .63 &
.40 & .65 \\
\hfill c & 
.48 & - & 
\textbf{.62} & - & 
.45  & - & 
.32  & - & 
.29  & - & 
.60  & - & 
.38 & - & 
.60 & - & 
.19  & - & 
.53  & - \\
\hline
    \end{tabular}
\end{table*}
%\end{center}
\end{landscape}

\begin{sidewaystable}
%%%%%%%%%%% TABLE 3 %%%%%%%%%%%%%
%%%%\begin{landscape}
%\begin{center}
%%%%\begin{table*}[ht]
\tiny
\caption{Pearson's correlation for original (o), disambiguated (d), and clean (c) datasets, and DBpedia without \textit{dbo:wikiPageWikiLink} links (\textit{w/o}), and with \textit{dbo:wikiPageWikiLink} links (\textit{w})}\label{tab:pearson}
\centering
    \begin{tabular} 
{p{1.7cm} 
p{0.3cm} p{0.2cm} 
p{0.3cm}  p{0.2cm} 
p{0.5cm} p{0.3cm} 
p{0.5cm} p{0.3cm}
p{0.3cm} p{0.5cm} 
p{0.5cm} p{0.3cm} 
p{0.5cm} p{0.5cm} 
p{0.5cm} p{0.3cm} 
p{0.2cm} p{0.2cm} 
p{0.5cm} p{0.2cm}}
    \hline
     Dataset & \multicolumn{2}{c}{\textit{WLM} \cite{Witten2008}} & 
     \multicolumn{2}{c}{\textit{LODDO} \cite{Zhou2012}} & 
     \multicolumn{2}{c}{\textit{LDSD} \cite{Passant2010}}  & 
     \multicolumn{2}{c}{\textit{LDSDGN} \cite{Piao2016}} & 
     \multicolumn{2}{c}{\textit{PLDSD} \cite{Alfarhood2017}} & 
     \multicolumn{2}{c}{\textit{ICM} \cite{Schuhmacher2014}} & 
     \multicolumn{2}{c}{\textit{REWOrD} \cite{Pirro2012}} & 
     \multicolumn{2}{c}{\textit{ExclM} \cite{Hulpus2015}}  & 
     \multicolumn{2}{c}{\textit{ASRMP$_m$} \cite{ElVaigh2020}} & 
     \multicolumn{2}{c}{\textit{ProxM} \cite{Leal2013}}\\ 

\hline
& 
\textit{w/o} & \textit{w} &
\textit{w/o} & \textit{w} &
\textit{w/o} & \textit{w} &
\textit{w/o} & \textit{w} &
\textit{w/o} & \textit{w} &
\textit{w/o} & \textit{w} &
\textit{w/o} & \textit{w} &
\textit{w/o} & \textit{w} &
\textit{w/o} & \textit{w} &
\textit{w/o} & \textit{w} \\

%% Atlasify %%
      \hline
\hfill o & 
.42 & .58 & 
.27 & .49 & 
.40  & .60 & 
.31  & .57 & 
.19  & .48 & 
.19  & .21 & 
\textbf{.51} & .33 & 
.34 & .35 & 
.15  & \textbf{.62} & 
.09  & .09 \\
\textbf{Atlasify}\hfill d & 
.42 & .51 & 
.27 & .49 &  
.40 & .58 & 
.29 & .53 & 
.19 & .41 & 
.19 & .19 & 
\textbf{.48} & .28 & 
.34 & .35 & 
.15 & \textbf{.61} &
.09 & .09 \\
\cite{Brent2012} \hfill c & 
\textbf{.55} & - & 
.33 & - & 
.50  & - & 
.32  & - & 
.19  & - & 
.20  & - & 
\textbf{.55} & - & 
.43 & - & 
.15  & - & 
.12  & - \\ 
\hline

%% B0 %%
\hfill o & 
.35 & \textbf{.55} & 
.31 & .47 & 
\textbf{.36}  & .51 & 
\textbf{.36}  & .52 & 
-  & .35 & 
.27  & .47 & 
.24 & .15 & 
.34 & .35 & 
-  & .47 & 
.34  & .35 \\
\textbf{B$_0$}\hfill d & 
.55 & .63 & 
.57 & .62 &  
.57 & .66 & 
\textbf{.61} & \textbf{.69} & 
.26 & .45 & 
.53 & .59 & 
.33 & .11 & 
.51 & .51 & 
.26 & .62 &
.46 & .52\\
\cite{Ziegler2006} \hfill c & 
.68 & - & 
.73 & - & 
.75  & - & 
\textbf{.86}  & - & 
.29  & - & 
.84  & - & 
.76 & - & 
.62 & - & 
.29  & - & 
.53  & - \\ 
\hline

%% B1 %%
\hfill o & 
.22 & .33 & 
.33 & \textbf{.46} & 
.30  & .33 & 
.20  & .28 & 
.30  & .23 & 
.23  & .35 & 
.14 & -.16 & 
\textbf{.36} & .37 & 
.30  & .45 & 
.30  & .30 \\
\textbf{B$_1$}\hfill d & 
.41 & .57 & 
\textbf{.42} & .49 &  
.41 & .57 & 
.22 & .54 & 
.30 & .33 & 
.31 & .31 & 
.32 & .16 & 
.40 & .44 & 
.30 & \textbf{.74} & 
.27 & .27 \\
\cite{Ziegler2006} \hfill c & 
\textbf{.57} & - & 
.46 & - & 
\textbf{.57}  & - & 
.42  & - & 
.37  & - & 
.25  & - & 
.25 & - & 
.49 & - & 
.32  & - & 
.20  & - \\ 
\hline

%% GM30 %%
\hfill o & 
.29 & .46 & 
.28 & .45 & 
\textbf{.34}  & .50 & 
\textbf{.34}  & .51 & 
-  & .49 & 
.27  & .30 & 
.01 & .36 & 
.29 & .31 & 
-  & \textbf{.53} & 
.27  & .27 \\
\textbf{GM30}\hfill d & 
.29 & .34 & 
.32 & .47 &  
\textbf{.43} & .63 & 
.38 & .58 & 
.14 & \textbf{.65} & 
.29 & .31 & 
.37 & .19 & 
.31 & .34 & 
.00 & .51 &
.27 & .27 \\
\cite{Gracia2008} \hfill c & 
.32 & - & 
.47 & - & 
\textbf{.55}  & - & 
.44  & - & 
.13  & - & 
.42  & - & 
.34 & - & 
.45 & - & 
.00  & - & 
.41  & - \\ 
\hline

%% MTURK %%
\hfill o & 
.17 & .38 & 
\textbf{.26} & .28 & 
\textbf{.26}  & .43 & 
.22  & .42 & 
.16  & .42 & 
.25  & .49 & 
.21 & .28 & 
.20 & .22 & 
.14  & \textbf{.50} & 
.10  & .17 \\
\textbf{MTurk}\hfill d & 
.25 & .42 & 
.22 & .31 &  
\textbf{.33} & .48 & 
.29 & .52 & 
.24 & .36 & 
.18 & .20 & 
.28 & .23 & 
.21 & .13 & 
.16 & \textbf{.58} &
.15 & .15 \\
\cite{Radinsky2011} \hfill c & 
.19 & - & 
.25 & - & 
\textbf{.36}  & - & 
.28  & - & 
.26  & - & 
.20  & - & 
.33 & - & 
.23 & - & 
.17  & - & 
.18  & - \\ 
\hline

%% Rel122 %%
\hfill o & 
.17 & .37 & 
\textbf{.20} & \textbf{.39} & 
.12  & .32 & 
.15  & .36 & 
.04  & .38 & 
.14  & \textbf{.39} & 
.14 & .30 & 
.16 & .21 & 
.04  & .37 & 
.15  & .18 \\
\textbf{Rel122}\hfill d & 
.26 & \textbf{.66} & 
\textbf{.32} & .48 &  
.26 & .56 & 
.22 & .57 & 
.08 & .59 & 
.29 & .57 & 
.09 & .32 & 
.25 & .29 & 
.08 & .53 &
.24 & .37 \\
\cite{Szumlanski2013} \hfill c & 
.32 & - & 
\textbf{.47} & - & 
.31  & - & 
.24  & - & 
.10  & - & 
.45  & - & 
.23 & - & 
.35 & - & 
.10  & - & 
.31  & - \\ 
\hline

%% WRG %%
\hfill o & 
.23 & .22 & 
.18 & .39 & 
.18  & .33 & 
.18  & .37 & 
.16  & .40 & 
\textbf{.24}  & .38 & 
.01 & .15 & 
.16 & .17 & 
.14  & \textbf{.44} & 
.14  & .17 \\
\textbf{WRG}\hfill d & 
\textbf{.29} & .33 & 
.16 & .32 &  
.24 & .41 & 
.24 & .48 & 
.17 & .40 & 
.09 & .12 & 
.17 & .24 & 
.20 & .21 & 
.15 & \textbf{.50} &
.03 & .03 \\
\cite{Agirre2009} \hfill c & 
\textbf{.41} & - & 
.19 & - & 
.33  & - & 
.31  & - & 
.20  & - & 
.07  & - & 
.16 & - & 
.28 & - & 
.21  & - & 
.02  & - \\ 
\hline

%% RG65 %%
\hfill o & 
.20 & .26 & 
.29 & .56 & 
.20  & .47 & 
.17  & .32 & 
\textbf{.32}  & .43 & 
.27  & \textbf{.60} & 
.09 & .23 & 
.20 & .21 & 
.19  & .51 & 
.22  & .30 \\
\textbf{RG65}\hfill d & 
.57 & .57 & 
\textbf{.60} & .63 &  
.58 & .68 & 
.57 & .66 & 
.25 & .62 & 
.56 & .57 & 
.42 & .44 & 
.52 & .52 & 
.00 & .74 &
.55 & \textbf{.75} \\
\cite{Rubenstein1965} \hfill c & 
.70 & - & 
\textbf{.78} & - & 
.66  & - & 
.68  & - & 
.20  & - & 
.77  & - & 
.66 & - & 
.77 & - & 
.00  & - & 
\textbf{.78}  & - \\ 
\hline

%% MC30 %%
\hfill o & 
.08 & .05 & 
.12 & \textbf{.48} & 
-.26  & .24 & 
-.19  & .10 & 
\textbf{.26}  & .26 & 
-.05  & .42 & 
-.10 & .32 & 
.20 & .25 & 
\textbf{.26}  & .44 & 
\textbf{.26}  & .46 \\
\textbf{MC30}\hfill d & 
.37 & .75 & 
\textbf{.43} & .70 &  
.27 & .70 & 
.32 & .75 & 
.20 & .54 & 
.41 & .65 & 
.11 & .34 & 
\textbf{.43} & .63 & 
.18 & \textbf{.80} &
.41 & .64 \\
\cite{Miller1991} \hfill c & 
.55 & - & 
\textbf{.68} & - & 
.47  & - & 
.53  & - & 
.08  & - & 
.66  & - & 
.14 & - & 
.64 & - & 
.00  & - & 
.65  & - \\ 
\hline

%% KORE %%
\hfill o & 
- & - & 
- & - & 
- & - & 
- & - & 
- & - & 
- & - & 
- & - & 
- & - & 
- & - & 
- & -  \\
\textbf{KORE}\hfill d & 
- & - & 
- & - & 
- & - & 
- & - & 
- & - & 
- & - & 
- & - & 
- & - & 
- & - & 
- & -  \\
\cite{Hoffart2012} \hfill c & 
- & - & 
- & - & 
- & - & 
- & - & 
- & - & 
- & - & 
- & - & 
- & - & 
- & - & 
- & -  \\ 
\hline

%% Avg. %%
\hline
\hline
\hfill o & 
.24 & .35 & 
\textbf{.25} & .44 & 
.21  & .41 & 
.19  & .38 & 
.16  & .38 & 
.20  & .40 & 
.14 & .22 & 
\textbf{.25} & .27 & 
.14  & \textbf{.48} & 
.21  & .25 \\
\textbf{Avg.}\hfill d & 
.38 & .53 & 
.37 & .50 &  
\textbf{.39} & .59 & 
.35 & .59 & 
.20 & .48 & 
.32 & .39 & 
.29 & .26 & 
.35 & .38 & 
.14 & \textbf{.63} &
.27 & .34 \\
\hfill c & 
.48 & - & 
.48 & - & 
\textbf{.50}  & - & 
.45  & - & 
.20  & - & 
.43  & - & 
.38 & - & 
.47 & - & 
.14  & - & 
.36  & - \\
\hline
    \end{tabular}
%%%%\end{table*}
%\end{center}
%%%%\end{landscape}

\vspace{2\baselineskip}

%%%%%%%%%%% TABLE 4 %%%%%%%%%%%%%
%%%%\begin{landscape}
%\begin{center}
%%%%\begin{table*}[ht]
\tiny
\caption{Means of the average Spearman's and Pearson's correlations for disambiguated datasets and DBpedia with \textit{dbo:wikiPageWikiLinks} links}\label{tab:overall}
\centering
    \begin{tabular} 
{p{1.2cm} 
p{1.4cm} 
p{1.2cm} 
p{1.6cm} 
p{1.4cm} 
p{1.0cm} 
p{1.6cm} 
p{1.2cm} 
p{1.6cm} 
p{1.3cm} }
    \hline
     \textit{WLM} \cite{Witten2008} & 
     \textit{LODDO} \cite{Zhou2012} & 
     \textit{LDSD} \cite{Passant2010}  & 
     \textit{LDSDGN} \cite{Piao2016} & 
     \textit{PLDSD} \cite{Alfarhood2017} & 
     \textit{ICM} \cite{Schuhmacher2014} & 
     \textit{REWOrD} \cite{Pirro2012} & 
     \textit{ExclM} \cite{Hulpus2015}  & 
     \textit{ASRMP$_m$} \cite{ElVaigh2020} & 
     \textit{ProxM} \cite{Leal2013}\\ 

%% With overall %%
      \hline
.59 & .59 & .60 & .56 & .38 & .52 & .25 & .54 & \textbf{.63} & .50\\
\hline
    \end{tabular}
\end{sidewaystable}

\begin{sidewaystable}
%%%%%%%%%%% TABLE 5 %%%%%%%%%%%%%
%%%%\begin{landscape}
%\begin{center}
%%%%\begin{table*}[ht]
\tiny
\caption{Spearman's correlation for common nouns (\textit{c}) and proper nouns (\textit{p}) in disambiguated datasets and DBpedia with \textit{dbo:wikiPageWikiLinks} links}\label{tab:spearman_common_proper}
\centering
    \begin{tabular} 
{p{1.9cm} 
p{0.3cm} p{0.5cm} 
p{0.3cm}  p{0.5cm} 
p{0.5cm} p{0.3cm} 
p{0.5cm} p{0.3cm}
p{0.3cm} p{0.5cm} 
p{0.5cm} p{0.3cm} 
p{0.5cm} p{0.5cm} 
p{0.5cm} p{0.3cm} 
p{0.2cm} p{0.2cm} 
p{0.5cm} p{0.2cm}}
    \hline
     Dataset & \multicolumn{2}{c}{\textit{WLM} \cite{Witten2008}} & 
     \multicolumn{2}{c}{\textit{LODDO} \cite{Zhou2012}} & 
     \multicolumn{2}{c}{\textit{LDSD} \cite{Passant2010}}  & 
     \multicolumn{2}{c}{\textit{LDSDGN} \cite{Piao2016}} & 
     \multicolumn{2}{c}{\textit{PLDSD} \cite{Alfarhood2017}} & 
     \multicolumn{2}{c}{\textit{ICM} \cite{Schuhmacher2014}} & 
     \multicolumn{2}{c}{\textit{REWOrD} \cite{Pirro2012}} & 
     \multicolumn{2}{c}{\textit{ExclM} \cite{Hulpus2015}}  & 
     \multicolumn{2}{c}{\textit{ASRMP$_m$} \cite{ElVaigh2020}} & 
     \multicolumn{2}{c}{\textit{ProxM} \cite{Leal2013}}\\ 

\hline
& 
\textit{c} & \textit{p} &
\textit{c} & \textit{p} &
\textit{c} & \textit{p} &
\textit{c} & \textit{p} &
\textit{c} & \textit{p} &
\textit{c} & \textit{p} &
\textit{c} & \textit{p} &
\textit{c} & \textit{p} &
\textit{c} & \textit{p} &
\textit{c} & \textit{p} \\

%% Atlasify %%
      \hline
\textbf{Atlasify}\cite{Brent2012}&.70&.80&.77&.79&.74&.73&.64&.91&.45&.50&.76&.90&.51&.56&.69&.84&.70&.92&.68&.92\\
%% B0 %%
      \hline
\textbf{B$_0$}\cite{Ziegler2006}&.51&.89&.92&.85&.92&.81&.92&.86&.79&.24&.92&.82&.54&.59&.81&.92&.87&.49&.95&.62\\
%% B1 %%
      \hline
\textbf{B$_1$}\cite{Ziegler2006}&.40&.64&.70&.38&.80&.40&.10&.25&.87&-.22&-.21&.59&.30&-.34&.00&.62&.11&.44&.16&.27\\
%% GM30 %%
      \hline
\textbf{GM30}\cite{Gracia2008}&.54&-&.79&-&.72&-&.59&-&.40&-&.78&-&.28&-&.82&-&.65&-&.69&-\\
%% MTurk %%
      \hline
\textbf{MTurk}\cite{Radinsky2011}&.47&-.07&.56&-.24&.47&.38&.47&.28&.42&.38&.53&.20&.30&-.32&.54&.25&.50&.25&.53&.32\\
%% Rel120 %%
      \hline
\textbf{Rel122}\cite{Szumlanski2013}&.66&-&.65&-&.58&-&.57&-&.48&-&.61&-&.30&-&.66&-&.58&-&.57&-\\
%% WRG252 %%
      \hline
\textbf{WRG}\cite{Agirre2009}&.52&1.0&.57&1.0&.44&1.0&.50&1.0&.40&1.0&.54&1.0&.08&1.0&.56&1.0&.53&1.0&.55&1.0\\
%% RG65 %%
      \hline
\textbf{RG65}\cite{Rubenstein1965}&.70&-&.76&-&.67&-&.67&-&.48&-&.76&-&.62&-&.79&-&.75&-&.75&-\\
%% MC30 %%
      \hline
\textbf{MC30}\cite{Miller1991}&.81&-&.86&-&.76&-&.76&-&.30&-&.81&-&.54&-&.79&-&.79&-&.80&-\\
%% KORE-IT %%
      \hline
\textbf{KORE-IT}\cite{Hoffart2012}&-&.63&-&.76&-&.68&-&.32&-&-.01&-&.62&-&-.06&-&.75&-&.65&-&.68\\
%% KORE-HW %%
      \hline
\textbf{KORE-HW}\cite{Hoffart2012}&-&.53&-&.52&-&.73&-&.44&-&-.08&-&.66&-&.37&-&.71&-&.54&-&.72\\
%% KORE-VG %%
      \hline
\textbf{KORE-VG}\cite{Hoffart2012}&-&.70&-&.60&-&.48&-&.51&-&-.11&-&.54&-&.10&-&.67&-&.62&-&.43\\
%% KORE-TV %%
      \hline
\textbf{KORE-TV}\cite{Hoffart2012}&-&.69&-&.73&-&.59&-&.38&-&.37&-&.49&-&.09&-&.74&-&.66&-&.65\\
%% KORE-CN %%
      \hline
\textbf{KORE-CN}\cite{Hoffart2012}&-&.69&-&.73&-&.59&-&.38&-&.37&-&.49&-&.09&-&.74&-&.66&-&.65\\
\hline
\hline
\textbf{Avg.}&.59&.65&\textbf{.73}&.56&.68&.66&.58&.66&.51&.38&.61&.70&.39&.30&.63&\textbf{.73}&.61&.62&.63&.62\\

\hline
    \end{tabular}
%%%%\end{table*}
%\end{center}
%%%%\end{landscape}

\vspace{2\baselineskip}
%%%%%%%%%%% TABLE 6 %%%%%%%%%%%%%
%%%%\begin{landscape}
%\begin{center}
%%%%\begin{table*}[ht]
\tiny
\caption{Pearson's correlation for common nouns (\textit{c}) and proper nouns (\textit{p}) in disambiguated datasets and DBpedia with \textit{dbo:wikiPageWikiLinks} links}\label{tab:pearson_common_proper}
\centering
    
    \begin{tabular} 
{p{1.0cm} 
p{0.3cm} p{0.2cm} 
p{0.3cm}  p{0.5cm} 
p{0.5cm} p{0.3cm} 
p{0.5cm} p{0.5cm}
p{0.3cm} p{0.5cm} 
p{0.5cm} p{0.3cm} 
p{0.5cm} p{0.5cm} 
p{0.5cm} p{0.3cm} 
p{0.2cm} p{0.2cm} 
p{0.5cm} p{0.2cm}}
    \hline
     Dataset & \multicolumn{2}{c}{\textit{WLM} \cite{Witten2008}} & 
     \multicolumn{2}{c}{\textit{LODDO} \cite{Zhou2012}} & 
     \multicolumn{2}{c}{\textit{LDSD} \cite{Passant2010}}  & 
     \multicolumn{2}{c}{\textit{LDSDGN} \cite{Piao2016}} & 
     \multicolumn{2}{c}{\textit{PLDSD} \cite{Alfarhood2017}} & 
     \multicolumn{2}{c}{\textit{ICM} \cite{Schuhmacher2014}} & 
     \multicolumn{2}{c}{\textit{REWOrD} \cite{Pirro2012}} & 
     \multicolumn{2}{c}{\textit{ExclM} \cite{Hulpus2015}}  & 
     \multicolumn{2}{c}{\textit{ASRMP$_m$} \cite{ElVaigh2020}} & 
     \multicolumn{2}{c}{\textit{ProxM} \cite{Leal2013}}\\ 

\hline
& 
\textit{c} & \textit{p} &
\textit{c} & \textit{p} &
\textit{c} & \textit{p} &
\textit{c} & \textit{p} &
\textit{c} & \textit{p} &
\textit{c} & \textit{p} &
\textit{c} & \textit{p} &
\textit{c} & \textit{p} &
\textit{c} & \textit{p} &
\textit{c} & \textit{p} \\

%% Atlasify %%
      \hline
\textbf{Atlasify}\cite{Brent2012}&.56&.63&.48&.81&.66&.88&.65&.61&.54&.35&.21&.78&.44&.55&.36&.70&.65&.91&.13&.45\\
%% B0 %%
      \hline
\textbf{B$_0$}\cite{Ziegler2006}&.54&.74&.82&.83&.72&.79&.82&.76&.86&.35&.75&.82&.43&.41&.67&.57&.89&.60&.67&.58\\
%% B1 %%
      \hline
\textbf{B$_1$}\cite{Ziegler2006}&.54&.59&.98&.88&.03&.59&.84&.22&.59&-.01&-.61&.53&.43&-.35&-.51&.67&.97&.69&-.34&.55\\
%% GM30 %%
      \hline
\textbf{GM30}\cite{Gracia2008}&.34&-&.47&-&.63&-&.58&-&.65&-&.31&-&.19&-&.34&-&.51&-&.27&-\\
%% MTurk %%
      \hline
\textbf{MTurk}\cite{Radinsky2011}&.40&.29&.32&-.32&.50&.35&.51&-.10&.49&.10&.57&.10&.32&-.16&.19&.37&.55&.22&.42&.37\\
%% Rel120 %%
      \hline
\textbf{Rel122}\cite{Szumlanski2013}&.66&-&.48&-&.56&-&.57&-&.59&-&.57&-&.32&-&.29&-&.53&-&.37&-\\
%% WRG %%
      \hline
\textbf{WRG}\cite{Agirre2009}&.26&1.0&.31&1.0&.42&1.0&.49&1.0&.39&1.0&.12&1.0&.23&1.0&.20&1.0&.52&1.0&.04&1.0\\
%% RG65 %%
      \hline
\textbf{RG65}\cite{Rubenstein1965}&.57&-&.63&-&.68&-&.66&-&.62&-&.57&-&.44&-&.52&-&.74&-&.75&-\\
%% MC30 %%
      \hline
\textbf{MC30}\cite{Miller1991}&.75&-&.70&-&.70&-&.45&-&.54&-&.65&-&.34&-&.63&-&.80&-&.64&-\\
%% KORE %%
      \hline
\textbf{KORE}\cite{Hoffart2012}&-&-&-&-&-&-&-&-&-&-&-&-&-&-&-&-&-&-&-&-\\
%% AVERAGE %%
      \hline
Avg.&.51&.65&.58&.64&.54&.\textbf{72}&.62&.50&.59&.36&.35&.65&.35&.29&.30&.66&\textbf{.68}&.68&.33&.59\\
\hline
    \end{tabular}
%%%%\end{table*}
%\end{center}
%%%%\end{landscape}

\vspace{2\baselineskip}
%%%%%%%%%%% TABLE 7 %%%%%%%%%%%%%
%%%%\begin{landscape}
%\begin{center}
%%%%\begin{table*}[ht]
\tiny
\caption{Means of the average Spearman's and Pearson's correlations for common nouns (\textit{c}) and proper nouns (\textit{p}) in disambiguated datasets and DBpedia with \textit{dbo:wikiPageWikiLinks} links}\label{tab:overall_common_proper}
\centering
    \begin{tabular} 
{p{1.5cm} 
p{1.2cm} 
p{1.4cm} 
p{1.2cm} 
p{1.6cm} 
p{1.4cm} 
p{1.0cm} 
p{1.6cm} 
p{1.2cm} 
p{1.6cm} 
p{1.4cm} }
    \hline
     & \textit{WLM} \cite{Witten2008} & 
     \textit{LODDO} \cite{Zhou2012} & 
     \textit{LDSD} \cite{Passant2010}  & 
     \textit{LDSDGN} \cite{Piao2016} & 
     \textit{PLDSD} \cite{Alfarhood2017} & 
     \textit{ICM} \cite{Schuhmacher2014} & 
     \textit{REWOrD} \cite{Pirro2012} & 
     \textit{ExclM} \cite{Hulpus2015}  & 
     \textit{ASRMP$_m$} \cite{ElVaigh2020} & 
     \textit{ProxM} \cite{Leal2013}\\ 

%% Common %%
      \hline
\textit{c} & .55 & \textbf{.65} & .61 & .60 & .55 & .48 & .37 & .46 & \textbf{.65} & .48\\
\hline
\textit{p} & .65 & .63 & .68 & .52 & .30 & .64 & .25 & \textbf{.69} & .65 & .61\\
\hline
    \end{tabular}
\end{sidewaystable}

%QUI

\subsubsection{Discussion}

On the basis of  the experimentation of this work, the \textit{ASRMP$_m$} method shows the best performance by considering disambiguated datasets and DBpedia with the \textit{dbo:wikiPageWikiLink} predicate. 
Indeed, a peculiarity of this method consists in taking into account {\it all the directed paths} connecting two resources, rather than selecting one or more of them according to some criteria (see Eq. \ref{ASRMP_a} in the Appendix). In fact, by summarizing, the \textit{WLM}  method, rather than addressing  paths, relies on the information gathered by the nodes that are adjacent to the compared resources. The same also holds in the case of \textit{LODDO} although, according to the assumptions made, it implicitly  considers all the paths of maximum length 2, both directed and undirected.  
With regard to the  methods based on triple patterns, namely     \textit{LDSD},  \textit{LDSDGN}, and  \textit{PLDSD}, they aim at verifying the existence  of specific configurations of paths, involving further resources in the graph on the basis of the names of the triples' predicates. Among these methods, only \textit{PLDSD} addresses all the paths, directed and undirected, between the compared resources.
  Finally,  among the methods based on triple weights, \textit{ICM}  focuses on the information contents of both the predicates and the objects of the triples and, analogously,  \textit{ProxM} that, according to the assumptions made in order to implement it, relies on the information contents of the triples' predicates. \textit{REWOrD} selects the most informative path among the ones connecting the compared resources, whereas \textit{ExclM} focuses on the top-k undirected paths between the compared resources.
Therefore,  aggregating all the directed  paths between resources is a distinctive feature of \textit{ASRMP$_m$} that contributes to make it the best strategy in order to compute  semantic relatedness
in the presence of  datasets containing both common nouns and proper nouns. 

In addition, overall \textit{ASRMP$_m$} and \textit{LODDO}  show the best  performances by considering only pairs of common nouns from disambiguated datasets, whereas \textit{ExclM} outperforms the other methods when addressing only pairs of proper nouns.

With regard to time complexity, as mentioned,  the motivation of this work is a comparison about the correlations of the methods with human judgment by running them all at once against the same  datasets, and on the same DBpedia release. For this reason, in this paper,
the complexity analysis of the methods has not been given  (when present, it can be found in the original papers where the methods have been proposed). About the running times, we ran the experimentation on a
machine with 32GB of RAM and the Intel® Core™ i7-8665U CPU @ 1.90GHz $\times$ 8 octa-core processor. 
In Table \ref{tab:runningtimes}, for each method, the worst running times needed in order  to compute the relatedness of a single pair of resources are shown. In particular, in the table the worst running times for a pair of common nouns and a pair of proper nouns are distinguished. Such times for common nouns are in general significantly less than the ones needed for evaluating the relatedness of proper nouns.  In fact, usually, in DBpedia nodes labeled with common nouns are involved in less triples than nodes representing proper nouns.

%\begin{center}
	\begin{table}\caption{Worst running times (in seconds) for a single pair of common (\textit{c}) and proper nouns (\textit{p})}
	\centering
   \footnotesize
		\begin{tabular}{p{2.5cm} p{1.2cm} p{1.2cm} }
		    \hline
            Method & \textit{c}  & \textit{p}\\ 
            \hline
		    \textit{WLM} \cite{Witten2008} & 0.6 & 1.1\\
		    \textit{LODDO} \cite{Zhou2012} & 0.4 & 0.5\\ 
		    \textit{LDSD} \cite{Passant2010}& 11.8 & 21.0\\
		    \textit{LDSDGN} \cite{Piao2016} & 5.9 & 9.0\\
		    \textit{PLDSD} \cite{Alfarhood2017} & 96.3 & 840.1\\
            \textit{ICM} \cite{Schuhmacher2014} & 7.3 & 8.7\\
            \textit{REWOrD} \cite{Pirro2012} &  6.3 & 17.5\\
            \textit{ExclM} \cite{Hulpus2015} & 9.9 & 78.1\\
            \textit{ASRMP$_m$} \cite{ElVaigh2020} & 0.4 & 1.5\\
            \textit{ProxM} \cite{Leal2013} & 9.2 & 12.6\\
			\hline
		\end{tabular}
		\label{tab:runningtimes}
	\end{table}
%\end{center}

	\begin{table}\caption{Pros and Cons of the methods where $d$ is the distance between the compared resources}
	\centering
   \footnotesize
		\begin{tabular}{p{2.0cm} p{6.0cm} p{6.0cm} }
		\hline
            Method & Pros & Cons\\ 
            \hline
		    \textit{WLM} & 
            - running time 
            
            - straightforward
            &  
             - local
             
             - predicates are not addressed
                                       
             - null relatedness values  if $d$ $>$ $2$\\
        \hline
		    \textit{LODDO} & 
            - running time 
            
            - straightforward 
            
            - the best overall correlation for common nouns &
            
            - local
      
             - predicates are not addressed
             
             - null relatedness values  if $d$ $>$ $2$\\ 
        \hline
		    \textit{LDSD} & - straightforward

            - more selectivity
            & 
            - null relatedness values  if $d$ $>$ $2$\\
        \hline
		    \textit{LDSDGN} &             
            - global 

            - more selectivity
            & 
            - non-straightforward
      
            - complex formalization
            
            - null relatedness values  if $d$ $>$ $2$\\
        \hline
		    \textit{PLDSD} &          
      
            -  non-null relatedness values for any $d$
            & 
            - running time 

            - less selectivity

            - the worst overall correlation for proper nouns
             \\
        \hline
            \textit{ICM} & 
            - global 

            - non-null relatedness values for any $d$
            & 
            - less selectivity\\
        \hline
            \textit{REWOrD} & - global

             - non-null relatedness values for any $d$
            & 
            - non-straightforward 
            
            - less selectivity
            
            - the worst overall correlation

            - the worst overall correlation for common nouns\\
        \hline
            \textit{ExclM} & 
            - straightforward

            - non-null relatedness values for any $d$

            - the best overall correlation for proper nouns
            & 
            - tuning parameters
            
            - less selectivity\\
        \hline
            \textit{ASRMP$_m$} & 
            - running time
            
            - more selectivity

            - non-null relatedness values  for any $d$

            - the best overall correlation

            - the best overall correlation for common nouns
            &

         - non-intuitive fuzzy logic  operators for triple and path aggregations  \\
        \hline
            \textit{ProxM} & 
            - global

            -  non-null relatedness values for any $d$
            & 
            - no built-in functions to weigh triples
            
            - less selectivity \\
			\hline
		\end{tabular}
		\label{tab:proscons}
	\end{table}

In Table \ref{tab:proscons}, pros and cons of the 10 methods are shown. In the table, $d$ is the distance between the compared resources, as defined according to the standard notion of shortest-path distance in graph theory.  For ``straightforward'' we mean that the approach is intuitive and easy to implement,  whereas ``complex formalization'' is related to the complexity of the underlying formulas (that are shown in the Appendix). ``Local''  means that the method focuses on the information provided by the nodes that are adjacent to the compared resources,  whereas ``global'' implies that the  information contained in the whole graph is addressed. Furthermore, ``more selectivity'' stands for methods defining some criteria in order to detect the kinds of paths to be considered (as for instance \textit{ASRMP$_m$},  which focuses on directed paths, or   \textit{LDSD} and  \textit{LDSDGN} that leverage specific path configurations), whereas ``less selectivity'' means that any path is  considered a priori. 
Overall, in the case of  \textit{ASRMP$_m$}, we  observe  an  imbalance in favor of pros with respect to cons,  taking into account in particular the running times, 
the focus on directed paths that is a proper characteristic of this approach, and the best overall correlations in general, and in the specific case of common nouns from disambiguated datasets.

\section{Conclusion}
\label{sec:conclusion}
Evaluating semantic relatedness of resources in RDF knowledge graphs is still a challenge. In this paper,  10 methods have been selected and experimented against 14 benchmark golden datasets by using DBpedia as reference RDF knowledge graph. The 10 approaches have been organized according to three representative groups, namely, the methods based on adjacent resources,  triple patters, and triple weights, and  their differences and commonalities have been highlighted.

The experimental results show that, first of all, the disambiguation of the dataset plays a fundamental role in evaluating semantic relatedness. Furthermore,  the triples with the \textit{dbo:wikiPageWikiLink} predicate represent a  significant integration to the information provided by the  resources' infoboxes, that contain partial summaries of the most important data  associated with the resources.
Finally, with regard to the methods, according to the experimental results, overall the strategy  relying on triple weights, when combined with the evaluation of {\it all} the directed paths connecting the compared resources, shows the best performances.

It is important to recall that in this experiment, when for a given method more than one strategy, or variant, is present,  we have considered the one that the authors identify as the {\it best} strategy in order to evaluate semantic relatedness. However, in some cases we realized that in our experiment the best variant for the authors does not correspond to the one associated with the best correlation values. For this reason, as a future work, we are planning to run a wider experimentation where  all the strategies of the methods are addressed (approximately 30 in total) and compared.

\section{Funding}
This research did not receive any specific grant from funding agencies in the public, commercial, or not-for-profit sectors.

\bibliographystyle{plainnat}  
\bibliography{elsarticle-template}

%\section*{References}

\newpage
%\pagebreak

%\section*{Appendix A. The 10 selected methods}
\large{
\textbf{Appendix}
}
\appendix
\section
{The 10 selected methods}
In this Appendix, the 10 selected methods are described in details, and are compared by using a running example  based on the graph $\mathcal{G}$ shown in Figure \ref{fig:master}. As mentioned above,  it contains 13 nodes (resources), linked with directed edges labeled with the  predicates $p_1$, $p_2$, $p_3$, and \textit{rdf:type}. For each method, $r_a$ and $r_b$ are the resources whose relatedness will be  addressed in order to highlight the specific characteristics of the different approaches.
\subsection{Methods based on adjacent resources}
\label{MethodsAdjacent_app}
%The focus on local knowledge,  with respect to the compared resources, makes these methods very efficient from a computational complexity point of view.   

In the following the methods based on adjacent resources are described.

\subsubsection{Wikipedia Link-based Measure (\textit{WLM})} \label{sec:WLM_2}

According to \cite{Witten2008}, consider an RDF graph $\mathcal G$, and the set $R$ of all the resources as defined according to the notation recalled in Section \ref{sec:rdf_linked_data}. Assume  $r_a$, $r_b$ $\in$ $R$, and let $A$, $B$ be the sets of the resources that are subjects of the triples with $r_a$ and $r_b$ as objects, respectively, i.e.:

$A = \{ r_j \in R  \vert  \exists p_i : \langle r_j,p_i,r_a \rangle \in \mathcal{G}\}$, $B = \{ r_j \in R  \vert  \exists p_i: \langle r_j,p_i,r_b \rangle \in \mathcal G\}$

\noindent According to the SPARQL notation introduced above, the sets $A$, and $B$ can also be rewritten as follows:

$ A = \{?x  \vert  \langle ?x, ?y, r_a \rangle \in \mathcal{G} \} $, $B = \{?x  \vert  \langle ?x, ?y, r_b \rangle \in \mathcal{G} \}$

\noindent The \textit{WLM} measure  between the resources $r_a$ and $r_b$, $WLM(r_a,r_b)$, is a distance rather than a relatedness measure since it
 originates from the  Normalized Google Distance, and is defined according to Eq. \ref{eq:WLM}: 

\begin{footnotesize}
\begin{equation}
\label{eq:WLM}
WLM(r_a,r_b)=  \frac{log(max( \vert A \vert , \vert B \vert ))-log( \vert A\cap B \vert )}{log( \vert R \vert )-log(min( \vert A \vert , \vert B \vert ))}
\end{equation}
\end{footnotesize}
 
\noindent where, for any set $S$,  $ \vert S \vert $ is the cardinality of $S$.  Note that, in the case both $r_a$ and $r_b$   never occur  as  object in any triple, $WLM(r_a,r_b)= \frac{\infty}{\infty}$ = 1 is assumed. Furthermore, if $r_a$ and $r_b$ are linked to the same resources or $r_a$ $\equiv$ $r_b$,  then $A \equiv B$, therefore their distance is null ($WLM(r_a,r_b)$ = 0).
In the case $WLM$ $\geq$ 1,
$r_a$ and $r_b$ are very unrelated and, in particular, if there are no resources linked to both $r_a$ and $r_b$, their distance is infinite ($WLM(r_a,r_b)$ = $\infty$), and they provide the minimum relatedness degree. 
Since $WLM$ ranges in the interval $[0, \dots, +\infty)$,  in this paper in order to experiment and compare  it  against the other methods,  the  relatedness formulation $\frac{1}{1+WLM}$ has been used.

\begin{figure}
\centering
{\includegraphics[width=.70\textwidth]{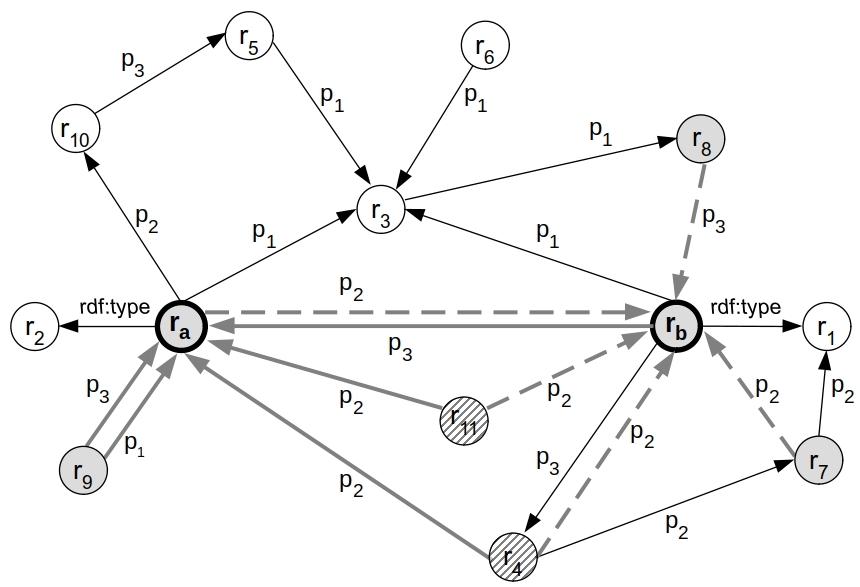}} 
\caption{\textit{WLM} applied to the running example graph}\label{fig:WLM}
\end{figure}

For instance, consider the nodes labeled with the resources $r_a$ and $r_b$ of the graph shown in Figure \ref{fig:master}. In order to evaluate the relatedness of $r_a$ and $r_b$ according to \textit{WLM}, only the nodes (resources)  with outgoing predicates towards $r_a$ and $r_b$ are considered, that are represented by the sets A = \{$r_4$, $r_9$, $r_{11}$, $r_b$\}, and B = \{$r_4$, $r_7$, $r_8$,  $r_{11}$, $r_a$\}, respectively. In Figure \ref{fig:WLM}, the related 
 links are  highlighted with  bold grey arrows, and  long dashed grey arrows, respectively. 
 In particular, the nodes $r_4$ and $r_{11}$, having outgoing predicates towards both $r_a$ and $r_b$ (the intersection of A and B) are filled with upward diagonals, whereas the nodes of all the other involved resources are in grey. Note that the resource $r_9$, although having two outgoing predicates towards $r_a$, appears only once because A is a set.

\subsubsection{Linked Open Data Description Overlap (\textit{LODDO})}
As mentioned in Section \ref{MethodsAdjacent}, the \textit{Linked Open Data Description Overlap} (\textit{LODDO}) method is based on the notion of resource's \textit{description} \cite{Zhou2012}.
Given a resource \textit{r}, the description $D$ of \textit{r},  $D(r)$, is the set of the resources\footnote{In \cite{Zhou2012} the authors state that the description of a resource is a vector without specifying if repetitions are allowed. In this paper, we assume that repetitions are not considered.} \textit{r$_i$} that are directly linked to \textit{r},  either via an incoming, or an outgoing predicate, plus the resource \textit{r} itself.  Furthermore, the method does not include in $D(r)$ all the resources that are linked to  $r$ via the predicate \textit{rdf:type}, exclusively. 
This is because, according to the authors of the method, given an RDF knowledge graph, almost every resource has \textit{owl:Thing} as type, and therefore the type assertions are considered ``noisy links'' that have to be ignored.
The description $D$  of the resource $r$, $D(r)$, can be formally defined according to the SPARQL notation  as follows:

\begin{footnotesize}
\begin{equation}
\label{eq:loddo_description}
\begin{aligned}
&D(r) = & \{?r_i \vert  \langle r, ?p_j, ?r_i \rangle \text{ $\in$ } \mathcal{G}, ?p_j \neq \textit{rdf:type} \} \cup  \\
& & \{?r_i \vert \langle ?r_i, ?p_j, r \rangle \text{ $\in$ } \mathcal{G}, ?p_j \neq \textit{rdf:type}\} \\
\end{aligned}
\end{equation}
\end{footnotesize}

\begin{figure}
\centering
{\includegraphics[width=.70\textwidth]{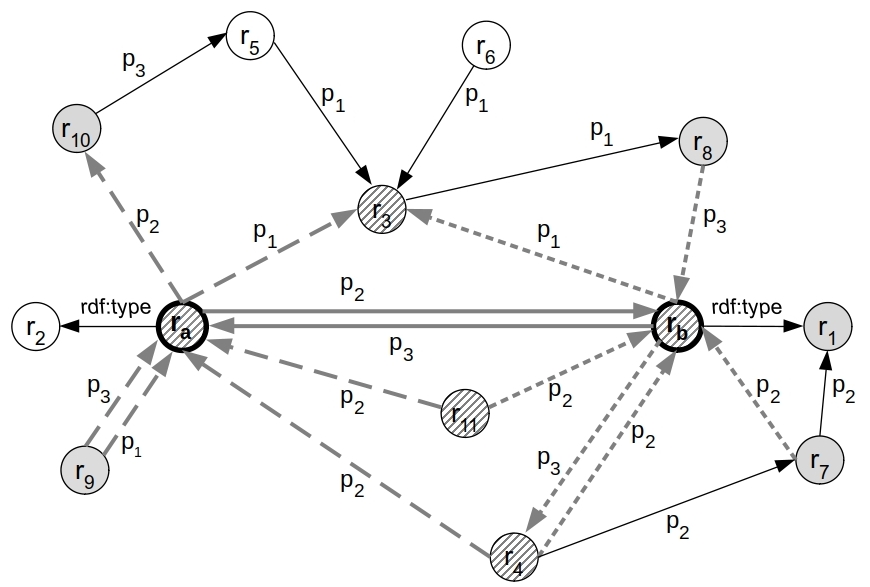}} 
\caption{\textit{LODDO} applied to the running example graph}\label{fig:LODDO}
\end{figure}

\noindent For instance, if we consider the resources $r_a$ and $r_b$ of our running example, $D(r_a) = \{r_a, r_b, r_3, r_4, r_9, r_{10}, r_{11}\}$, and $D(r_b) = \{r_b, r_a, r_3, r_4, r_7, r_8, r_{11}\}$. In Figure \ref{fig:LODDO}, the links that contribute to the descriptions of $r_a$ and $r_b$ are depicted as long dashed grey arrows and dashed grey arrows, respectively, whereas the links that contribute to the description of both the resources are in bold grey. 
Furthermore, the nodes in $D(r_a)$ and $D(r_b)$ have been highlighted.
It is worth noting that the resources $r_2$ and $r_1$ are not included in the above descriptions, since they are linked to $r_a$ and $r_b$ via the \textit{rdf:type} predicate only, respectively.

Given two resources, say $r_a$ and $r_b$, the following two strategies for computing the semantic relatedness between them are proposed, namely, \textit{LODJaccard} and \textit{LODOverlap}:

\vspace{0.15 cm}

\noindent \textbf{LODOverlap}.  The \textit{LODOverlap}  has a bias towards the resource, between the two, with a less rich description (Eq. \ref{eq:lodoverlap}),  where $min$ stands for the minimum cardinality between the descriptions of $r_a$ and $r_b$. 

\begin{footnotesize}
\begin{equation}
\label{eq:lodoverlap}
LODOverlap(r_a, r_b) = \frac{ \vert D(r_a) \cap D(r_b) \vert }{min\{ \vert D(r_a) \vert ,  \vert D(r_b) \vert \}}
\end{equation}
\end{footnotesize}

\noindent \textbf{LODJaccard}. $LODJaccard$  resembles the $Jaccard$ \textit{similarity coefficient}  \cite{Jaccard1912}, making no distinction between the two resources (Eq. \ref{eq:lodjaccard}). In fact, in place of the minimum, both the cardinalities of the descriptions are addressed. 

\begin{footnotesize}
\begin{equation}
\label{eq:lodjaccard}
LODJaccard(r_a, r_b) = \frac{ \vert D(r_a) \cap D(r_b) \vert }{ \vert D(r_a) \vert  +  \vert D(r_b) \vert  -  \vert D(r_a) \cap D(r_b) \vert }
\end{equation}
\end{footnotesize}

\noindent For both the strategies, the bigger the intersection between the descriptions of the compared resources the higher their semantic relatedness degree.

The method does not explicitly search for paths linking $r_a$ and $r_b$. However, the semantic relatedness between $r_a$ and $r_b$ is non-null  if there exists at least one undirected path of length 2 between them. In fact, only in this case, the intersection between $D(r_a)$ and $D(r_b)$ is non-empty.
In Figure \ref{fig:LODDO}, the intersection between the descriptions $D(r_a)$ and $D(r_b)$ is represented by the set $\{r_a, r_b, r_3, r_4, r_{11}\}$, whose corresponding nodes  are filled with upward diagonals.
In the case $r_a \equiv r_b$ the semantic relatedness between the resources is equal to $1$.
According to the experimentation given in \cite{Zhou2012}, the \textit{LODOverlap} strategy performs better than the \textit{LODJaccard} one.

\subsection{Methods based on triple patterns}
\label{Methods_patterns_app}
In this section, the second group of methods is addressed. They are based on the identification of \textit{path patterns} that satisfy specific criteria in the knowledge graph with respect to the compared resources.
Since the methods presented in this group represent distances that range in the interval $[0, \dots, 1]$, in this paper in order to experiment and compare them, if $v$ is the distance obtained according to one of these methods, we use the corresponding $1-v$ relatedness  formulation.\\

\subsubsection{Linked Data Semantic Distance (\textit{LDSD})\label{sec:Passant_2}}
In \cite{Passant2010}, the author presents a family of three measures for semantic distance named \textit{Linked Data Semantic Distance} (\textit{LDSD}). These measures are recalled below. 

\vspace{0.2 cm}
\noindent \textbf{\textit{LDSD}$_{dw}$}. The first measure is  the {\it direct weighted}  $LDSD$ distance, indicated as $LDSD_{dw}$,  which considers only the incoming and outgoing direct links between the resources to be compared. In particular, given a graph $\mathcal{G}$,
let $C_d$ be  a function that computes the number
of direct and distinct links between resources in the graph as follows. Given two resources $r_a$, $r_b$ and the predicate  $p_j$, $C_d(p_j,r_a,r_b)$ = 1 if there exists  a link labeled with  $p_j$ from the resource $r_a$ to the resource $r_b$, i.e., a triple $\langle r_a,p_j,r_b \rangle$, otherwise $C_d(p_j,r_a,r_b)$ = 0. Furthermore, 
%//// questo forse non serve piu  $C_d(r_a,r_b)$ /// era %$C_d(n,r_a,r_b)$/// computes the total number of direct and distinct links from $r_a$ to $r_b$,//// 
$C_d(p_j,r_a)$\footnote{In the original work \cite{Passant2010}, this function is defined as $C_d(p_j,r_a,n)$, where $n$ represents the  result of the function $C_d(p_j,r_a)$.} is the total
number of links  labeled with the predicate $p_j$ from $r_a$ to any node (i.e., the total number of resources that can be reached from $r_a$ via $p_j$). Therefore, given the resources $r_a$ and $r_b$, $LDSD_{dw}(r_a,r_b)$ is defined according to Eq. \ref{eq:LDSD_dw}:

\begin{footnotesize}
\begin{equation}
\label{eq:LDSD_dw}
LDSD_{dw}(r_a,r_b)=  \frac{1}{1+\sum_{p_j \in W}\frac{C_d(p_j,r_a,r_b)}{1+log(C_d(p_j,r_a))} + \sum_{p_j \in Z}\frac{C_d(p_j,r_b,r_a)}{1+log(C_d(p_j,r_b))}}
\end{equation}
\end{footnotesize}

\noindent where $W \subseteq R$  is the set of the predicates $p_j$ in the graph $\mathcal{G}$  such that $C_d(p_j,r_a,r_b) = 1$, and $Z \subseteq R$ is the set of the predicates $p_j$ in $\mathcal{G}$ such that $C_d(p_j,r_b,r_a) = 1$.\\

\begin{figure}
\centering
{\includegraphics[width=.70\textwidth]{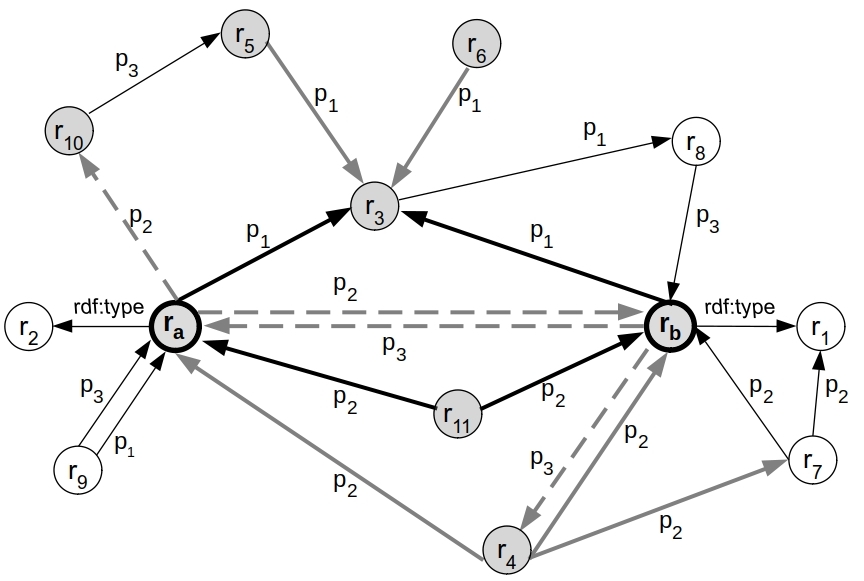}} 
\caption{\textit{LDSD} applied to the running example graph}\label{fig:passant}
\end{figure}

For instance, in the graph    of the running example,  the links  involved in the computation of  $LDSD_{dw}(r_a,r_b)$ are highlighted with long dashed grey arrows,  as shown in Figure \ref{fig:passant}.

\vspace{0.2 cm}
\noindent {\bf {\textit {LDSD}$_{iw}$}}. The second measure is the \textit{indirect weighted LDSD}, indicated as $LDSD_{iw}$. It basically considers all the path patterns in the graph identified by those resources linked  to both the compared resources via the same predicate.
Let  $C_{io}$ and  $C_{ii}$ be functions that compute the number of indirect and distinct links between resources,   outgoing and incoming respectively, as follows. Given two resources $r_a$, $r_b$ and a predicate  $p_j$,
$C_{io}(p_j,r_a,r_b)$ = 1 if there exists a resource $r_n$ that satisfies both $\langle r_a,p_j,r_n \rangle$, and $ \langle r_b,p_j,r_n \rangle$, otherwise $C_{io}(p_j,r_a,r_b)$ = 0.
Analogously, $C_{ii}(p_j,r_a,r_b)$ = 1  if there exists
a resource $r_n$ that satisfies both $\langle r_n,p_j,r_a \rangle$, and $\langle r_n,p_j,r_b \rangle$, otherwise $C_{ii}(p_j,r_a,r_b)$ = 0. 
Furthermore, let
$C_{io}(p_j,r_a)$  and $C_{ii}(p_j,r_a)$\footnote{Analogously to the function $C_d$, in \cite{Passant2010}, these two functions are defined as $C_{io}(p_j,r_a,n)$  and $C_{ii}(p_j,r_a,n)$, respectively.} 
be the total number of resources indirectly 
linked  to $r_a$ via the predicate $p_j$, outgoing and incoming respectively.
Hence, given the resources $r_a$ and $r_b$,  $LDSD_{iw}(r_a,r_b)$  addresses the indirect incoming and outgoing links between the resources, and is defined according to Eq. \ref{eq:LDSD_iw}.

\begin{footnotesize}
\begin{equation}
\label{eq:LDSD_iw}
LDSD_{iw}(r_a,r_b)=  \frac{1}{1+\sum_{p_j \in U}\frac{C_{io}(p_j,r_a,r_b)}{1+log(C_{io}(p_j,r_a))} + 
\sum_{p_j \in V}\frac{C_{ii}(p_j,r_a,r_b)}{1+log(C_{ii}(p_j,r_a))}}
\end{equation}
\end{footnotesize}

\noindent where $U \subseteq R$ is the set of the predicates $p_j$ in the graph $\mathcal{G}$ such that $C_{io}(p_j,r_a,r_b) = 1$, and $V \subseteq R$ is the set of the predicates $p_j$ in $\mathcal{G}$ such that $C_{ii}(p_j,r_a,r_b) = 1$.\\

\noindent In the example of Figure \ref{fig:master},  the resource $r_3$, with incoming links labeled with the predicate $p_1$ (one outgoing from $r_a$ and the other one from $r_b$),  and both the resources $r_4$ and $r_{11}$, with   outgoing links labeled with the predicate $p_2$ (and incoming to $r_a$ and $r_b$, accordingly),  satisfy the above conditions for $C_{io}(p_1,r_a,r_b)$ and
$C_{ii}(p_2,r_a,r_b)$ respectively, therefore $C_{io}(p_1,r_a,r_b)$  = 1 and
$C_{ii}(p_2,r_a,r_b)$ = 1. Note that only one resource with indirect incoming links is needed in order to have  $C_{ii}(p_2,r_a,r_b)$ = 1, for instance $r_{11}$. For this reason, in order to highlight this point, in Figure \ref{fig:passant}, besides the indirect outgoing links related to $r_3$, only the   indirect incoming links related to $r_{11}$ have been drawn in bold.  

\vspace{0.2 cm}

\noindent {\bf {\textit {LDSD}$_{cw}$}}.
Finally, the author proposes the {\it combined weighted}  $LDSD$ distance between the resources $r_a$ and $r_b$, indicated as $LDSD_{cw}(r_a,r_b)$, that is a combination of the previous distances, the direct and indirect ones, defined as follows:

\begin{footnotesize}
\begin{equation}
\label{eq:LDSD_cw}
LDSD_{cw}(r_a,r_b)= \frac{1}{1+ f_1 + f_2}\\
\end{equation}
\end{footnotesize}

\noindent where:\\

\begin{footnotesize}
$ f_1 = \sum_{p_j \in W}\frac{C_d(p_j,r_a,r_b)}{1+log(C_d(p_j,r_a))} +
\sum_{p_j \in Z}\frac{C_d(p_j,r_b,r_a)}{1+log(C_d(p_j,r_b))}$\\

$f_2 = \sum_{p_j \in U}\frac{C_{io}(p_j,r_a,r_b)}{1+log(C_{io}(p_j,r_a))} +
\sum_{p_j \in V}\frac{C_{ii}(p_j,r_a,r_b)}{1+log(C_{ii}(p_j,r_a))}$\\
\end{footnotesize}

\noindent and 
$W \subseteq R$  is the set of the predicates $p_j$ in the graph $\mathcal{G}$  such that $C_d(p_j,r_a,r_b) = 1$,
$Z \subseteq R$ is the set of the predicates $p_j$ in $\mathcal{G}$ such that $C_d(p_j,r_b,r_a) = 1$, 
$U \subseteq R$ is the set of the predicates $p_j$ in $\mathcal{G}$ such that $C_{io}(p_j,r_a,r_b) = 1$, 
and $V \subseteq R$ is the set of the predicates $p_j$ in $\mathcal{G}$ such that $C_{ii}(p_j,r_a,r_b) = 1$.\\

In our running example, the $LDSD_{cw}$ measure involves all the links and the nodes highlighted in Figure \ref{fig:passant}.
Note that the distance defined according to  Eq. \ref{eq:LDSD_cw} is not symmetric.
%, in fact:
%$LDSD_{cw}(r_b,r_a)$ = ...\\
%///se venisse lo stesso risultato non c'e' bisogno di scrivere questo sopra tanto e' evidente che la formula cw non e' simmetrica //
This point is addressed by the measure recalled in the next subsection.
According to the results given in \cite{Passant2010}, the $LDSD_{cw}$ measure performs better than the $LDSD_{dw}$ and the $LDSD_{iw}$ measures.

\subsubsection{LDSD with Global Normalization (\textit{LDSDGN})} \label{PiaoBreslin2016_2}

The \textit{LDSD with Global Normalization} (\textit{LDSDGN}) measure \cite{Piao2016} is an evolution of the approach presented by Passant \cite{Passant2010}, where the normalization addresses both the compared resources and the global appearances of specific path patterns in the graph. In a previous work \cite{Piao2015}, the authors claim that, given  two resources $r_a$ and $r_b$, with $r_a$ $\neq$ $r_b$, a distance measure $d$ should satisfy the following three axioms:

 (i) Equal self-distance, i.e., 
{\it d}$(r_a,r_a)$ = {\it d}$(r_b,r_b) = 0$.

 (ii) Symmetry, i.e., {\it d}$(r_a,r_b)$ = {\it d}$(r_b,r_a)$.

 (iii) Minimality, i.e., {\it d}$(r_a,r_a)$ $<$ {\it d}$(r_a,r_b)$. 

\noindent Hence, they put in evidence that all the measures introduced by Passant do not satisfy both the axioms (i) and (iii), and this is because the distance between any resource and itself depends on its incoming and outgoing links. Furthermore, the $LDSD_{cw}$ measure does not even satisfy the symmetry axiom because  Eq. \ref{eq:LDSD_cw} addresses only the total number of resources indirectly linked to $r_a$, whereas the ones linked to $r_b$ are not considered. 
For this reasons,  in order to meet the mentioned requirements,  in \cite{Piao2016}  the authors propose a family of $LDSD$ measures satisfying the three axioms above.
In particular, in the following, the distances the $LDSD_\alpha$, $LDSD_\beta$ and $LDSD_\gamma$ are recalled.

\vspace{0.2 cm}

\noindent {\bf LDSD$_\alpha$}.
Given a graph $\mathcal{G}$, analogously to the notation used by  Passant in \cite{Passant2010}, below $C_d(p_j,r_a,r_b)$ = 1 if in the graph there exists a triple $\langle r_a,p_j,r_b \rangle$, otherwise $C_d(p_j,r_a,r_b)$ = 0, and  the total number of resources that can be reached from $r_a$ by means of  the predicate $p_j$ is indicated by $C_d(p_j,r_a)$ (analogously, $C_d(p_j,r_b)$).
Similarly, $C_{io}(p_j,r_a)$ and $C_{ii}(p_j,r_a)$ are the
total number of resources indirectly linked to $r_a$ via outgoing and incoming links labeled with the predicate $p_j$, respectively.
Furthermore, on the basis of the assumption that resources are more related if there is a great number of them linked via a given predicate $p_k$,  the  $C_{io}$ ($C_{ii}$) function defined  by Passant has been generalized by using the function  $C'_{io}$ ($C'_{ii}$) as follows: $C'_{io}(p_k,r_a,r_b)$ ($C'_{ii}(p_k,r_a,r_b)$) computes the total number of 
resources linked to $r_a$ and $r_b$ via an outgoing (incoming)
predicate $p_k$. Therefore, given the resources $r_a$, $r_b$, the first distance is the $LDSD_\alpha(r_a,r_b)$ measure defined in Eq. \ref{eq:LDSDalpha}.

\begin{footnotesize}
\begin{equation}
\label{eq:LDSDalpha}
LDSD_\alpha(r_a,r_b)= \frac{1}{1+ f_1 + f_2}\\
\end{equation}
\end{footnotesize}

\noindent where:\\

\begin{footnotesize}
$f_1 = \sum_{p_j \in U}\frac{C_d(p_j,r_a,r_b)}{1+log(C_d(p_j,r_a))} +
\sum_{p_j \in V}\frac{C_d(p_j,r_b,r_a)}{1+log(C_d(p_j,r_b))}\\$

$f_2 = \sum_{p_j \in W}\frac{C'_{io}(p_j,r_a,r_b)}{1+log(C_{io}(p_j,r_a))} +
\sum_{p_j \in Z}\frac{C'_{ii}(p_j,r_a,r_b)}{1+log(C_{ii}(p_j,r_a))}$\\
\end{footnotesize}

\noindent and 
$U \subseteq R$ is the set of the predicates $p_j$ in the graph $\mathcal{G}$ such that $C_{d}(p_j,r_a,r_b) = 1$,
$V \subseteq R$ is the set of the predicates $p_j$ in $\mathcal{G}$ such that $C_{d}(p_j,r_b,r_a) = 1$,
$W \subseteq R$ is the set of the predicates $p_j$ in $\mathcal{G}$   such that $C'_{io}(p_j,r_a,r_b) > 0$, and
$Z \subseteq R$ is the set of the predicates $p_j$ in $\mathcal{G}$   such that $C'_{ii}(p_j,r_a,r_b) > 0$.\\

\begin{figure}
\centering
{\includegraphics[width=.70\textwidth]{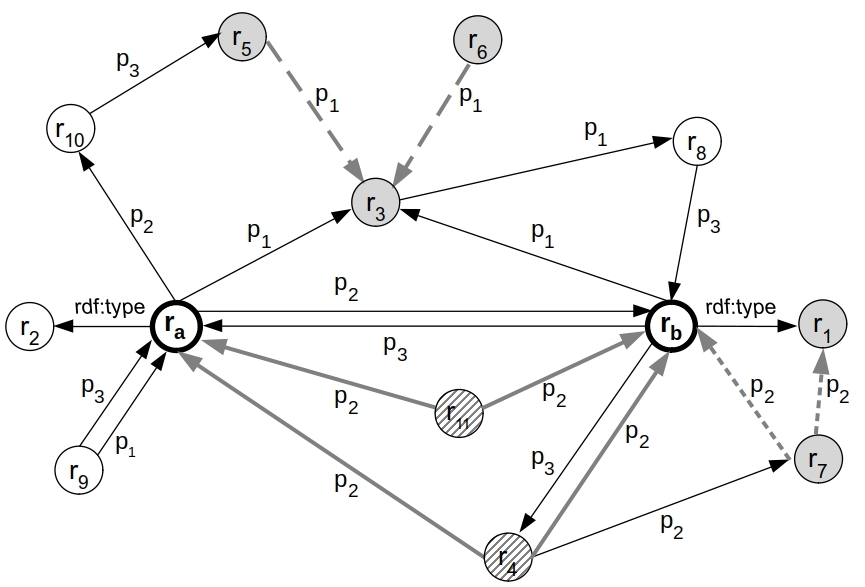}} 
\caption{\textit{LDSDGN} applied to the running example graph}\label{fig:piao}
\end{figure}

\noindent In the example of Figure \ref{fig:master},  $C'_{ii}(p_2,r_a,r_b)$ = 2 since 
there are two resources, namely $r_4$ and $r_{11}$, both with  incoming links to $r_a$ and $r_b$ labeled with the predicate $p_2$. In Figure \ref{fig:piao} these resources are filled with  upward diagonals, and the involved links are highlighted in grey.

\vspace{0.2 cm}

\noindent {\bf LDSD$_\beta$}. With respect to  $LDSD_\alpha$, in the  $LDSD_\beta$ measure   the last two addenda at the denominator are modified by addressing the averages between  $C_{io}(p_j,r_a)$, $C_{io}(p_j,r_b)$, and  $C_{ii}(p_j,r_a)$, $C_{ii}(p_j,r_b)$ respectively, in order to achieve symmetry. In particular, given the  resources $r_a$, and $r_b$, $LDSD_\beta(r_a,r_b)$ is defined according to Eq. \ref{eq:LDSDbeta}.

\begin{footnotesize}
\begin{equation}
\label{eq:LDSDbeta}
LDSD_\beta(r_a,r_b) = \frac{1}{1+ f_1 + f_2}
\end{equation}
\end{footnotesize}

\noindent where:\\

\begin{footnotesize}
$f_1 = \sum_{p_j \in U}\frac{C_d(p_j,r_a,r_b)}{1+log(C_d(p_j,r_a))} +
\sum_{p_j \in V}\frac{C_d(p_j,r_b,r_a)}{1+log(C_d(p_j,r_b))}$\\

$f_2 = \sum_{p_j \in W}\frac{C'_{io}(p_j,r_a,r_b)}{1+log(\frac{C_{io}(p_j,r_a)+C_{io}(p_j,r_b)}{2})} +
\sum_{p_j \in Z}\frac{C'_{ii}(p_j,r_a,r_b)}{1+log(\frac{C_{ii}(p_j,r_a)+C_{ii}(p_j,r_b)}{2})}$\\
\end{footnotesize}

\noindent and 
$U \subseteq R$ is the set of the predicates $p_j$ in the graph $\mathcal{G}$ such that $C_{d}(p_j,r_a,r_b) = 1$,
$V \subseteq R$ is the set of the predicates $p_j$ in  $\mathcal{G}$ such that $C_{d}(p_j,r_b,r_a) = 1$,
$W \subseteq R$ is the set of the predicates $p_j$ in $\mathcal{G}$   such that $C'_{io}(p_j,r_a,r_b) > 0$, and
$Z \subseteq R$ is the set of the predicates $p_j$ in $\mathcal{G}$   such that $C'_{ii}(p_j,r_a,r_b) > 0$.\\

\noindent Therefore, in  the $LDSD_\beta(r_a,r_b)$ distance further links are captured by the function $C_{ii}(p_2,r_b)$, that  are highlighted with dashed grey arrows in Figure \ref{fig:piao}, with the nodes of the involved resources, $r_7$ and $r_1$,  in grey.

\vspace{0.2 cm}
\noindent {\bf LDSD$_\gamma$}. In \cite{Piao2016} the authors state that all the above recalled  measures, including Eq. \ref{eq:LDSD_cw} of Passant,  involve a {\it local normalization} that takes into account the  paths ``in the context'' of the resources. For this reason, in the mentioned paper, the authors propose a further measure, namely $LDSD_\gamma$, relying on a {\it global normalization} notion that essentially considers the importance of a path between
two resources according to the number of its occurrences in the whole graph $\mathcal{G}$. In the following, 
let $C_{dp}(p_j)$ be
the global occurrences of the link $p_j$ between two resources in $\mathcal{G}$.
Furthermore, 
$C_{io}(p_k,r_j,r_a,r_b)$ = 1 if there exists a resource $r_j$ such that $\langle r_a,p_k,r_j \rangle$ and $\langle r_b,p_k,r_j \rangle$, and
$C_{ii}(p_k,r_j,r_a,r_b)$ = 1 if there exists a resource $r_j$ such that $\langle r_j,p_k,r_a \rangle$ and $\langle r_j,p_k,r_b \rangle$.

The normalizations of $C_{io}(p_k,r_j,r_a,r_b)$ and  $C_{ii}(p_k,r_j,r_a,r_b)$  are carried out by using the global occurrences $C_{iop}(p_k,r_j)$ and $C_{iip}(p_k,r_j)$ of $r_j$ as follows.
$C_{iop}(p_k,r_j)$  returns the global occurrences of  $r_j$ in the undirected paths $[\langle r_n,p_k,r_j \rangle$, $\langle r_s,p_k,r_j \rangle$],  for any  resources $r_n$, $r_s$ in the graph $\mathcal{G}$ and, analogously, 
$C_{iip}(p_k,r_j)$  computes the global occurrences of  $r_j$ in the undirected paths $[\langle r_j,p_k,r_n \rangle$, $\langle r_j,p_k,r_s \rangle]$,  for any  resources $r_n$, $r_s$ in $\mathcal{G}$.   

According to the above assumptions, given $r_a$ and $r_b$,  $LDSD_\gamma(r_a,r_b)$ is  defined in Eq. \ref{eq:LDSDgamma}.

\begin{footnotesize}
\begin{equation}
\label{eq:LDSDgamma}
LDSD_\gamma(r_a,r_b)= \frac{1}{1+ f_1 + f_2}
\end{equation}
\end{footnotesize}

\noindent where:\\

\begin{footnotesize}
$f_1 = \sum_{p_j \in U}\frac{C_d(p_j,r_a,r_b)}{1+log(C_{dp}(p_j))} +
\sum_{p_j \in V}\frac{C_d(p_j,r_b,r_a)}{1+log(C_{dp}(p_j))}$\\

$ f_2 = \sum_{(p_k, r_j) \in W}\frac{C_{io}(p_k,r_j,r_a,r_b)}{1+log(C_{iop}(p_k,r_j))} + 
\sum_{(p_k, r_j) \in Z}\frac{C_{ii}(p_k,r_j,r_a,r_b)}{1+log(C_{iip}(p_k,r_j))}$\\

\end{footnotesize}

\noindent and 
$U \subseteq R$ is the set of the predicates $p_j$ in the graph $\mathcal{G}$ such that $C_{d}(p_j,r_a,r_b) = 1$,
$V \subseteq R$ is the set of the predicates $p_j$ in $\mathcal{G}$ such that $C_{d}(p_j,r_b,r_a) = 1$,
$W \subseteq R\times R$ is the set of the pairs ($p_k$, $r_j$) such that $C_{io}(p_k, r_j, r_a, r_b) = 1$, and
$Z \subseteq R\times R$ is the set of the pairs ($p_k$, $r_j$)  such that $C_{ii}(p_k, r_j, r_a, r_b) = 1$.\\

%\begin{equation}
%LDSD_\gamma(r_a,r_b)=  \frac{1}{1+g_1(r_a,r_b) + g_2(r_a,r_b) + g_3(r_a,r_b) + g_4(r_a,r_b)}
%\end{equation}

%\noindent where:\\

%$g_1(r_a,r_b)=\sum_j\frac{C_d(p_j,r_a,r_b)}{1+log(C_{dp}(p_j))}$

%$g_2(r_a,r_b)=\sum_j\frac{C_d(p_j,r_b,r_a)}{1+log(C_{dp}(p_j))}$

%$g_3(r_a,r_b)=\sum_k\sum_j\frac{C_{io}(p_k,r_j,r_a,r_b)}{1+log(C_{iop}(p_k,r_j))}$

%$g_4(r_a,r_b)=\sum_k\sum_j\frac{C_{ii}(p_k,r_j,r_a,r_b)}{1+log(C_{iip}(p_k,r_j))}$ \\

\noindent For instance, in our running example, assume $k$ = 1, and $j$ = 3. Then, $C_{io}(p_1,r_3,r_a,r_b)$  is equal to 1 because there exists the resource $r_3$, and $\langle r_a,p_1,r_3 \rangle$ and $\langle r_b,p_1,r_3 \rangle$ are triples belonging to the graph. It is normalized according to $C_{iop}(p_1,r_3)$, that returns the global occurrences of the resource $r_3$ in the graph, identified by the links $\langle r_5,p_1,r_3 \rangle$, $\langle r_6,p_1,r_3 \rangle$,  highlighted with
long dashed grey arrows in Figure \ref{fig:piao}.

%In order to quantify the relatedness of the resources $r_a$, $r_b$ of the graph of Figure \ref{fig:master}, the following holds:

%$LDSD_\alpha(r_a,r_b)$ = 0.74

%$LDSD_\beta(r_a,r_b)$ = 0.73

%$LDSD_\gamma(r_a,r_b)$ = 0.72 //metterei solo questo//

According to the results presented in \cite{Piao2016},  $LDSD_{\gamma}$  performs better than the $LDSD_{\alpha}$ and  $LDSD_{\beta}$ measures.

\subsubsection{Propagated Linked Data Semantic Distance (\textit{PLDSD})} 

The \textit{Propagated Linked Data Semantic Distance} ($PLDSD$) \cite{Alfarhood2017} allows 
 the evaluation of the relatedness of two resources
by considering the distance computed according to $LDSD_{cw}$ (see Eq. \ref{eq:LDSD_cw}) between the adjacent resources in all the paths linking the compared resources, up to a given length. As a result, with respect to the  $LDSD_{cw}$ measure, in this approach additional pairs of resources are considered in the semantic relatedness evaluation.
Note that the aforementioned paper focuses on recommendation systems and, in order to face with efficiency problems,  the authors reduce the knowledge graph by addressing only  the resources identified by the  system in the given domain.

Given the resources $r_a$ and $r_b$ in a knowledge graph $\mathcal G$, and $h>0$ that is the maximum length of the paths to be considered, the semantic relatedness $PLDSD_h(r_a, r_b)$  can be summarized according to Eq. \ref{eq:pldsd}:

\begin{footnotesize}
\begin{equation}
\label{eq:pldsd}
PLDSD_h(r_a, r_b) = \underset{P \in \mathcal{P}^h}{max}\prod_{i=1}^{length(P)} (1 - LDSD_{cw}(s_i, o_i)) 
\end{equation}
\end{footnotesize}

\noindent where:
\begin{itemize}
    \item 
$\mathcal{P}^h$ is the set of  the undirected paths $P$ connecting $r_a$ and $r_b$ with length less than or equal to $h$.
\item $s_i$ and $o_i$ are the subject and the object, respectively, of the \textit{i-th} triple in the path $P$.
\item $length(P)$ is the length of the path $P$.
\end{itemize}

\begin{figure}
\centering
{\includegraphics[width=.70\textwidth]{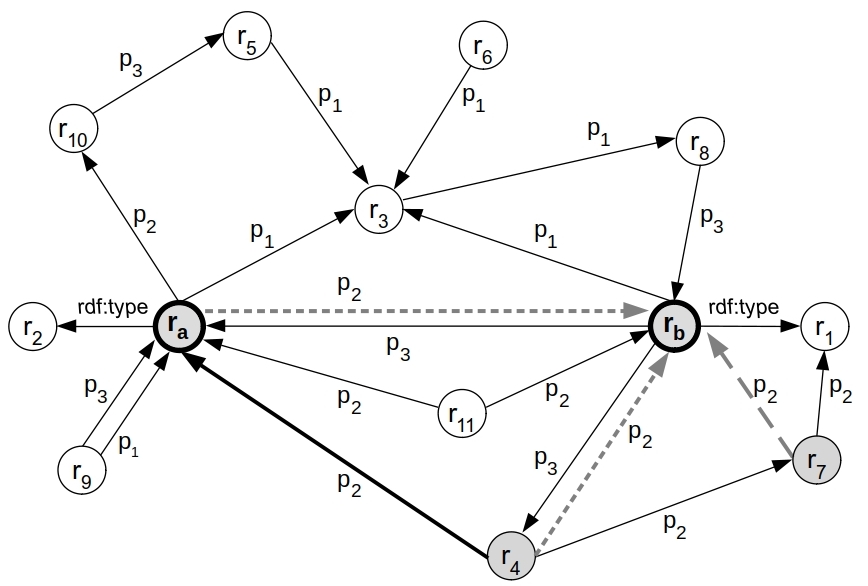}} 
\caption{\textit{PLDSD} applied to the running example graph}\label{fig:pldsd}
\end{figure}

\noindent According to the above formula, it is possible to see the reason why, with respect to the  $LDSD_{cw}$, additional resources are considered in the  relatedness evaluation. For instance, consider the resources $r_a$ and $r_b$ of the running example of Figure \ref{fig:master}. In Figure \ref{fig:passant} we have seen that the resource  $r_4$ has been considered in the evaluation of the  $LDSD_{cw}(r_a,r_b)$ since it has  outgoing links towards both $r_a$ and $r_b$ labeled with the predicate $p_2$, but the resource $r_7$, for instance, is not involved in the computation. This is not the case of the $PLDSD$ approach, where also  $r_7$  is addressed. In fact, when considering the link highlighted in bold in Figure \ref{fig:pldsd}, corresponding to the triple $\langle r_4, p_2, r_a\rangle$ of the path [$\langle r_4, p_2, r_a\rangle$, $\langle r_4, p_2, r_b\rangle$],   the $LDSD_{cw}$ applied to the  pair $(r_4, r_a)$ (whose resources are indirectly linked via $p_2$, as shown by the dashed grey arrows drawn in Figure \ref{fig:pldsd}) involves $r_7$ since it is indirectly connected to $r_4$ via the predicate $p_2$, as highlighted by the additional long dashed grey arrow of Figure  \ref{fig:pldsd}.

\subsection{Methods based on triple weights}
\label{Methods_weights_app}
In this subsection the methods belonging to the third group are described, which require the association of weights with triples in order to evaluate the contributions of the different paths.

\subsubsection{Information Content-based Measure (\textit{ICM})}
The \textit{Information Content-based Measure} (\textit{ICM}) requires the evaluation of  the weights of the triples belonging to the undirected paths linking the  compared resources, up to a given length \cite{Schuhmacher2014}.
Such a weight is computed on the basis of the \textit{information content} notion recalled below.

Given a random variable $X$ in the set $\{x_i\}$, and a probability distribution $Pr(X)$ over $X$, the information content (\textit{IC}) associated with $X=x_i$, i.e., the event that the variable $X$ assumes the value $x_i$, is defined in  Eq. \ref{eq:IC}:

\begin{footnotesize}
\begin{equation}
\label{eq:IC}
IC_{Pr(X)}(X=x_i)=-log(Pr(X=x_i))
\end{equation}
\end{footnotesize}

\noindent  that can also be written as $IC(x_i) = -log(Pr(x_i))$ for short.
Hence, according to the $IC$ notion,  specificity is a good proxy for relevance, and the less the probability of an event, the higher its information content.

If the random variable $X$ describes only the predicate of a triple, the weight of the triple depends only on the probability associated with that predicate. Consequently, two triples with   the same predicate have the same weight.
However, it can be intuitively assessed that two triples having the same predicate, but different objects, in general, convey different amounts of information. This is the case of the following two triples extracted from DBpedia, representing two statements about the resource \textit{dbr:Dante\_Alighieri}:

$\langle$\textit{dbr:Dante\_Alighieri}, \textit{rdf:type}, \textit{dbo:Person}$\rangle$

$\langle$\textit{dbr:Dante\_Alighieri}, \textit{rdf:type}, \textit{dbo:Writer}$\rangle$

They both have the same predicate, i.e., \textit{rdf:type}, but since writer is a term more specific than person, the latter  represents a more accurate and richer piece of information.
 For this reason, in order to compute the weight of a triple, both the predicate and the object of the triple are considered, and in the following we assume they are described by the random variables $X$ and $Y$, respectively.

In \cite{Schuhmacher2014}, the authors propose the following three strategies for computing the weight of a triple $t = \langle r_s, p, r_o \rangle$. 

\begin{figure}
\centering
{\includegraphics[width=.70\textwidth]{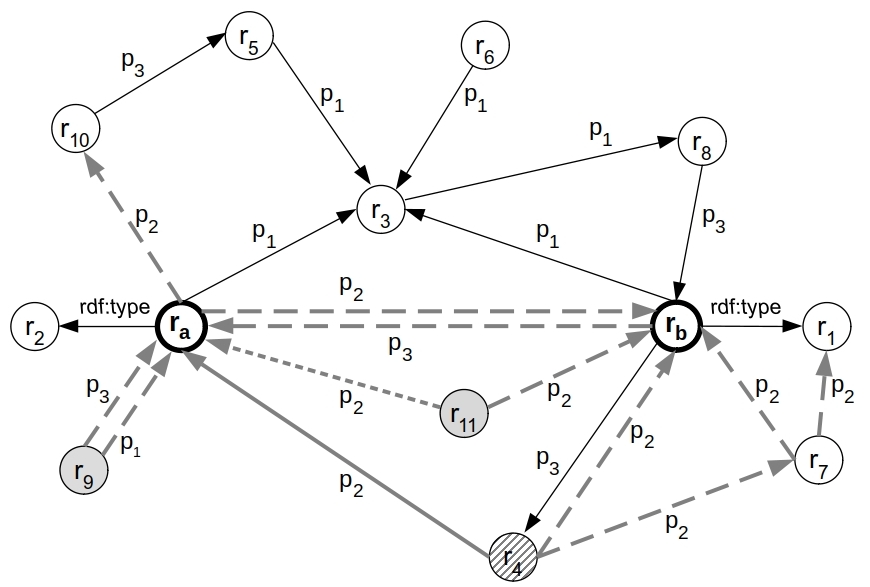}} 
\caption{\textit{ICM} applied to the running example graph}\label{fig:IC_JOINTandCOMB}
\end{figure}

\vspace{0.2 cm}
\noindent \textbf{Joint Information Content (jointIC)}. 
In the case of the {\it jointIC} strategy, the weight of the triple $t$, $w_{jointIC}(t)$, is computed according to Eq. \ref{eq:ic_joint}:  

\begin{footnotesize}
\begin{equation}
\label{eq:ic_joint}
w_{jointIC}(t) = IC(p) + IC(r_o \vert p)
\end{equation}
\end{footnotesize}

\noindent where 
$ IC(p) = IC_{Pr(X)}(X=p)$ is the information content associated with probability that the random variable $X$ assumes the value $p$,
and 
$ IC(r_o \vert p) = IC_{Pr(Y), Pr(X)}(Y=r_o \vert X=p)$
is the information content associated with the conditional probability that the variable $Y$ assumes the value $r_o$, supposing that the variable $X$ assumes the value $p$.
Note that, $w_{jointIC}(t)$ is equivalent to the $IC$  of the joint probability $Pr(p, r_o)$\footnote{$IC(p) + IC(r_o \vert p) = IC(Pr(X=p)) + IC(Pr(Y=r_o \vert X=p)) = -log(Pr(X=p)) -log(Pr(Y=r_o \vert X=p)) = 
-log(Pr(X=p)Pr(Y=r_o \vert X=p)) = -log(Pr(X=p), Pr(Y=r_o)) = IC(Pr(X=p), Pr(Y=r_o)) = IC(Pr(p, r_o)) = IC(p, r_o)$.}, that is the probability that the variables $X$ and $Y$ assume the values $p$ and $r_o$, respectively, and represents the likelihood of randomly selecting, from the considered RDF graph, a triple with $p$ and $r_o$ as predicate and object, respectively. Therefore, Eq. \ref{eq:ic_joint} can be also written as $w_{jointIC}(t) = IC(p, r_o) = IC_{Pr(Y), Pr(X)}(X=p, Y=r_o)$, emphasizing that the triples that contribute to the computation of  {\it jointIC} are those with predicate $p$ and object $r_o$.   
For instance, if we consider the triple $\langle r_4, p_2, r_a\rangle$ of the running example, which has been represented with a grey arrow in the graph of Figure \ref{fig:IC_JOINTandCOMB},  the  edges  relevant to compute {\it jointIC}($\langle r_4, p_2, r_a\rangle$) are the triple itself, and the triple $\langle r_{11}, p_2, r_a\rangle$,  which has been highlighted with a dashed  grey arrow in the same figure, because in the graph there are no other  triples with predicates $p_2$ and object $r_a$.

\vspace{0.2 cm}
\noindent \textbf{Combined Information Content (combIC)}. The {\it combIC} strategy aims at mitigating the possible penalization of the {\it jointIC} measure, in the case of infrequent objects that occur with infrequent predicates, as shown  by Eq (\ref{eq:ic_comb}):

\begin{footnotesize}
\begin{equation}
\label{eq:ic_comb}
w_{combIC}(t) = IC(p) + IC(r_o)
\end{equation}
\end{footnotesize}

\noindent where, with respect to Eq. \ref{eq:ic_joint}, $IC(r_o)$ is evaluated independently of the predicate $p$.  In fact,  the   \textit{combIC} approach is applied while making an independence assumption between the predicate and the object. Consequently, the weight of the triple $t$ results in the sum of the \textit{ICs} of the predicate and the object.
If we consider again the triple $\langle r_4, p_2, r_a\rangle$ in the graph of Figure \ref{fig:IC_JOINTandCOMB}, according to the {\it combIC} strategy, additional triples have to be considered with respect to {\it jointIC}, that have been highlighted with long dashed grey arcs in the same figure, i.e.,  all the triples having either $p_2$ as  predicate or $r_a$
as the object.

\vspace{0.2 cm}
\noindent  \textbf{Information Content and Pointwise Mutual Information (IC+PMI)}. According to the {\it IC+PMI} strategy, the weight of the triple $t$ can be defined by  Eq. \ref{eq:ic_icpluspmi}:

\begin{footnotesize}
\begin{equation}
\label{eq:ic_icpluspmi}
w_{IC+PMI}(t) = IC(p) + PMI(p, r_o)
\end{equation}
\end{footnotesize}

\noindent where:

\begin{footnotesize}
\begin{equation}
\label{eq:ic_pmi}
PMI(p, r_o) = log \frac{Pr(p, r_o)}{Pr(p) Pr(r_o)}
\end{equation}
\end{footnotesize}

\noindent In particular, $PMI$ measures the mutual dependence between the two random variables describing the predicate and the object of a triple, and can be seen as a measure of the deviation from independence between the two outcomes. With respect to the previous strategies, by means of the addendum \textit{PMI}, {\it IC+PMI} represents a balance between the assumptions of full dependence ({\it jointIC}) and independence ({\it combIC}) between predicates and objects. 
In Eq. \ref{eq:ic_icpluspmi}, the $IC$ of the predicate is summed  with $PMI$ in order to bias the weight towards less frequent, and thus more informative, predicates.

Once a strategy has been adopted, given the resources $r_a$ and $r_b$, 
%with $r_a \neq r_b$, 
 and  the maximum length $h>0$ of the paths to be considered, the semantic relatedness between $r_a$ and $r_b$, \textit{ICM}$_{h}(r_a, r_b)$, is computed according to Eq. \ref{eq:ic_relatedness}:

\begin{footnotesize}
\begin{equation}
\label{eq:ic_relatedness}
\textit{ICM}_{h}(r_a, r_b) = \frac{1}{\underset{P \in \mathcal{P}^h}{min} \underset{t_i \in P}{\sum} (w_{max} - w(t_i))} 
\end{equation}
\end{footnotesize}

\noindent where:
\begin{itemize}
    \item 
$\mathcal{P}^h$ is the set of the undirected paths connecting $r_a$ and $r_b$ with length less than or equal to $h$.

\item $t_i$ is the \textit{i-th} triple in the path $P$ of the set $\mathcal{P}^h$.

\item $w(t)$ is the weight of the triple $t$, and $w_{max}$ is the maximum weight a triple in the graph can assume, according to one of the above three strategies.
\end{itemize}

\noindent In the case $r_a \equiv r_b$ the semantic relatedness between the resources is assumed to be equal to $1$.
On the basis of the results of the experimentation presented in \cite{Schuhmacher2014}, the measure obtained according to  \textit{combIC}  outperforms the other two.

\subsubsection{REWOrD}
According to \cite{Pirro2012}, the {\it REWOrD} method is based on the notion of \textit{informativeness} of predicates, in line with the \textit{Term Frequency-Inverse Document Frequency} (\textit{TF-IDF}) approach.
 \textit{TF-IDF} is generally used in information retrieval to evaluate the relevance of a term \textit{w} in a document \textit{d} belonging to a collection \textit{D} of documents.
The \textit{Term Frequency} (\textit{TF}) of the term \textit{w} with respect to the document \textit{d} represents the number of times \textit{w} appears in  \textit{d} divided by the total number of terms in \textit{d}.
The \textit{Inverse Document Frequency} (\textit{IDF}) represents the logarithm 
of the ratio between the total number of documents in \textit{D} and the number of documents containing the term \textit{w}.

In the case of an RDF graph, say $\mathcal{G}$, \textit{TF-IDF} deals with predicates instead of terms, and resources and triples instead of documents, therefore becomes \textit{Predicate Frequency-Inverse Triple Frequency} (\textit{PF-ITF}). As mentioned in Section \ref{Methods_weights}, we need to distinguish between incoming and outgoing \textit{Predicate Frequency} (\textit{PF}). 
In particular, the  incoming \textit{PF} of a predicate \textit{p} with respect to a resource \textit{r}, say $PF_{i}^r(p)$, resembles the \textit{TF} as defined in Eq. \ref{eq:pf_i}:

\begin{footnotesize}
\begin{equation}
\label{eq:pf_i}
PF_{i}^r(p) = \frac{ \vert T_i^r(p) \vert }{ \vert T^r \vert }
\end{equation}
\end{footnotesize}

\noindent where:
\begin{itemize}
    \item 
$ \vert T_i^r(p) \vert  =  \vert \{\langle?r_i, p, r\rangle \} \vert $ is the number of triples having predicate \textit{p} and  object \textit{r}, i.e., incoming to $r$.

\item $ \vert T^r \vert  =  \vert \{\langle r, ?p_i, ?r_k\rangle\} \cup  \{\langle?r_h, ?p_j, r\rangle\} \vert $ is the number of triples where the resource $r$ appears. 
\end{itemize}

\noindent The \textit{Inverse Triple Frequency} (\textit{ITF}) resembles the \textit{IDF} as defined in Eq. \ref{eq:idf}:

\begin{footnotesize}
\begin{equation}
\label{eq:idf}
ITF(p) = log\frac{ \vert \mathcal{G} \vert }{ \vert T(p) \vert }
\end{equation}
\end{footnotesize}

\noindent where:
\begin{itemize}

\item $ \vert \mathcal{G} \vert  =  \vert \{\langle ?r_i, ?p_j, ?r_h \rangle\} \vert $ is the total number of triples in the graph $\mathcal{G}$.

\item $ \vert T(p) \vert  =  \vert \{\langle ?r_i, p, ?r_j \rangle \} \vert $ is the number of triples with $p$ as predicate.

\end{itemize}

\noindent Finally, the \textit{incoming} \textit{PF-ITF} of the predicate $p$ with respect to the resource $r$, \textit{PF-$ITF_i^r$(p)}, is defined in Eq. \ref{eq:pfitf_i}:

\begin{footnotesize}
\begin{equation}
\label{eq:pfitf_i}
\textit{PF-}ITF_i^r(p)=PF_i^r(p) \cdot ITF(p)
\end{equation}
\end{footnotesize}

\noindent and stands for the \textit{informativeness} of the incoming predicate $p$ with respect to $r$. 
Analogously, the informativeness of the outgoing predicate $p$ with respect to $r$ is indicated as \textit{PF-$ITF_o^r$(p)}.
For example, consider the resource $r_a$ and the predicate $p_2$ in the knowledge graph of Figure \ref{fig:master}. We have  $ PF_i^{r_a}(p_2)= \frac{2}{9} = 0.22$, since the number of triples with predicate $p_2$ and object $r_a$ is $2$, and the number of triples where $r_a$ appears is $9$. Furthermore, $ITF(p_2) = log(\frac{22}{9}) = 0.39$, since the total number of triples in the graph is $22$, and the triples with predicate $p_2$ are $9$.
Therefore, $\textit{PF-}ITF_i^{r_a}(p_2)$, i.e., the informativeness of the incoming predicate $p_2$ with respect to $r_a$, is equal to $0.22 \cdot 0.39 = 0.08$.

The relatedness space of the resource $r$, say $RS(r)$, is the vector of  weighted predicates (either incoming or outgoing), where weights are the predicates' informativeness with respect to $r$.  
When addressing semantic relatedness between resources, their relatedness spaces can be enriched with the informativeness of the predicates occurring in the \textit{most informative path} (\textit{mip}) linking them. The \textit{mip} is the path with  the greatest informativeness, among those connecting the  resources, up to a given length.
Note that, the method considers undirected paths.
Given an undirected path \textit{P}$_n$ of length \textit{n}, the informativeness of \textit{P}$_n$ is the sum of the informativeness of the sub-paths of length 1, i.e., the single triples, divided by $n$, as defined according to Eq. \ref{eq:pathInf2}:

\begin{footnotesize}
\begin{equation}
\label{eq:pathInf2}
I({P_n}) = (I({t_{1})} + I({t_{2})} + ... + I({t_{n}}))/n
\end{equation}
\end{footnotesize}

\noindent where, for $i$ = $1$ \dots $n$:

$I({t_{i})} = I(\langle r_{i}, p_i, r_{i+1}\rangle) = (\textit{PF-}ITF_{o}^{r_{i}}(p_i) + \textit{PF-}ITF_{i}^{r_{i+1}}(p_i))/2$, 
 and 
$\langle r_{i}$, $p_i$, $r_{i+1} \rangle $ is the \textit{i-th} triple of the path $P_{n}$.

\noindent The method proposes five strategies for computing the relatedness between two resources $r_a$ and $r_b$,  referred to as \textit{reword\_{incoming}}, \textit{reword\_{outgoing}}, \textit{reword\_{average}}, \textit{reword\_{mip}}, and \textit{reword}. 

According to \textit{reword\_{incoming}}, the relatedness spaces for $r_a$ and $r_b$ are built by considering only the incoming predicates  to $r_a$ and $r_b$, respectively.

\begin{figure}
\centering
{\includegraphics[width=.70\textwidth]{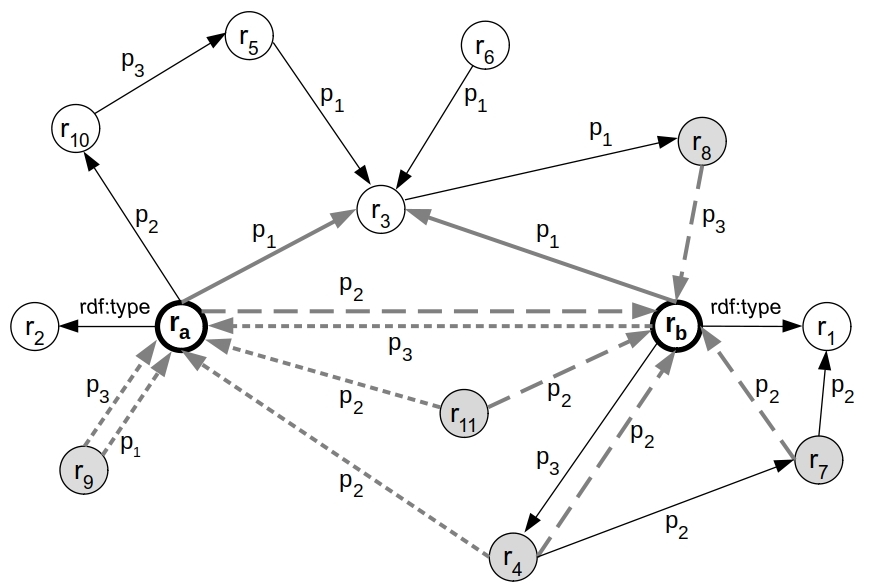}} 
\caption{\textit{REWOrD} applied to the running example graph}\label{fig:REWORD}
\end{figure}

For instance, consider  Figure \ref{fig:REWORD}, where the incoming predicates  to $r_a$ and $r_b$ are represented with dashed  and long dashed grey arrows, respectively.
On the basis of this strategy, the relatedness spaces of $r_a$, and $r_b$ are:

$RS(r_a) = [$ $(p_1,$ \textit{PF-ITF}$_i^{r_a}(p_1)), $
 $(p_2,$ \textit{PF-ITF}$_i^{r_a}(p_2)), $  
 $(p_3,$ \textit{PF-ITF}$_i^{r_a}(p_3)) $ $]$\\
\indent $RS(r_b) = [$   $(p_2,$ \textit{PF-ITF}$_i^{r_b}(p_2)), $
 $(p_3,$ \textit{PF-ITF}$_i^{r_b}(p_3)) $
$]$

\noindent   each containing the corresponding resource's incoming predicates,  associated  with their informativeness. 
Analogously, in the case of the \textit{reword\_{outgoing}}, only the outgoing predicates  from $r_a$ and $r_b$ are considered. The \textit{reword\_{average}} performs the arithmetic mean between the informativeness computed according to the \textit{reword\_{incoming}} and \textit{reword\_{outgoing}} strategies.
In the case of \textit{reword\_{mip}}, the  relatedness between $r_a$ and $r_b$ is evaluated by relying on the informativeness of the \textit{mip} between the resources.
Finally, according to the \textit{reword} strategy, the relatedness spaces of $r_a$ and $r_b$ as defined in the case of the \textit{reword\_incoming} approach are considered, both enriched with the informativeness of the predicates in the \textit{mip}. In particular, for each triple $\langle r_i, p_j, r_k\rangle$ in the \textit{mip}, the predicate $p_j$, and the related informativeness,  is added  to both the relatedness spaces of $r_a$ and $r_b$ and, if $p_j$ is already present in one or both the relatedness spaces, its informativeness will increase the existing ones. 
For example consider Figure \ref{fig:REWORD}, where we assume  that the \textit{mip} connecting $r_a$ and $r_b$ is the undirected path composed of the triples $\langle r_{a}, p_1, r_{3}\rangle$ and $\langle r_{b}, p_1, r_{3}\rangle$, highlighted in bold grey. According to the \textit{reword} strategy, the resulting relatedness spaces become:

$RS(r_a) = [$ $(p_1,$ \textit{PF-ITF}$_i^{r_a}(p_1)$ + 
$I(\langle r_{a}$, $p_1$, $r_{3}\rangle$) + 
$I(\langle r_{b}$, $p_1$, $r_{3}\rangle$)), \\
\indent \indent $(p_2,$ \textit{PF-ITF}$_i^{r_a}(p_2)), $
 $(p_3,$ \textit{PF-ITF}$_i^{r_a}(p_3)) $
$]$

$RS(r_b) = [$ $(p_1, $
$I(\langle r_{a}$, $p_1$, $r_{3} \rangle$) + 
$I(\langle r_{b}$, $p_1$, $r_{3} \rangle$)), \\
\indent \indent
$(p_2,$ \textit{PF-ITF}$_i^{r_b}(p_2)), $
 $(p_3,$ \textit{PF-ITF}$_i^{r_b}(p_3)) $
$]$

\noindent where the informativeness of both the triples of the \textit{mip} have been added to the informativeness of the already existing  predicate $p_1$ in the relatedness space of $RS(r_a)$, whereas a further element with the same informativeness has been added to the relatedness space of  $r_b$ since the predicate $p_1$ is not defined in  $RS(r_b)$.

Finally, the relatedness between $r_a$ and $r_b$ is computed as the \textit{cosine} between the two relatedness spaces.
In accordance with the results of the experimentation given in \cite{Pirro2012}, the \textit{reword} strategy outperforms the others.

\subsubsection{Exclusivity-based Measure (\textit{ExclM})}
As mentioned in Section \ref{Methods_weights}, the \textit{Exclusivity-based Measure} (\textit{ExclM}) computes the weight of a triple on the basis of the notion of  \textit{exclusivity} \cite{Hulpus2015}. 

Given a triple $t = \langle r_i, p, r_j \rangle$, the exclusivity of $t$ is  formally defined according to  Eq. \ref{eq:exclusivity}: 

\begin{footnotesize}
\begin{equation}
\label{eq:exclusivity}
exclusivity(\langle r_i, p, r_j \rangle) = \frac{1}{ \vert \{\langle r_i, p, ?r_x \rangle\} \vert  +  \vert \{\langle ?r_y, p, r_j \rangle \} \vert  - 1}
\end{equation}
\end{footnotesize}

\noindent where:
\begin{itemize}
    \item $\{\langle r_i, p, ?r_x \rangle\}$ is the set of triples with subject $r_i$ and predicate $p$.
\item $\{\langle ?r_y, p, r_j \rangle\}$ is the set of triples with object $r_j$ and predicate $p$.
\end{itemize}
Since the triple $\langle r_i, p, r_j \rangle$ belongs to both the above sets, 1 is subtracted at the denominator to avoid  that triple being counted twice. 

\begin{figure}
\centering
{\includegraphics[width=.70\textwidth]{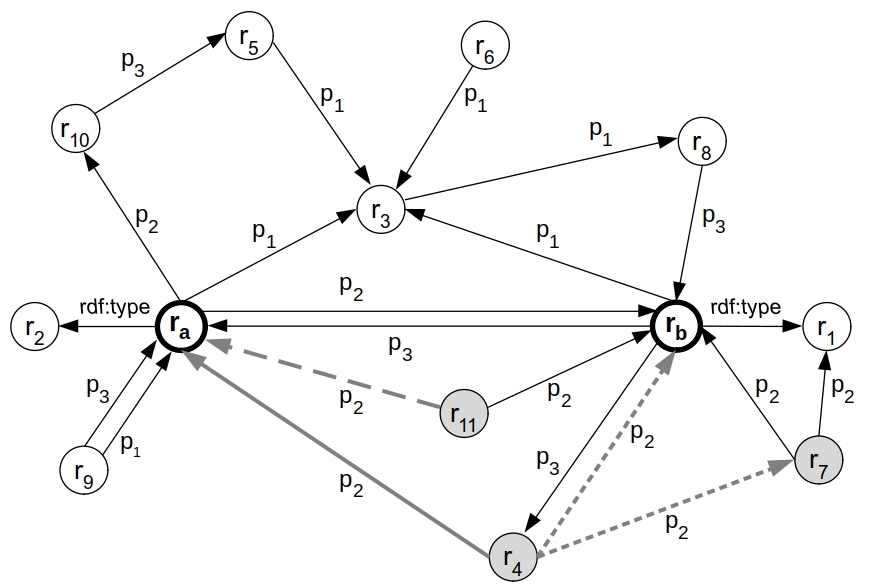}} 
\caption{\textit{ExclM} applied to the running example graph}\label{fig:EXCLUSIVITY}
\end{figure}

Consider in our running example the triple $t = \langle r_4, p_2, r_a \rangle$, highlighted with a bold grey arrow in  Figure \ref{fig:EXCLUSIVITY}.  
In addition to $t$, the  triples having predicate $p_2$ and subject $r_4$ are 2, whose edges are highlighted with dashed grey arrows in the same figure, therefore they are 3 in total.  
The triples having  predicate $p_2$ and object $r_a$ are 2, and are the triple $t$, and the triple drawn with a long dashed grey arrow.
Therefore, $exclusivity(t) = 1/(3+2-1) = 1/4 = 0.25$.

\noindent Based on the exclusivity function,  a path weighting function is introduced. This approach assumes that links in an RDF graph can be traversed in both directions and, for this reason, undirected paths are considered.

Let $P=[t_1, ..., t_n]$ be a sequence of triples representing an undirected path, the {\it weight} of $P$,  $w(P)$,  is defined in Eq. \ref{eq:excl_path_weight}.

\begin{footnotesize}
\begin{equation}
\label{eq:excl_path_weight}
w(P) = \frac{1}{\sum_{1}^{n}\frac{1}{exclusivity(t_i)}}
\end{equation}
\end{footnotesize}

\noindent Finally, given two resources $r_a$ and $r_b$, two integers $h$ and $k$ greater than $0$, and a constant $\alpha$, the relatedness $Eexcl_{h, k}^{\alpha}(r_a, r_b)$ is defined in Eq. \ref{eq:excl_rel}:

\begin{footnotesize}
\begin{equation}
\label{eq:excl_rel}
Excl_{h,k}^{\alpha}(r_a, r_b) = \sum_{P_i \in \mathcal{P}^h_k} \alpha^{length(P_i)} \cdot w(P_i)
\end{equation}
\end{footnotesize}

\noindent where:
\begin{itemize}
    \item 
$\mathcal{P}^h_k$ is the set of the top-k undirected paths with length less than or equal to $h$, among the ones connecting $r_a$ and $r_b$, i.e., the $k$ paths with the greatest weights.

\item $0 < \alpha \leq 1$ is a constant raised to the power of the length of the path $P_i$, \textit{length($P_i$)}, in $\mathcal{P}^h_k$ that inspired by the Katz's centrality measure, and aims at penalizing longer paths.
\item $w(P_i)$ is the weight of the path $P_i$ defined above.
\end{itemize}
In \cite{Hulpus2015}, the authors consider $k \in \{1, 5, 10\}$, and $\alpha \in \{0.25, 0.5, 0.75, 1\}$ in their experiment, and show that $k = 5$ and $\alpha = 0.25$ lead to better results.

\subsubsection{$ASRMP_m$} 
As mentioned in Section \ref{Methods_weights}, the $ASRMP_m$ family of relatedness measures originates from the previous proposal of the authors named  {\it Weighted Semantic Relatedness Measure} ($WSRM$) \cite{ElVaigh2019}. Therefore,
let us start by recalling the $WSRM$ measure and, successively, the $ASRMP_m$   family.

Consider an RDF graph $\mathcal G$, and the set $R$ of all the resources labeling such a graph, as defined in Section \ref{sec:rdf_linked_data}. Given two resources  $r_a$, $r_b$ $\in$ $R$, the $WSRM(r_a, r_b)$ between $r_a$, $r_b$ is defined in Eq. \ref{WSRM}:

\begin{footnotesize}
\begin{equation}
\label{WSRM}
WSRM(r_a, r_b) = \frac{ \vert \{p  \vert  \langle r_a, p, r_b \rangle \in \mathcal{G}  \} \vert  }{\sum_{r' \in R}  \vert \{p'  \vert  \langle r_a ,p', r' \rangle \in \mathcal{G}\} \vert   }
\end{equation}
\end{footnotesize}

\noindent where for any set S, $ \vert S \vert $ is the cardinality of  S.
According to the mentioned paper, the authors propose different strategies  to evaluate semantic relatedness.

\vspace{0.2 cm}
\noindent {\bf  {\textit {ASRMP}$_m^a$}}.
We start by recalling  the $ASRMP_m^a$ measure that considers all the paths between the compared resources of length \textit{equal} to $m$. In particular, given two resources $r_i$, $r_j$, $ASRMP_m^a(r_i, r_j)$ is defined as shown in Eq. \ref{ASRMP_a}:

\begin{footnotesize}
\begin{equation}
\label{ASRMP_a}
ASRMP_m^a(r_i, r_j) = \oplus_{q\in \mathcal{P}^m} \otimes_{k=1}^{m} WSRM(r_k,r_{k+1})
\end{equation}
\end{footnotesize}

\noindent where:
\begin{itemize}
    \item $\mathcal{P}^m$ is the set of the directed paths between $r_i$ and $r_j$ with length equal to $m$.
    \item $r_k$ is  the $k^{th}$ resource of the path $q$ (therefore $r_1$ = $r_i$, and $r_{m+1}$ = $r_j$).
    \item $\otimes$ and $\oplus$ are the {\it t-norm} and the related {\it s-norm} aggregators, respectively, the former for the edges of a given path, and the latter for  different paths of length $m$.
\end{itemize}

Note that, among the different aggregators available in the literature, the fuzzy logic operators {\it t-norm} for $\otimes$, and the  {\it s-norm} for $\oplus$, have been chosen by the authors in order to ensure transitivity.
In particular, according to the experimental results defined in the literature, the  Hamacher t-norm operator recalled in Eq. \ref{t-norm}:      

\begin{footnotesize}
\begin{equation}
\label{t-norm}
T_{H,0}(x,y)=  \frac{xy}{x+y-xy}
\end{equation}
\end{footnotesize}

\noindent with its associated s-norm, has been selected by the authors as the best aggregator.

\vspace{0.2 cm}
\noindent {\bf  {\textit {ASRMP}$_m^b$}}. Given the resources $r_i$, $r_j$, the second measure proposed by the authors is $ASRMP_m^b(r_i, r_j)$
that, with respect the previous one, aggregates all the paths of length {\it less than or equal to}  $m$, as defined in Eq. \ref{ASRMP_b}.

\begin{footnotesize}
\begin{equation} 
\label{ASRMP_b}
ASRMP_m^b(r_i, r_j) = \oplus_{q\in \mathcal{P}^m} \otimes_{k=1}^{ \vert q \vert } WSRM(r_k,r_{k+1})
\end{equation}
\end{footnotesize}

\begin{figure}
\centering
{\includegraphics[width=.70\textwidth]{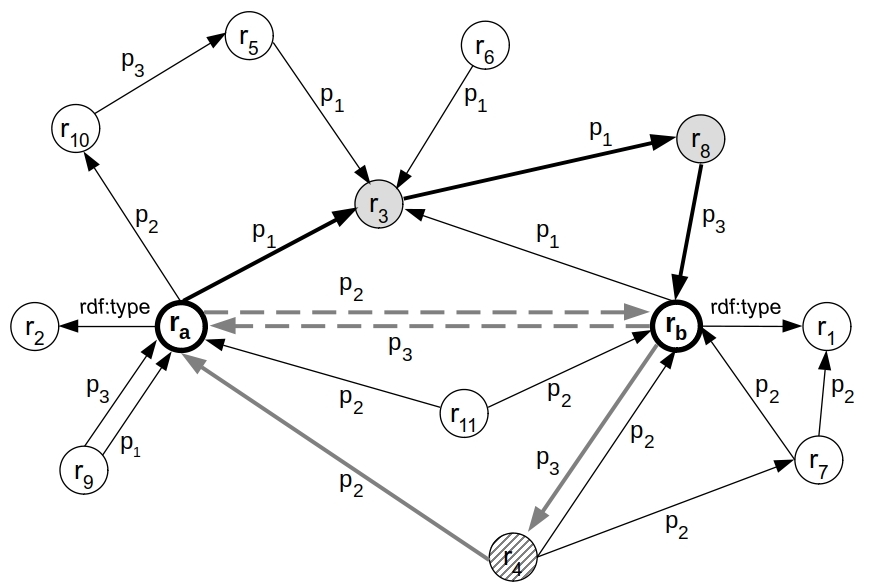}} 
\caption{\textit{ASRMP$_m$} applied to the running example graph}\label{fig:ASRMP}
\end{figure}

\noindent where $\mathcal{P}^m$ is the set of the directed paths between $r_i$ and $r_j$ of length less than or equal to $m$. However, the authors state that  direct links should represent stronger relations, whereas indirect ones should account for weaker relations and, therefore, the longer the path, the weaker the relation. For this reason, they propose a third measure, recalled below.

\vspace{0.2 cm}

\noindent {\bf  {\textit {ASRMP}$_m^c$}}. According to $ASRMP_m^c(r_i, r_j)$,  paths are weighted on the basis of their length $n$, $n$ = $1$..$m$, as shown in Eq. \ref{ASRMP_c}: 

\begin{footnotesize}
\begin{equation}
\label{ASRMP_c}
ASRMP_m^c(r_i, r_j) = \sum_{n=1}^m\sum_{q\in \mathcal{P}^n} w_n \otimes_{k=1}^{n} WSRM(r_k,r_{k+1})
\end{equation}
\end{footnotesize}

\noindent where:
\begin{itemize}
    \item $\mathcal{P}^n$ is the set of the directed paths between $r_i$ and $r_j$ with length equal to $n$.
    \item $w_n$ is a length-dependent weight, approximately corresponding to the percentage of paths of length n.
\end{itemize}

\noindent Finally, in order to achieve symmetry,  the three strategies above are reformulated according to the $\psi_m^x(r_i,r_j)$  relatedness measure defined in  Eq. \ref{ASRMP_media}.

\begin{footnotesize}
\begin{equation}
\label{ASRMP_media}
\psi_m^x(r_i,r_j) = \frac{1}{2}(ASRMP_m^x(r_i, r_j)+ASRMP_m^x(r_j, r_i)), \,\,\,\,\,  x \in \{a,b,c\}
\end{equation}
\end{footnotesize}

\noindent Consider again the resources $r_a$, $r_b$ of the running example of Figure \ref{fig:master}, and assume   $m$ = $3$.
In the case of the measure  $\psi_3^a(r_a,r_b)$\footnote{Superscript $a$ in the name of the measure $\psi_m^a$ and subscript $a$ in the name of the resource $r_a$ is just a case occurring in this running example, and analogously later in the case of $b$, for $\psi_m^b$ and  $r_b$.},  the paths of length $3$ are considered that are represented  by the  only path
[$\langle r_a,p_1,r_3\rangle$, $\langle r_3,p_1,r_8\rangle$, $\langle r_8,p_3,r_b\rangle$],  that is highlighted with bold arrows in Figure \ref{fig:ASRMP}. Whereas, in the case of the measure $\psi_3^b(r_a,r_b)$, besides the previous one, also the paths with lengths less than $3$ are addressed, that are the one of length $2$, i.e.,   [$\langle r_b,p_3,r_4\rangle$, $\langle r_4,p_2,r_a\rangle$], highlighted with bold grey arrows in Figure \ref{fig:ASRMP}, and  the ones of length $1$, i.e., [$\langle r_a,p_2,r_b\rangle$], and [$\langle r_b,p_3,r_a\rangle$], represented with  long dashed grey arrows in the same figure. 

Among the proposed strategies, the authors state that $ASRMP_m^a$ is the best one, in particular for Entity Linking tasks.

\subsubsection{Proximity-based Method (\textit{ProxM})}
\label{sec:proximity_2}
According to the \textit{Proximity-based Method} (\textit{ProxM}) \cite{Leal2013},
given the resources $r_a$ and $r_b$, and an integer $h$ standing for the maximum length of a path, the relatedness ({\it proximity}) between them, $prox_h(r_a, r_b)$, is defined in Eq. \ref{eq:proximity}: 

\begin{footnotesize}
\begin{equation}
\label{eq:proximity}
prox_h(r_a, r_b) = \frac{1}{\Omega(\mathcal{G})}\sum_{n=1}^{h} \frac{1}{2^n\Delta(\mathcal{G})^n}\sum_{P \in \mathcal{P}^n} \sum_{t_i\in P}w(p_i)
\end{equation}
\end{footnotesize}

\noindent where: 

\begin{itemize}
\item $\Omega(\mathcal{G})$ is the maximum weight a predicate in $\mathcal{G}$  can be associated with.

\item $\Delta(\mathcal{G})$ is the maximum outdegree of the nodes in $\mathcal{G}$. 

\item $\mathcal{P}^n$ is the set of  the undirected paths connecting $r_a$ and $r_b$ with length $1 \leq n \leq h$.

\item $w(p_i)$ is a function that associates the predicate $p_i$ of the triple $t_i$ with a weight,  which is manually assigned in the experiment provided in \cite{Leal2013}.

\end{itemize}

\noindent Finally, if  $r_a \equiv r_b$, $prox_h(r_a, r_b)$ is assumed to be equal to 1.

\begin{figure}
\centering
{\includegraphics[width=.70\textwidth]{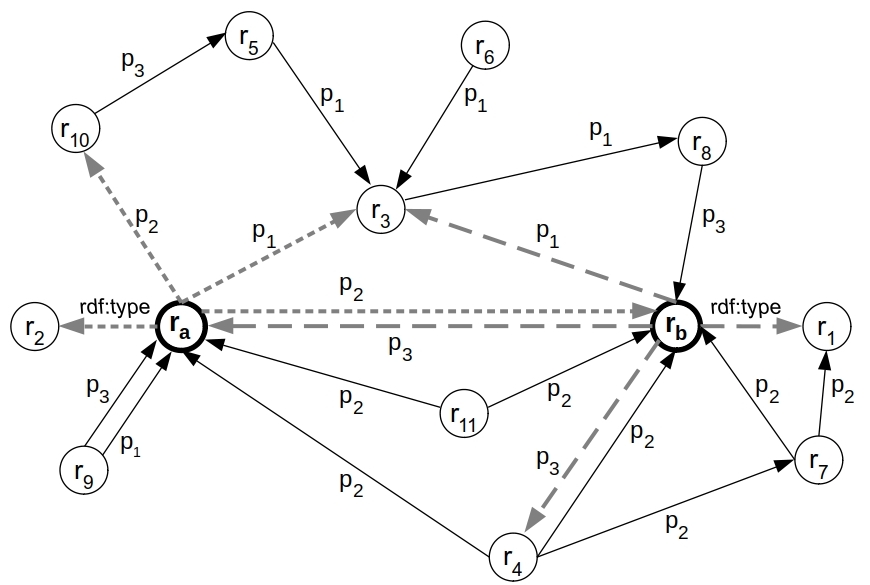}} 
\caption{\textit{ProxM} applied to the running example graph}\label{fig:PROXIMITY}
\end{figure}

For instance, consider the graph of our running example. We have $\Delta(\mathcal{G}) = 4$ since 4 is the maximum outdegree of the nodes of the graph. In particular, both $r_a$ and $r_b$ have outdegree  equal to 4 (see the  dashed grey and long dashed grey lines, respectively, in Figure \ref{fig:PROXIMITY}).
Furthermore, suppose that the predicates $p_1$, $p_2$, $p_3$, $p_4$, and \textit{rdf:type} have been associated with the weights $0.5$, $0.3$, $0.2$, $0.7$, and $0.6$, respectively, then $\Omega(\mathcal{G})  = 0.7$  that is the maximum among the predicates weights.

\end{document}